\documentclass[3p,preview]{elsarticle}

\usepackage[utf8]{inputenc}
\usepackage{graphicx} 
\usepackage[hidelinks]{hyperref}
\usepackage{tabularx}
\usepackage{multirow}
\usepackage{xcolor}
\usepackage{subcaption}
\usepackage{amsmath,amsfonts,bm}

\newcommand{\pder}[2]{\frac{\partial #1}{\partial #2}}
\newcommand{\pderInl}[2]{\partial #1 / \partial #2}

\newcommand{\gradfn}[2]{\partial #1 \left( \partial #2 \right)}
\newcommand{\gradfnN}[3]{\partial #1 \left( \partial #2 \right)_{#3}}

\newcommand{\RHS}[1]{{#1}^{\text{RHS}}}
\newcommand{\vel}{\mathrm{\mathbf{u}}}
\newcommand{\velRHS}{\RHS{\vel}}
\newcommand{\velcomp}{u}
\newcommand{\velcompRHS}{\RHS{\velcomp}}
\newcommand{\velu}{u} 
\newcommand{\velv}{v}
\newcommand{\velw}{w}
\newcommand{\dens}{\rho}

\newcommand{\pressure}{p}
\newcommand{\pressureRHSvec}{\mathrm{\mathbf{h}}}
\newcommand{\pressureRHSveccomp}{h}
\newcommand{\pressureRHS}{h}
\newcommand{\viscosity}{\nu}

\newcommand{\source}{S}

\newcommand{\guess}{*}
\newcommand{\AdvMat}{C}
\newcommand{\AdvDiag}{A} 
\newcommand{\AdvDiagInv}{\AdvDiag^{-1}}
\newcommand{\AdvNDiag}{H}

\newcommand{\PMat}{P}

\newcommand{\velwall}{\velcomp_\tau}
\newcommand{\Rewall}{\text{Re}_\tau}
\newcommand{\Recenter}{\text{Re}_\text{cl}}

\newcommand{\stats}{\lambda}
\newcommand{\statsSet}{\Lambda}

\newcommand{\ETTeq}{t \velwall / \delta}
\newcommand{\ETTtext}{ETT}
\newcommand{\ETT}{\ETTtext} 

\newcommand{\normal}{\vec{n}}

\newcommand{\laxis}{\xi}

\newcommand{\transform}{{\mathbf{T}^{-1}}}
\newcommand{\transforminv}{\mathbf{T}}
\newcommand{\transformJ}{J}

\newcommand{\cvel}{\mathbf{U}}
\newcommand{\cvelcomp}{U}

\newcommand{\facesign}{N}
\newcommand{\cell}{P}
\newcommand{\nb}{F}
\newcommand{\area}{a}
\newcommand{\Snb}{\textbf{\nb}}
\newcommand{\nbf}{f}
\newcommand{\Snbf}{\textbf{\nbf}}
\newcommand{\nbb}{B}
\newcommand{\Snbb}{\textbf{\nbb}}
\newcommand{\nbfb}{b}
\newcommand{\Snbfb}{\textbf{\nbfb}}
\newcommand{\tangentialDir}{{D}}
\newcommand{\nbd}{{\tangentialDir_\nb}}
\newcommand{\Snbd}{{\textbf{\tangentialDir}_\nb}}
\newcommand{\nbdf}{{\tangentialDir_\nbf}}
\newcommand{\nbdfb}{{\tangentialDir_\nbfb}}
\newcommand{\Snbdfb}{{\textbf{\tangentialDir}_\nbfb}}

\newcommand{\evalloc}[2]{{\left[ #1 \right]_{#2}}}
\newcommand{\evalcell}[1]{\evalloc{#1}{\cell}}
\newcommand{\evalnb}[1]{\evalloc{#1}{\nb}}

\newcommand{\evalnbf}[1]{\evalloc{#1}{\nbf}}
\newcommand{\evalnbfb}[1]{\evalloc{#1}{\nbfb}}

\newcommand{\lerp}[1]{\overline{#1}}

\newcommand{\NNw}{\theta}

\newcommand{\CFL}{\text{CFL}}
\newcommand{\dt}{\Delta t}

\newcommand{\Lowres}{\text{No-Model}} 
\newcommand{\NNeight}{\text{NN}_{8}}
\newcommand{\NNsixteen}{\text{NN}_{16}}
\newcommand{\NNthirty}{\text{NN}_{30}}
\newcommand{\NNforty}{\text{NN}_{40}}

\newcommand{\myshortname}{PICT} 
\newcommand{\quoteEN}[1]{``#1''}
\newcommand{\myrefsec}[1]{section~\ref{#1}}
\newcommand{\myrefapp}[1]{\ref{#1}}
\newcommand{\myreffig}[1]{figure~\ref{#1}}
\newcommand{\myreftab}[1]{table~\ref{#1}}
\newcommand{\myrefeq}[1]{eq.~(\ref{#1})}

\title{\myshortname ~-- 
A Differentiable, GPU-Accelerated Multi-Block PISO Solver for Simulation-Coupled Learning Tasks in Fluid Dynamics
}
\date{2025}

\author[1]{Aleksandra Franz\corref{cor1}}
\ead{aleksandra.franz@tum.de}

\author[1]{Hao Wei}
\ead{hao.wei@tum.de}

\author[1]{Luca Guastoni}
\ead{luca.guastoni@tum.de}

\author[1]{Nils Thuerey}
\ead{nils.thuerey@tum.de}

\cortext[cor1]{Corresponding Author}

\affiliation[1]{organization={Technical University of Munich},
    city=Munich,
    country=Germany}

\begin{document}

\begin{abstract}
    Despite decades of advancements, the simulation of fluids remains one of the most challenging areas of in scientific computing. Supported by the necessity of gradient information in deep learning,
    differentiable simulators have emerged as an effective tool for optimization and learning in physics simulations. 
    In this work, we present our fluid simulator \textit{PICT}, a differentiable pressure-implicit solver coded in PyTorch with Graphics-processing-unit (GPU) support.
    We first verify the accuracy of both the forward simulation and our derived gradients in various established benchmarks like lid-driven cavities and turbulent channel flows
    before we
    show that the gradients provided by our solver can be used to learn complicated turbulence models in 2D and 3D. We apply both supervised and unsupervised training regimes using physical priors to match flow statistics.
    In particular, we learn a stable sub-grid scale (SGS) model for a 3D turbulent channel flow purely based on reference statistics.
    The low-resolution corrector trained with our solver runs substantially
    faster than the highly resolved references, while keeping or even surpassing their accuracy.
    Finally, we give additional insights into the physical interpretation of different solver gradients, and motivate a physically informed regularization technique.
    To ensure that the full potential of PICT can be leveraged, it is published as open source: \url{https://github.com/tum-pbs/PICT}.
\end{abstract}


\maketitle

\section{Introduction}

The simulation of fluids represents one of the most challenging and computationally demanding areas in scientific computing. Accurate and efficient simulations of fluid dynamics are essential across a broad range of applications, from engineering design to climate modeling.
Starting from the early direct numerical simulations (DNSs) performed by Kim et al.~\cite{Kim_Moin_Moser_1987}, numerical simulation has emerged as an effective tool for both scientific discovery and engineering design. These simulations 
rely on numerical solvers, whose accuracy plays an important role when investigating the physics of fluid flows.
In many engineering applications, computation is performed on a mesh fitted into the domain, using algorithms based on finite differences~\cite{leveque2007finite}, finite volumes~\cite{versteeg2007introduction}, or spectral elements~\cite{patera1984spectral}. While other higher accuracy options are available, the \textit{Pressure Implicit with Splitting of Operators} (PISO) algorithm~\cite{issa1986solution} remains a popular method for the simulation of incompressible flows due to its simplicity and stability, as well as its inclusion in software packages such as OpenFOAM~\cite{weller1998tensorial}. While other formulations have been proposed to tackle specific problems~\cite{Issa01122001,NORDLUND2016199}, the original algorithm is still widely used.

In this work, we introduce a novel implementation of a solver for incompressible fluid flows based on the PISO scheme called \textit{PICT}, whose distinguishing feature is its differentiability. This allows gradients to flow through the solver, enabling end-to-end treatment of various optimization and machine learning (ML) tasks. 
Differentiable simulations have become increasingly prevalent in the field of robotics~\cite{hu2019difftaichi, freeman2021braxdifferentiablephysics, howell2023dojodifferentiablephysicsengine}, and are garnering growing interest within the fluid dynamics community~\cite{jaxfluids2023, holl2024phiflow, fan2024differentiable}.
Differentiable solvers can be employed to optimize time-dependent problems step-by-step, a scenario that would be computationally very expensive to address with traditional adjoint formulations~\cite{griewank2003mathematical}.
The use of differentiable solvers~\cite{hu2019difftaichi,holl2020,Sirignano2020} in combination with machine learning algorithms can provide significant advantages. Typical supervised machine learning methods attempt to learn from examples and neural network models are optimized based on the gradient with respect to the mismatch between the network output and the reference ground truth. While these methods work well in tasks like 
replacing a numerical solver~\cite{thuerey2020deep,fukami2020assessment},
super-resolution~\cite{fukami2019super,xie2018},
flow field reconstruction~\cite{sirignano2018dgm,Guastoni2020},
or turbulent inflow generation~\cite{Fukami2019TurbulentInflowGenerator},
trying to predict the temporal evolution of a physical system can lead to the accumulation of errors~\cite{kohl2023turbulent}. This issue can be addressed by integrating this evolution into the training via \textit{unrolling}~\cite{um2020solver,list2025differentiability}, and by learning to predict consistent sequences. 
This concept has been successfully applied to turbulence modeling~\cite{list2022_Learned,sirignano2023pde,agdestein2025discretize}, for solver acceleration~\cite{Kochkov}, and weather forecasting~\cite{lam2022graphcast}.
Similar to inductive biases like curl formulations \cite{mohan2023embedding}, differentiable solvers inherently incorporate physical constraints into the optimization process, reducing the risk of physically implausible solutions. E.g. they can restrict correction terms to be divergence free in incompressible settings~\cite{list2022_Learned}, wheres physics-informed losses would yield soft constrains~\cite{tompson2017}.
In order to facilitate the integration of machine learning models in the simulations, we implement our solver in the widely used \textit{PyTorch} framework~\cite{pytorch}. It will be made available as open source software\footnote{\url{https://github.com/tum-pbs/PICT}}, leverages GPU acceleration, and provides custom gradient computations tailored to the solver.

One application in which temporal consistency is particularly important is the development of learned turbulence closure models~\cite{Duraisamy_2019}. Different supervised approaches have been tested, both in the context of subgrid-scale (SGS) modeling~\cite{beck2018deep, sarghini2003neural, Xie2019NNLES} and wall-modeling. The latter approach has been implemented by training a classifier to choose among different models~\cite{Lozano2023}, or by using physics-informed neural networks~\cite{Yang2019wallPINN}.
Turbulence models for steady-state solutions via Reynolds-averaging have likewise been targeted with a PINN approach~\cite{wu2018physics}. 
In order to address the problem of accumulating errors for transient problems, reinforcement learning (RL) based solutions to the turbulence modeling problem have been proposed~\cite{turbulenceMARL2021, SciMARL2022, beck2023towards,koh2025physics}. By formulating the problems as a Markov decision process (MDP), these approaches prevent the limitations of auto-regressive predictions in a purely supervised a-priori context. It should be noted, however, that the training procedure in this case requires extensive exploration of the solution space and induces high sample complexity. Differentiable solvers, on the other hand, directly compute gradients of the loss function with respect to system parameters, effectively yielding an a-posteriori training~\cite{sanderse2024scientific}
and bypassing the need for iterative policy updates or extensive simulation runs. In this paper, we demonstrate the potential of our solver by training neural network models on wake and obstacle flows in 2D, and as subgrid-scale models for a turbulent channel flow (TCF) in 3D.
Our paper is aligned with recent works that have investigated the use of differentiable solvers for SGS models using graph neural networks~\cite{kim2024generalizabledatadriventurbulenceclosure} and for shell models of turbulence~\cite{freitas2024solverintheloopapproachturbulenceclosure}.
A distinctive feature of our work is that the PICT solver allows for training the SGS model while only supervising in terms 
of velocity moments. I.e.,~no pre-computed training data sets are required in this case, and training only induces a moderate computational cost. While the training process itself is stochastic, giving no formal guarantees on the exact results obtained, the learned model significantly outperforms reference solvers in accuracy and runtime, and retains the target statistics for arbitrary timespans and different initial conditions.

The remainder of the article is organized as follows: in \myrefsec{sec:method}, we describe the problem setting and the solution algorithm, with particular focus on the gradient computation using automatic differentiation (AD).
In \myrefsec{sec:learningloss}, we describe the optimization tasks that we use to showcase the potential of the solver's differentiability.
Furthermore, our solver is carefully validated with respect to the forward and backwards passes, see \myrefsec{sec:validation} and \myrefapp{app:validation} for details.
In \myrefsec{sec:DLresults}, the results of challenging flow modeling tasks in 2D and 3D are reported. 
Finally, in \myrefsec{sec:conclusion}, we provide a summary and the conclusions of the study.

\section{Solution and Backpropagation Algorithm} \label{sec:method}
\begin{figure}
    \centering
    \includegraphics[width=\textwidth]{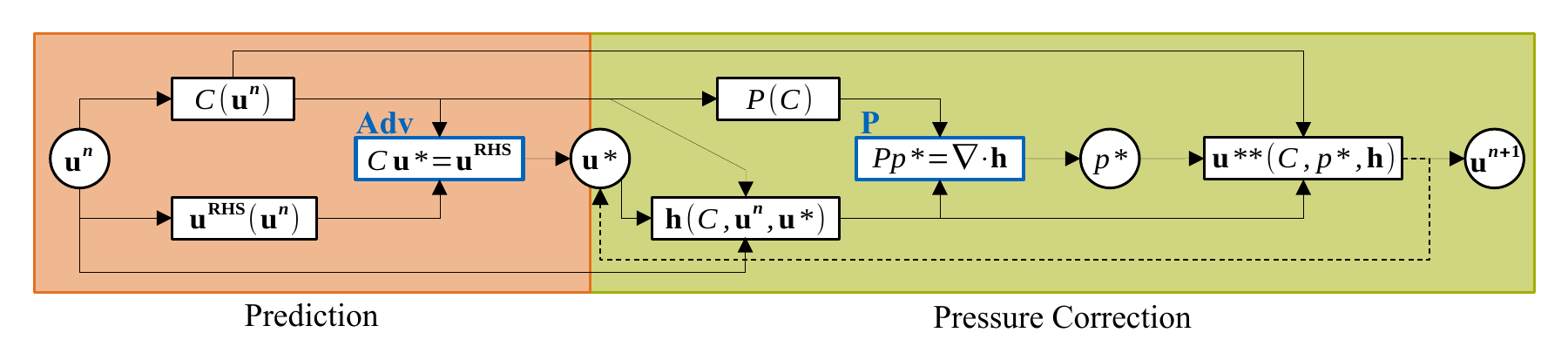}
    \caption{A flow chart showing the high level computational graph of our solver.
    Viscosity, boundaries, velocity sources, transformations, and non-orthogonal correction have been omitted for clarity.
    The dashed line represents the pressure correction loop, and both linear solves (\textit{Adv} and \textit{P}) are highlighted in blue.
    All shown paths are differentiable.
    }
    \label{fig:PISOgraph}
\end{figure}

In this section, we lay the necessary groundwork by introducing
the various building blocks and algorithms used to implement our simulator,
before discussing its differentiability and the opportunities that arise from our modular approach.

\subsection{The PISO Algorithm}
With velocity $\vel$, pressure $\pressure$, viscosity $\viscosity$, external sources $\source$, and time $t$,
the governing Navier-Stokes equations that describe the dynamics of incompressible flows take the form of
momentum equation
\begin{equation} \label{eq:momentum}
    \pder{\vel}{t} + \nabla \cdot (\vel \vel) - \viscosity \nabla^2 \vel  = -\nabla \pressure + \source
\end{equation}
and continuity equation
\begin{equation} \label{eq:continuity}
    \nabla \cdot \vel = 0.
\end{equation}
To simulate these dynamics, we use the PISO algorithm introduced by Issa \cite{issa1986solution},
which comprises a predictor step to solve the momentum equation (\ref{eq:momentum})
and a corrector step to enforce continuity (\ref{eq:continuity}).
The predictor step advances the simulation in time, resulting in a velocity guess $\vel^\guess$,
and is typically followed by 2 corrector steps that each compute a pressure which in turn is used to make $\vel^\guess$ divergence free.
We discretize the PISO algorithm on a collocated grid using the finite volume method (FVM)
by following Maliska \cite{maliska2023fundamentals} and Kajishima and Taira \cite{kajishima2016computational}, adapting their formulations to the PISO structure.
As we chose an implicit Euler scheme for the time advancement,
the discretization produces two linearized systems, the first being
\begin{equation}\label{eq:implEuler1}
    \AdvMat \vel^\guess = \frac{\vel^n}{\Delta t} - \nabla \pressure + \source
\end{equation}
where the (sparse) matrix $\AdvMat$ contains the advection and diffusion terms and which is solved  for the velocity guess $\vel^\guess$. The second one, with $\AdvDiag$ being the diagonal of $\AdvMat$, is
\begin{equation}\label{eq:implEuler2}
    \nabla^2 (\AdvDiagInv \pressure^{\guess}) = \nabla \cdot \pressureRHSvec ,
\end{equation}
which is solved for a pressure that makes the velocity guess (included in $\pressureRHSvec$) divergence free.
Details about the exact PISO formulation used and the discretization can be found in \ref{app:method}, a schematic overview is shown in \myreffig{fig:PISOgraph}.

\subsection{Multi-Block Grids and Transformations} \label{sec:MBGtransform}
\begin{figure*}
    \centering
    \includegraphics[width=\textwidth]{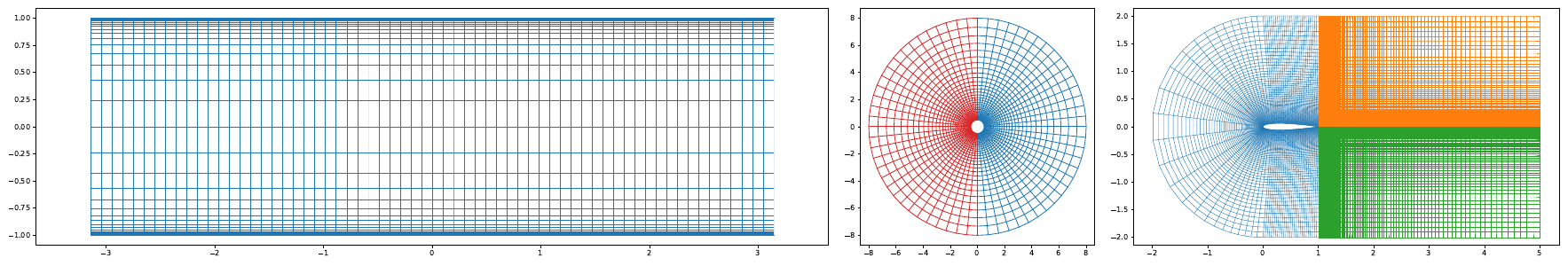}
    \caption{Three examples of transformed multi-block meshes that can be handled by our simulator.
    From left to right:
    a channel flow grid refined towards the walls,
    a ring grid with a round obstacle at the center,
    a refined C-grid around a NACA 0012 airfoil.
    The meshes have been coarsened for improved visibility and colors indicate different blocks.
    } \label{fig:meshes}
\end{figure*}
As PICT is intended to be a general, differentiable solver for the incompressible Navier-Stokes equations, applicable to a broad range of learning tasks for fluid simulation, we choose a first order finite volume scheme for discretization.
The domain of interest is further split into multiple blocks,
based on the geometry of the boundaries,
in order to obtain a multi-block grid, examples of which are shown in \myreffig{fig:meshes}.
Each block comprises a regular grid of quadrilateral (2D) or hexahedral (3D) elements that can be refined and aligned to boundaries with precomputed transformations.
These transformations are represented as matrix $\transforminv$ where the elements relate the Cartesian computational grid space with directions $\laxis_j$ to the physical space with directions $x_i$ via
$\transforminv_{ji} = \pderInl{\laxis_j}{x_i}$.
Each side of a block can have a single boundary specified, either a connection to another block with matching resolution to create a conformal mesh or a prescribed quantity.
Each block has separate velocity and pressure tensors that together make up the global fields.
The linear systems for prediction and correction steps are assembled from the blocks and solved globally for the complete domain.

Compared to unstructured meshes, our approach makes memory handling and flux computations easier while still allowing boundary refinement and alignment. The multi-block grid structure allows for a straightforward integration of convolutional neural networks (CNNs) for deep-learning tasks, only requiring custom padding for the block connections to avoid artifacts at the block edges, which we provide for \myshortname's multi-block structure. On the other hand, unstructured meshes would require more costly graph networks or resampling.
Compared to a cut-cell approach \cite{salih2019thin},
where boundaries arbitrarily intersect cells, the boundary handling is simpler and unused regions of memory are avoided.
A potential drawback of the multi-block grids is the relatively complicated mesh generation of fitting multiple regular grids to complex geometries,
which influences solver stability and accuracy \cite{spekreijse2002simple}.

\subsection{Differentiation of the Discrete Algorithm} \label{sec:diffdiscrete}

For many optimization tasks, including the training of neural networks in deep learning, derivatives are required. Typical algorithms 
rely on scalar loss functions, and hence it is sufficient to compute a gradient vector  with respect to the quantity being optimized. The gradient is the result of contracting the full Jacobian with respect to the scalar loss. 
In our setting, this translates to the requirement to provide gradients of the solver's output with respect to its inputs or other parameters in the form of Jacobian-vector-products.
As an example, we consider the optimization of an initial velocity field $\vel^0$ based on a loss $L(\vel^n)$ on a converged state $\vel^n$. The gradient $\pderInl{L(\vel^n)}{\vel^0}$ can be acquired by backpropagation through the complete rollout of the simulation using the chain rule:
\begin{equation}
    \pder{L(\vel^n)}{\vel^0} = \pder{L(\vel^n)}{\vel^{n}} \prod_{i=1}^{n} \pder{\vel^i}{\vel^{i-1}}.
\end{equation}
Generally speaking, there are two approaches to compute these gradients~\cite{kidger2022neuraldifferentialequations}: \quoteEN{Discretize-then-Optimize} (DtO), which corresponds to standard backpropagation, provides accurate gradients based on the numerical discretization of the PDE,
while \quoteEN{Optimize-then-Discretize} (OtD), also called \textit{continuous adjoint method}, consists in solving numerically an additional differential equation backwards in time.
In comparison, DtO requires more memory as intermediate results of the compute graph need to be stored, but it is fast and provides accurate gradients. It also requires a solver written in an AD framework, or, as in our case, custom analytical gradients.
On the other hand, OtD tends to be slower due to the full backwards solve, which has to be derived for the specific problem at hand,
and the gradients tend to be less accurate with respect to the forward discretization.
OtD requires less memory, but for non-linear problems the solution of the forward problem still needs to be stored at different timesteps as 
required for the adjoint solve.

In our algorithm, we combine the two approaches: we employ DtO for the overall structure and backpropagate the gradients through the computational graph of our solver, as depicted in \myreffig{fig:PISOgraph}, while we use OtD for the embedded solution of the linear systems.
When computing gradients through the linear solvers, we do not backpropagate through the solution procedure of $Ax=b$ (here with a general matrix $A$) but instead solve the system $A^T \partial b= \partial x$ for $\partial b$ given an output gradient $\partial x$ \cite{giles2008extended}, which corresponds to an OtD treatment of these operations.
The gradients with respect to the matrix entries are computed with the outer product $\partial A = - \partial b \otimes x$. Since $A$ is a sparse matrix in our case, only elements that exist on $A$ are realized on $\partial A$.

For the remaining operations we derive the analytical gradients based on our discretization of the PISO algorithm.
E.g. the components of the pressure gradient used to correct the velocity guess via
$\vel^{\guess\guess} = \pressureRHSvec - \nabla \pressure$ are computed with finite differences as
\begin{equation}
    \left( \nabla \pressure \right)_i = \sum_j \transforminv_{ji} \frac{\pressure_{j+1} - \pressure_{j-1}}{2}.
\end{equation}
The analytical gradient of this discretization is 
\begin{equation}
    \gradfn{\pressure}{\vel^{\guess\guess}} = \sum_{\nb \rightarrow j} -0.5 \facesign_\nbf \evalnb{\left( \transforminv \AdvDiagInv \partial \vel^{\guess\guess} \right)_j},
\end{equation}
where the sum traverses all neighboring cells $\nb$, $j$ indicates the computational normal axis of the connecting face, and $\facesign_\nbf$ is the sign (direction) of that face.

These individual gradient functions are then chained together with the backwards linear solves in an AD framework to enable back-propagation of gradients through the whole PISO algorithm.
This has the added benefit that we can manually chose which inputs need to be kept for backpropagation,  further reducing memory requirements.
This modular DtO approach also makes the solver and its gradients highly customizable. Changes and additions, such as adding a learned corrector or replacing the pressure correction with a learned alternative, can be made 
flexibly within PyTorch's AD framework.
In the OtD setting, such changes would require a new derivation of the adjoint problem.
Details of the gradients of the individual operations are reported in \ref{app:gradients}.

\subsection{Gradient Paths}\label{sec:gradpaths}

An interesting opportunity that arises from the modular design is the possibility to investigate individual parts of the gradient functions and paths with respect to their impact on accuracy and runtime.
Thus, choosing appropriate gradients for a given optimization task yields the flexibility to reduce the optimization time without impacting the result accuracy.
This is especially attractive when training deep neural networks, where many optimization iterations (gradient update steps) and thus many simulation roll-outs can be required to optimize the network parameters.
We make the following observations regarding optimization tasks that utilize a differentiable solver, be it direct optimization or deep learning applications:
When training, the loss is expected to decrease over time, meaning that a neural network is typically very approximate at first, converging towards more accurate solutions over the course of training.
Furthermore, in our solver, the advection solve drives the dynamics, and is thus also responsible for the transport of gradients in the backwards pass, while the pressure solve has a primarily diffusive influence.
In the following, we consider several approximate variants of the gradient calculation, in addition to the full backpropagation. These variants are particularly interesting for resource efficiency, which becomes an important consideration for scaling up learning and optimization methods to larger, three-dimensional scenarios.

As can be seen in the computational graph in \myreffig{fig:PISOgraph}, there are several paths through the PISO step for a gradient from output to input, i.e., $\pderInl{\vel^{n}}{\vel^{n-1}}$.
Formulating derivatives for all steps of the full PISO algorithm as detailed in \myrefapp{app:gradients},
we identify three central, additive groups of Jacobians from the computational graph:
$\pderInl{\vel^{n}}{\vel^{n-1}} = J^\text{{Adv}} + J^P  + J^{\text{none}} $, 
where $J^{\text{Adv}}$ denotes the Jacobians for backpropagation paths that pass through the linear solve for advection, $\AdvMat \vel^\guess = \velRHS$, and $J^P$ the ones through the pressure solve, $\PMat \pressure^\guess = \nabla \cdot \pressureRHSvec$. Interestingly, this leaves a third set of \textit{bypass} paths $J^{\text{none}}$ in the PISO algorithm which consists of
\begin{equation}\label{eq:jnone}
    \begin{split}
        &\pder{\AdvMat \left( \vel^{n-1} \right)}{\vel^{n-1}}
        \pder{\vel^{n}\left( \AdvMat, \pressure^\guess, \pressureRHSvec \right)}{\AdvMat},\\
        &\pder{\AdvMat \left( \vel^{n-1} \right)}{\vel^{n-1}}
        \pder{\pressureRHSvec \left( \AdvMat, \vel^{n-1}, \vel^\guess \right)}{\AdvMat}
        \pder{\vel^{n}\left( \AdvMat, \pressure^\guess, \pressureRHSvec \right)}{\pressureRHSvec},\text{ and}\\
        &\pder{\pressureRHSvec \left( \AdvMat, \vel^{n-1}, \vel^\guess \right)}{\vel^{n-1}}
        \pder{\vel^{n}\left( \AdvMat, \pressure^\guess, \pressureRHSvec \right)}{\pressureRHSvec} , 
    \end{split}
\end{equation}
assuming a single corrector step for simplicity.
Despite bypassing both linear solves, these terms still provide direct per-cell contributions, e.g. the last term provides $\frac{\partial \vel^{n}}{\AdvDiagInv \Delta t}$ to $\partial \vel^{n-1}$, and hence represent a gradient flow from output to input of the solver.
In light of the previous discussion, especially the observation that a NN is expected to be inaccurate in the early phases of training,
the question arises whether all three terms contribute equally strongly to the update direction of learning and optimization tasks.
This is especially interesting as computing the three terms shows huge differences in computational complexity:
both $J^\text{{Adv}}$ and $J^P$ involve solving a linear system, are correspondingly expensive, typically showing a super-linear complexity in terms of system size $N$. 
The term $J^{\text{none}}$, on the other hand, stems from relatively simple operations with vectors. Correspondingly, it is linear in $N$, and can be computed efficiently.
Without backpropagation through the advection via $J^\text{{Adv}}$, the influence of the dynamics is missing from the gradients and errors are mainly propagated cell by cell to the previous time step via $J^{\text{none}}$. Nonetheless, as we show in 
\myrefsec{sec:gradPathAblation}, this is a suitable approximation when the error is high or the optimization problem is not primarily driven by the dynamics. 
This is, e.g., suitable for cases with shorter rollouts, where transport dynamics play only a minor role and $J^\text{{Adv}}$ yields only a minor contribution.
In such cases, the optimization of the learning process can still reach its intended target with a substantially lower computational cost.
The gradients of the pressure solve resulting from $J^P$ have an even lower impact on the overall gradient accuracy, and our tests indicate that they are negligible in most scenarios.

\subsection{Operators for Turbulence Statistics} \label{sec:OnlineMoments}
In addition to the proposed solver, we also provide an implementation to compute differentiable, arbitrary-order (co)moments \cite{moments2016} in an online fashion. This allows us to accumulate statistics without the need to store entire simulation sequences, from which we can compute turbulence statistics or the turbulent energy budget terms
\begin{equation}
    \begin{split}
        \text{Production: } P_{ij} &= - \left( \overline{\velcomp_i'\velcomp_k'}\pder{\overline{\velcomp_j}}{x_k} 
            + \overline{\velcomp_j'\velcomp_k'}\pder{\overline{\velcomp_i}}{x_k} \right)\\
        \text{Dissipation: } \epsilon_{ij} &= 2 \overline{\pder{\velcomp_i'}{x_k} \pder{\velcomp_j'}{x_k}}  \\
         \text{Turbulent transport: } T_{ij} &= - \pder{\overline{\velcomp_i'\velcomp_j'\velcomp_k'}}{x_k}\\
        \text{Viscous diffusion: } D_{ij} &= \frac{\partial^2 \overline{\velcomp_i'\velcomp_j'}}{\partial x_k^2} \\
        \text{Velocity pressure-gradient term: } \Pi_{ij} &= - \left( \overline{\velcomp_i'\pder{\pressure}{x_j} + \velcomp_j'\pder{\pressure}{x_i}} \right),
    \end{split}
\end{equation}
where $\overline{\Box}$ represents averaging over time and homogeneous directions and summation over the vector components $k$ is implied.

\section{Optimization and Learning via Automatic Differentiation} \label{sec:learningloss}

The flexibility of our solver allows for the formulation of different initial and boundary condition optimization problems, as well as control tasks. In the present study, we focus our attention on developing \textit{correction} models~\cite{um2020solver,Kochkov} for highly under-resolved simulations.
This means a neural network modifies the state of a simulation in order to conform to a learning task.
In particular, our objective is to optimize deep neural networks $G(\cdot\,; \NNw)$ parameterized by $\NNw$ to output corrections that bring the simulation state closer to that of a higher-fidelity simulation.
The output can be either a residual correction to the instantaneous velocity, $\vel_\NNw := G(\cdot\,; \NNw)$ added to $\vel^n$ between simulation steps, or a correcting force added as additional source term $\source_\NNw := G(\cdot\,; \NNw)$ in \myrefeq{eq:momentum}. 
While both are viable options, we focus on the latter in the following as it more naturally couples with the PISO solver.

In deep learning tasks, the choice of the training loss $L$ plays an important role, not only to drive the non-linear optimization of the neural network towards the desired goal, but also to prevent sub-optimal results and training instabilities.
The network weights are optimized using gradient descent-based algorithms which utilize the gradients $\pderInl{L}{\NNw}$ to iteratively update $\NNw$. These gradients are obtained by backpropagation through the loss function and network and, in our case, also through the simulation rollout (see also \myrefsec{sec:diffdiscrete}).

\subsection{Loss Terms and Physical Constraints} \label{sec:learninglossPhysconstraint}

In the context of differentiable simulations, a training setup for a learned operator embedded in the solver should also ensure that constraints from the physical model are preserved.
In our incompressible flow scenario, the conservation of mass in the form of divergence-free motions is the most important constraint.
The corresponding pressure solve projects solutions onto the closest divergence free motion. As such, a single solution $\vel$ is obtained from all $\vel^* + \tilde{\vel}$ , where $\tilde{\vel}$ denotes an arbitrary irrotational velocity field.
The ambiguity of this surjective mapping can impede learning tasks that aim to provide or correct $\vel$ via a learned operator $\vel_\NNw := G(\cdot;\NNw)$, as substantially different outputs of the neural network can yield identical results within the solver.
Hence, the learning process itself can start to oscillate around different solutions once it has reached sufficient accuracy.
This is similar to nullspace issues of classic iterative solvers~\cite{strangBook},
where this problem can prevent an iterative solver from converging. Interestingly, this issue is not directly solved in a differentiable PISO solver: the differentiable pressure projection operator is completely agnostic to divergent parts, and hence provides no learning direction with respect to different divergent solutions.
We also observed that divergent solutions can cause issues within the simulation when correcting the velocity directly. Specifically, when divergence is introduced before the predictor step, the advection may introduce oscillations in the velocity that are not recovered by the following pressure projection.
While divergent source terms as an alternative to velocity corrections do not cause these oscillations, the nullspace issues still apply.

Since the particular choice of $\tilde{\vel}$ or $\tilde{\source}$ is irrelevant to the final solution, 
we consider different approaches to guide the optimization. The simplest way to address this issue is to use classic techniques for regularization. In particular, we consider weight decay~\cite{goodfellow2016deep} as additional term in our loss function:
\begin{equation} \label{eq:wdloss}
L_{\text{WD}} = \lambda_{\text{WD}} ||\NNw||^2.
\end{equation}
This loss term favors solutions with small magnitudes of the NN parameters, and in this way yields a better posed learning problem with a smaller number of solutions. 
It should be noted, however, that this approach typically yields reduced output magnitudes for the network, and can lead to overly penalizing favorable solutions that require larger network weights.

Alternatively, we may seek to minimize $\tilde{\vel}$ and thus to stabilize training by preventing the network from learning outputs that induce non-divergence-free motions~\cite{agdestein2025discretize}.
A naive option consists in using the divergence of the velocity as a soft constraint of the form $||\nabla \cdot \vel_\NNw||^2$. 
While this could drive the network towards divergence-free output, it is a local feedback for a global problem and thus could take many iterations to converge.
Additionally, in our setting the central differencing used to compute the divergence would be prone to cause checkerboard artifacts~\cite{hopman2025quantifying}.
A more principled alternative is to compute an additional pressure correction $\nabla^2\pressure_\NNw = \nabla \cdot \vel_\NNw$. The spatial pressure gradient $\nabla \pressure_\NNw$ is then the exact, globally correct feedback that drives the velocity towards fulfilling the continuity equation, as $\vel_\NNw - \nabla \pressure_\NNw$ represents a divergence free motion.
To integrate this feedback we directly change the gradient $\pderInl{L}{\NNw}$ to be computed as 
\begin{equation} \label{eq:SgradPsgrad}
    \pder{L}{\NNw} = \pder{\vel_\NNw}{\NNw}\left( \pder{L}{\vel_\NNw} + \lambda_{\nabla\cdot\vel} \nabla \pressure_\NNw \right),
\end{equation}
where the scaling factor $\lambda_{\nabla\cdot\vel}$ is used for balancing the loss terms.
This gradient modification directly transfers to modifications of $\source_\NNw$ instead of $\vel_\NNw$.

\subsection{Losses from Turbulence Statistics}

For highly chaotic flows, the pairs of high and low resolution simulation frames needed for supervised training no longer match 
as soon as simulations at different resolutions produce de-correlated trajectories.
Down-sampling high resolution frames likewise does not produce valid low resolution statistics.
To circumvent this issue we follow~\cite{turbulenceMARL2021,list2022_Learned} and include more physical guidance into the training by using the differentiable turbulence statistics from \myrefsec{sec:OnlineMoments}, which allow us to define statistics losses with respect to~those of a reference simulation $\hat{\velcomp}$.
For wall-bounded turbulence simulations, like turbulent channel flows, the loss terms for mean and second order statistics are
\begin{equation}
    \begin{split}
        L_{U_i}^{n:m} &= \frac{1}{Y} \sum_{y=0}^{Y-1} ||\overline{\velcomp_i}^{n:m}(y) - \overline{\hat{\velcomp}_i}(y) ||_2^2\\
        L_{u'_{ij}}^{n:m} &= \frac{1}{Y} \sum_{y=0}^{Y-1} ||\overline{\velcomp'_i \velcomp'_j}^{n:m}(y) - \overline{\hat{\velcomp}'_i \hat{\velcomp}'_j}(y)||_2^2,\\
    \end{split}
\end{equation}
where $Y$ is the resolution in wall normal direction and $\overline{\Box}^{n:m}$ denotes averaging over homogeneous directions and rolled out steps $n$ to $m$.
The loss terms for statistics are combined into a loss formulation that contains both temporally averaged statistics and per-frame statistics:
\begin{equation} \label{eq:TCFstatsLossGeneric}
    L_{\text{stats}} = \sum_{i=0}^2 \lambda_{U_i} L_{U_i}^{0:N} + \sum_{i=0}^2 \sum_{j=0}^2 \lambda_{u'_{ij}} L_{u'_{ij}}^{0:N}
    + \sum_{n=0}^{N} \lambda^n_{stats}\left( \sum_{i=0}^2 \lambda_{U_i} L_{U_i}^n + \sum_{i=0}^2 \sum_{j=0}^2 \lambda_{u'_{ij}} L_{u'_{ij}}^n \right),
\end{equation}
with $N$ being the number of rolled out steps.
The inclusion of per-frame turbulence statistics helps to prevent a learned operator from compensating undershooting a particular statistic with a delayed overshoot.
This leads to a more consistent matching of the statistics with fewer temporal oscillations.

\section{Validation} \label{sec:validation}

\begin{figure}
    \centering
    \begin{subfigure}{0.49\textwidth}
        \includegraphics[width=\textwidth]{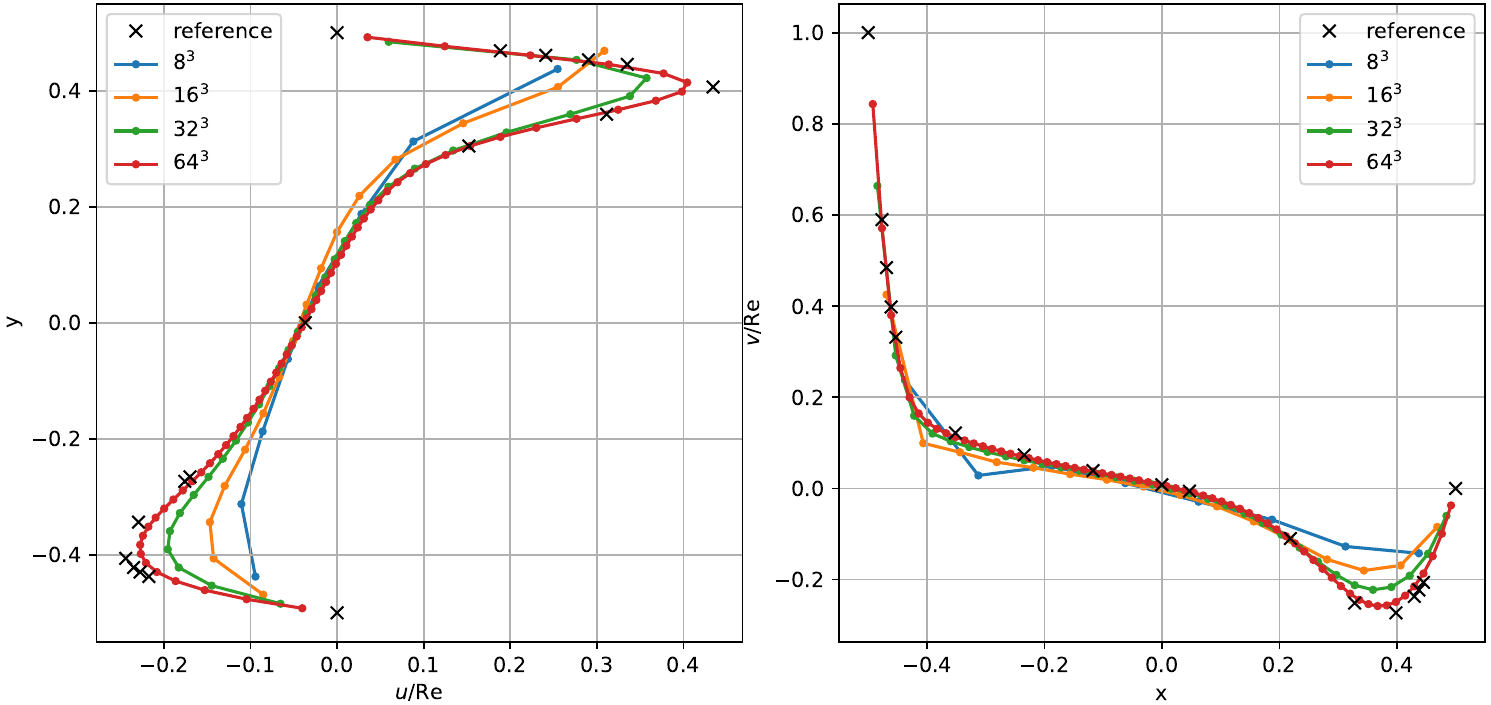}
        \caption{Uniform grid} \label{fig:Lid3d1000:uniform}
    \end{subfigure}
    \hfill
    \begin{subfigure}{0.49\textwidth}
        \includegraphics[width=\textwidth]{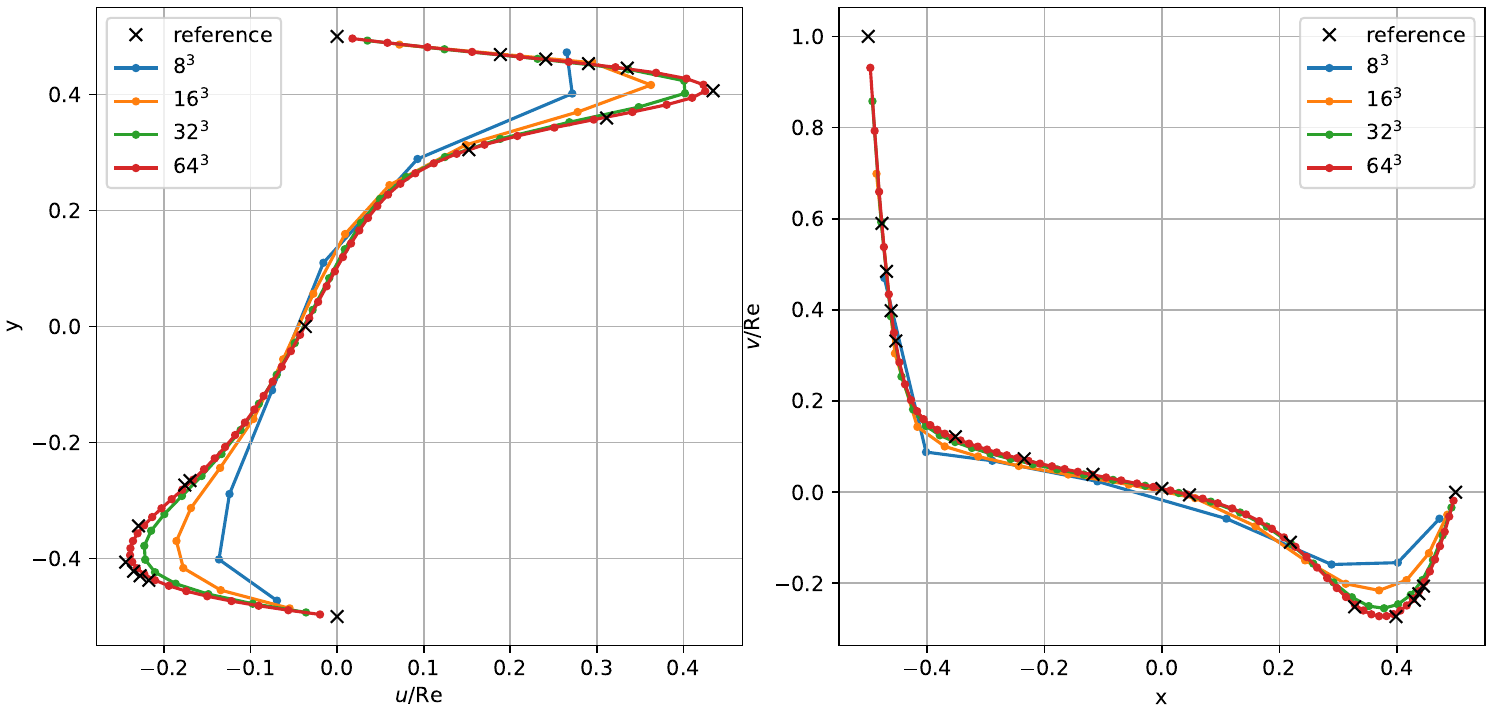}
        \caption{Refined grid} \label{fig:Lid3d1000:refined}
    \end{subfigure}
    \caption{Velocity profiles for the 3D lid-driven cavity with $\text{Re}=1000$ for increasing resolutions.
    The left image in each subfigure is the $\velu$-velocity on the vertical center line, and the right is the $\velv$-velocity on the horizontal center line.
    (b) uses a grid that was refined towards all boundaries.
    The velocities are normalized with the Reynolds number.
    The reference is a high-resolution DNS~\cite{Lid3D}.
    } \label{fig:Lid3d1000}
\end{figure}

\begin{figure}
    \centering
    \includegraphics[width=\textwidth,trim={0.1in 0.15in 0.1in 0.1in},clip]{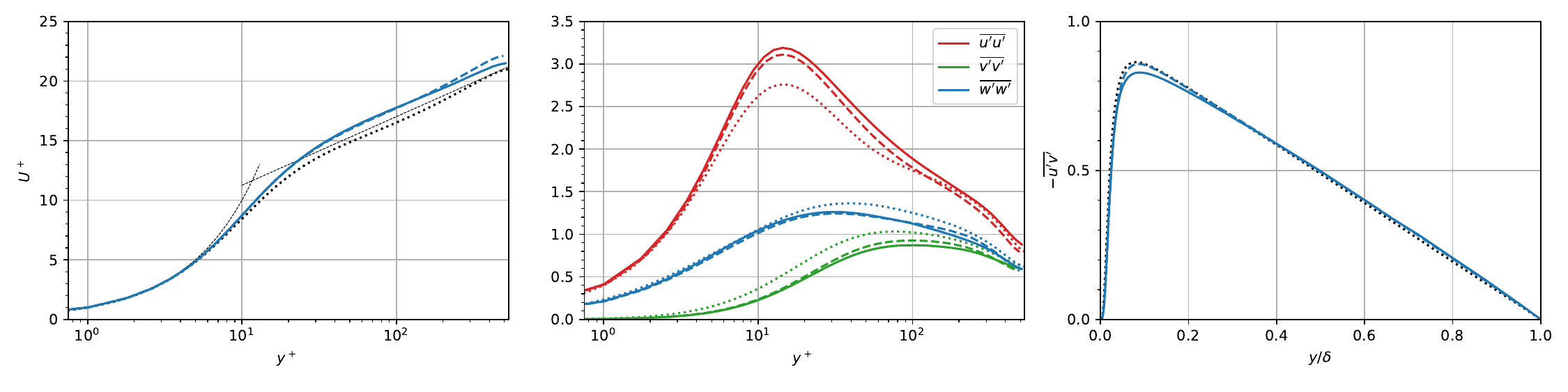}
    \caption{
    Turbulence statistics for a 3D TCF with $\Rewall = 550$.
    The statistics are averaged over time and stream- and span-wise direction,
    normalized with the average $\velwall$ of the corresponding simulation,
    and plotted against the wall-normal direction.
    The solid lines show results from our solver, while the dashed lines show those of OpenFOAM's PISO implementation using the same computational mesh.
    Dotted lines indicate the spectral reference from Hoyas and Jim\'enez \cite{TCF_2008_10}, while
    fine dotted lines in the $U^+$ plot are the log-law and law of the wall.
    } \label{fig:TCF180-TCF550}
\end{figure}
\begin{figure}
    \centering
    \includegraphics[width=\textwidth]{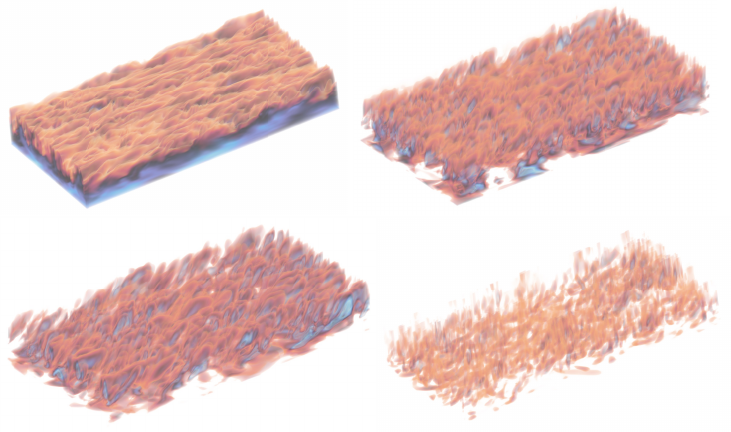}
    \caption{
    Qualitative visualizations of the pressure and velocity components of our TCF benchmark. Only the flow close to the bottom wall in the computational space is shown.
    The color gradient and iso-surface values are chosen individually to highlight the detail of the flow structures.
    Top: $\velu$ and $\velv$, bottom: $\velw$ and $\pressure$.
    } \label{fig:TCFvis}
\end{figure}

In this section, we first give a  brief overview of benchmark scenarios that were used to validate the forward simulations, focusing on 3D cases. Details and additional validations are provided in \myrefapp{app:validation}.
We then validate the gradient derivations and investigate them in optimization settings.
While PICT supports single and double precision, all experiments were run with single precision, matching the standard in learning applications, and conforming to recent analysis from turbulence modeling~\cite{karp2025effectslowerfloatingpointprecision}.

\subsection{Forward Simulation} \label{sec:valFWD}

For a lid-driven cavity setup, the plots in \myreffig{fig:Lid3d1000} show that the PICT solver correctly converges to the reference solution obtained from a high-resolution DNS~\cite{Lid3D}. Grid refinement, shown on the right of \myreffig{fig:Lid3d1000}, further improves the results.
We also simulate a 3D turbulent channel flow (TCF)
at $\Rewall = 550$
and compare to established numerical references~\cite{TCF_2008_10} and solvers~\cite{weller1998tensorial, OpenFOAM}.
The resulting turbulence statistics, accumulated over 20 \ETT{} (eddy turnover time) after convergence of the simulation, can be found in \myreffig{fig:TCF180-TCF550}.
The flows are statistically stationary and the averaged, inner-scaled statistics 
of PICT closely match those obtained with OpenFOAM's PISO implementation using the same mesh.
Both solvers are also reasonably close to the spectral reference, despite the relatively low resolution. The resulting averaged $\Rewall$ of 534 is also very close to the target.
A qualitative visualization of the simulated boundary layer in terms of velocity components and pressure is shown in \myreffig{fig:TCFvis}. The visualizations highlight the anisotropy of the velocity fields due to the presence of the wall and the highly directional nature of the flow, with the velocity streaks oriented in the streamwise direction.

\subsection{Gradients}
\label{sec:gradients}

As differentiability is a key feature of the PICT solver, the individual gradient operations 
are validated numerically using PyTorch's \texttt{gradcheck} tool \cite{TorchGradcheck}, which compares the analytic gradients provided by our custom operations to numerical approximations of the gradients via finite differences.
The tests confirm accuracy of the custom gradient operations up to numerical precision.
Equipped with gradients for the building blocks of the simulator, verifying the capabilities of direct optimization tasks is a natural next step to verify the correctness of gradients backpropagated through the full solver.
This task does not involve a neural network, but treats parameters of the simulation as degrees of freedom to be optimized.

\begin{figure}
    \centering
    \begin{subfigure}{0.245\textwidth}
        \includegraphics[width=\textwidth]{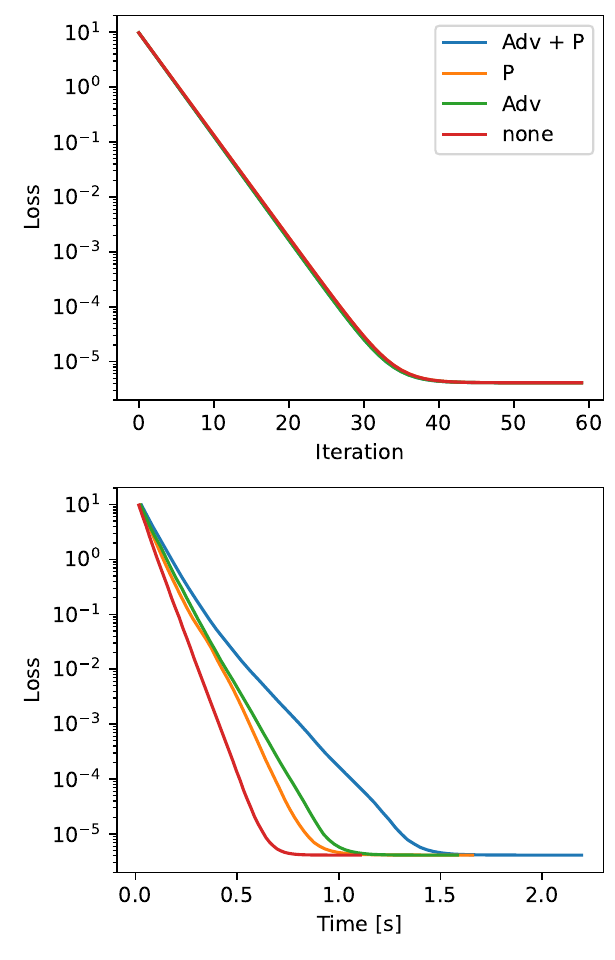}
        \caption{$n = 1$, lr $=0.01$} \label{fig:gradpathu:n1}
    \end{subfigure}
    \hfill
    \begin{subfigure}{0.245\textwidth}
        \includegraphics[width=\textwidth]{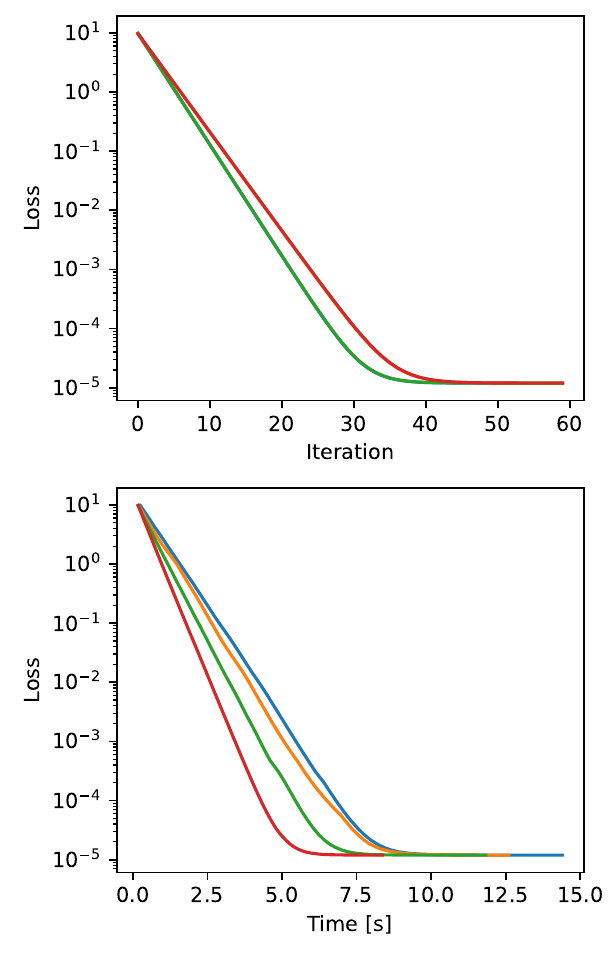}
        \caption{$n = 10$, lr $=0.01$} \label{fig:gradpathu:n10}
    \end{subfigure}
    \hfill
    \begin{subfigure}{0.245\textwidth}
        \includegraphics[width=\textwidth]{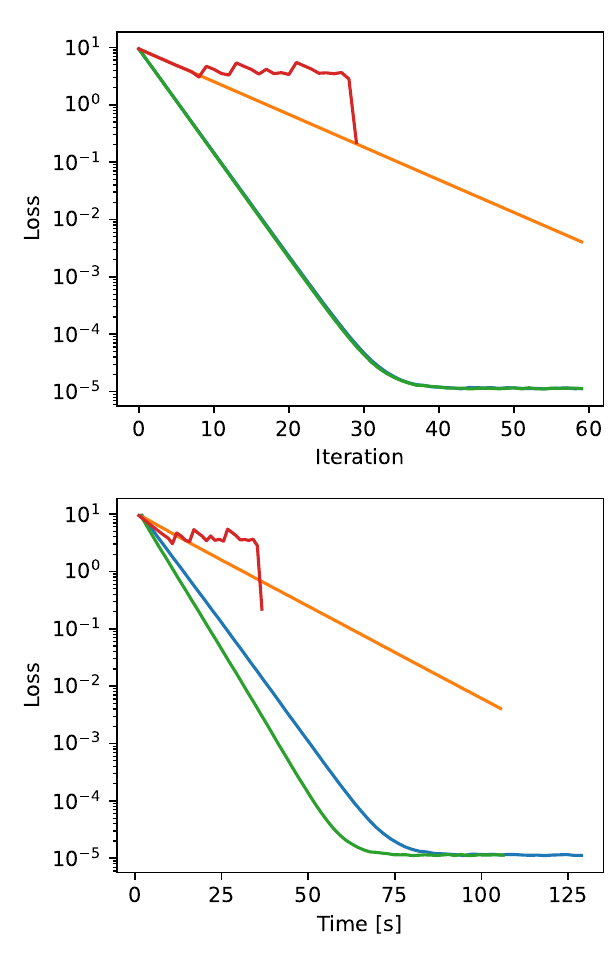}
        \caption{$n = 100$, lr $=0.01$} \label{fig:gradpathu:n100}
    \end{subfigure}
    \hfill
    \begin{subfigure}{0.245\textwidth}
        \includegraphics[width=\textwidth]{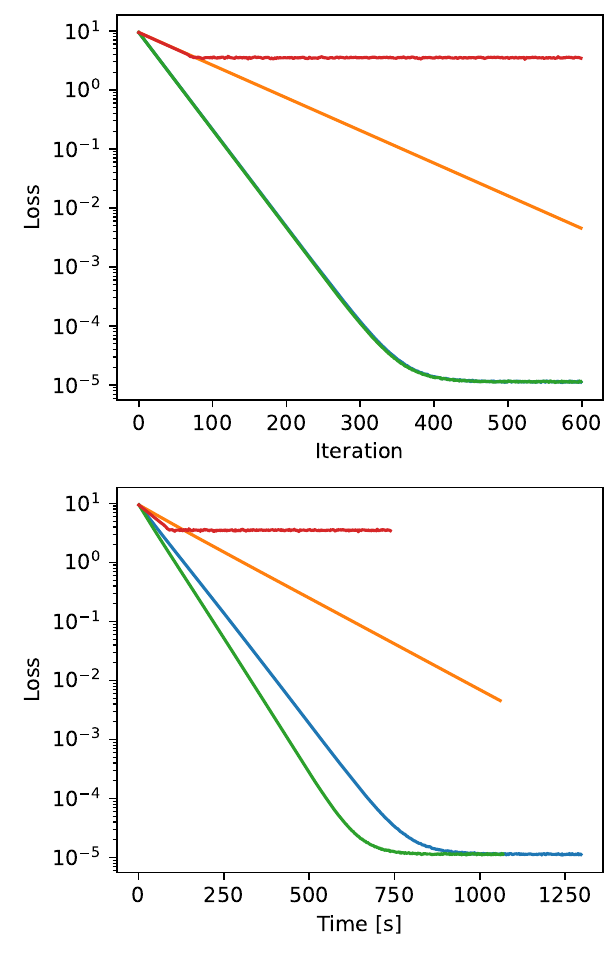}
        \caption{$n = 100$, lr $=0.001$} \label{fig:gradpathu:n100slow}
    \end{subfigure}
    \caption{Loss curves for the optimization task and gradient path ablations (top row), and corresponding runtimes (bottom row). 
    \quoteEN{Adv+P} denotes the full version, while 
    \quoteEN{Adv} indicates that only the gradient of the advection-diffusion linear solve is used, \quoteEN{P} only that of the pressure linear solve. The \quoteEN{none} version relies on a gradient path without both linear solves.
    The columns show results with different rollout length $n$.     
    The step size (equivalent to the learning rate) is additionally reduced from $0.01$ to $0.001$ for the longest rollout.
    } \label{fig:gradpathu}
\end{figure}

The simulation setup used is a periodic box of resolution $18\times16$ where we initialize the $\velu$-velocity with a 2D gauss profile. The objective is to optimize for a single degree of freedom that represents an unknown scaling of this initial velocity based on a L2 velocity-loss that is computed after $n$ simulation steps. 
The optimization was performed via gradient descent iterations, the convergence for which is shown in \myreffig{fig:gradpathu}.
The full solver, denoted by \textit{Adv+P} in this graph successfully converges towards the reference factor up to numerical precision, as indicated by the loss approaching a level of below $10^{-5}$.
This scenario shows that the differentiable solver provides correct gradients, 
which we also verify with an additional lid driven cavity optimization presented in the appendix.
Next, we use the task of \myreffig{fig:gradpathu} to investigate the different options of backpropagation paths through the solver, as outlined in \myrefsec{sec:gradpaths}. 

\subsection{Gradient Path Ablation}
\label{sec:gradPathAblation}

Computationally, the most expensive parts of the simulation in terms are the linear solves for the advection-diffusion and pressure systems, which typically take 70-90\% of the total runtime. While they are critical for the forward simulation, 
we investigated their influence on the gradient calculation, 
as alternative paths in the computational graph exist,  cf. \myreffig{fig:PISOgraph} and \myrefsec{sec:gradpaths}.
The versions we consider are \textit{Adv},
where the gradients $\pder{\vel^{\guess}}{\AdvMat}$ and $\pder{\vel^{\guess}}{\velRHS}$ of the advection solve are included,
and \textit{P}, utilizing $\pder{\pressure}{\PMat}$ and $\pder{\pressure}{\nabla \cdot \pressureRHS}$.
We then optimize the unknown scaling parameter for the initial velocity through a varying number of steps $n$ by computing the gradient $\pderInl{\vel^{n}}{\vel^{0}}$ via  backpropagation through  all $n$ steps of the solver.
For shorter lengths of $n=1$ and 10, the difference is non-existent or negligible. However, when plotted against the wall clock time, there is a significant performance advantage in skipping the linear solves in the backwards pass, with a speed up of up to 2x when skipping both, as shown in \myreffig{fig:gradpathu}.
For larger $n$, on the other hand, the advection becomes more relevant, and when skipping both solves in the backwards pass the optimization can start to diverge. This happens in our setup when no linear solve gradients are used with $n=100$ steps and the default learning rate of $10^{-2}$.
A significantly lower learning rate of $10^{-3}$, as used for the rightmost result shown in the figure, can compensate for this effect to some extent, but the optimization nonetheless converges to a sub-optimal minimum with a higher loss. 
In the 100 step case, the versions without the advection-diffusion solve gradients also no longer have a convergence-runtime advantage over the versions with these terms. Only the pressure solve gradient can be neglected, which still results in a faster convergence in the runtime comparisons, reaching the same loss in the same number of optimization steps. 

A runtime comparison of the different versions can be found in \myreftab{tab:gradpathtimes}.
The lower learning rate also means that more optimization steps are required to reach a certain level of accuracy, indicating that this extra time might be better spent on including the backwards pass linear solves in combination with a higher learning rate.
Overall, the parameter $n$ represents a key hyperparameter that controls the \textit{unrolling} length when computing losses for training neural networks~\cite{list2025differentiability}.
Hence, this ablation provides that an interesting tradeoff between accuracy and runtime performance based on unrolling length, learning rate, and desired convergence. For small and moderate $n$ in comparison to the representative chaotic timescale of the system,
it is a viable option to exclude the linear solves' gradients in the initial training. Once training has converged to a steady state, the paths can be re-activated to achieve a more accurate results as a \textit{fine tuning} phase.
In the following, we will explicitly state if approximate gradients with excluded paths in the computational graph were used.

\begin{table}
    \centering
    \begin{tabularx}{\textwidth}{l X X X X} \hline
        \       & $n=$1 & $n=$10 & $n=$100 & $n=$100, $lr=0.001$ \\\hline
        Adv + P & 1.084 & 6.853 & 63.20 & 674.3 \\
        P       & 0.689 & 6.707 & 157.5 & 1611 \\
        Adv     & 0.775 & 5.479 & \textbf{52.11} & \textbf{552.1} \\
        none    & \textbf{0.515} & \textbf{4.393} & - & - \\\hline
    \end{tabularx}
    \caption{Wall clock time to reach loss $< 10^{-4}$ in seconds.
    All experiments run for 60 optimization steps with a learning rate of $lr=0.01$, except for $n=100$ which used 600 steps with $lr=0.001$.}
    \label{tab:gradpathtimes}
\end{table}

\section{Results - Deep Learning Applications} \label{sec:DLresults}

In this section we show the efficacy of our differentiable solver by developing learned correctors and SGS models to improve coarse simulations in various flow scenarios.
As described in \myrefsec{sec:learningloss}, these models are neural networks $G(\cdot;\NNw)$ that are tasked to estimate a correcting force $\source_\NNw$. We target two 2D scenarios, a vortex street and backward facing step, in addition to a turbulent channel flow in 3D. The cases highlight PICT's support for different learning formulations, matching both instantaneous flow fields and statistics, as well as non-uniform discretizations, as they employ grid refinement near obstacles and walls.

\subsection{2D Vortex Street}

The vortex street case, typically associated with flow past a cylinder or square, is a classical and well-understood problem in fluid mechanics. It serves as a canonical benchmark for validating numerical solvers \cite{braza1986numerical, jackson1987finite} and machine learning models \cite{raissi2019physics, sharifi2023spatial, um2020solver, drygala2024comparison}. The vortex street phenomenon  introduces non-linear, unsteady flow characteristics that are challenging to model. As such, it provides an ideal starting point to test our solver’s performance when coupled with machine learning models.
We target a correction setup, where $G(\cdot;\NNw)$ has the task to let an approximate low-resolution simulation match a high-resolution reference by correcting the source term of the PISO solver. 
As neural network architecture, this scenario used a 2D CNN with 7 layers and 16, 32, 64, 64, 64, 64, and 2 filters, respectively. The kernel sizes of the filters are \(7^2\), \(5^2\), \(5^2\), \(3^2\), \(3^2\), \(1^2\), and \(1^2\), respectively, for a total of 144750 parameters. The stride is 1 in all layers and we use ReLU as the activation function. To handle correct padding between blocks, this network (and subsequent ones) use PICT's custom multiblock convolutions from \myrefsec{sec:MBGtransform}.

In this work, we adopt a simulation setup which employs a square bluff body to generate the vortex street~\cite{kelkar1992numerical}; further details of the setup are provided in \myrefapp{app:2d_vs}. Using a square obstacle instead of a circular one is motivated by the need to rigorously evaluate our method's capacity to address pronounced numerical instabilities and complex flow patterns. The sharp corners of a square obstacle introduce significant computational challenges, particularly as the grid resolution decreases. In low-resolution simulations, the sharp edges of the square create corner-induced disturbances that propagate upstream as checkerboard patterns, which indicate numerical artifacts and oscillations. Such instabilities can degrade simulation accuracy and lead to divergence if left uncorrected. Thereby it provides a stringent test of our solver's stability and the neural network's corrective capabilities.

The vortex shedding behavior transitions through distinct regimes depending on the Reynolds number. In this study, we focus on two cases with $Re=500$ and $Re=600$, where the flow already exhibits chaotic behavior~\cite{saha2000transition}. To introduce variability between training and test sets, the obstacle height $y_s$ and $Re$ are varied. A detailed breakdown of the geometry, Reynolds number, and sample ranges for each set is provided in \myreftab{tab:train_test_Karman}. 
For training and inference, we target a version down-sampled by 4$\times$ in the two spatial directions, giving a resolution of $67 \times 36$, and a 10$\times$ larger timestep.
We use an adaptive time stepping method to ensure \CFL $\le 0.8$ for all low- and high-resolution cases.
Since the grids are not uniformly distributed with refinement, a coordinate-based approach that interpolates velocity values between the high- and low-resolution grids has been used to downsample high-resolution data. We employ a curriculum-based training strategy~\cite{um2020solver}, starting with 4 unrolled steps and progressively increasing to 8 and finally 16 steps.
The corrector models trained with 8 and 16 unrolled steps are referred to as $\NNeight$ and $\NNsixteen$, respectively.
The training steps of $\NNeight$ are extended to match $\NNsixteen$ to ensure that the only variable differentiating $\NNeight$ and $\NNsixteen$ is the final number of unrolled steps.
While this system is chaotic, the scale separation of flow features is much smaller than it would be in 3D turbulent flows, allowing us to train the NN corrector with a supervised strategy in line with the rich literature of learning for 2D flows~\cite{List_Chen_Thuerey_2022, kochkov2021machine, um2020solver, freitas2024solverintheloopapproachturbulenceclosure}.
Thus, for training we use a simple MSE loss between the instantaneous prediction and target velocity field, which is evaluated at every other time step of the time integration.
The ground-truth trajectory remains a valid and meaningful target for the used rollout lengths, as the model learns to correct the error at each simulation step and thus ensures that the simulation stays close to the true dynamics.
We further apply the weight decay defined in\myrefeq{eq:wdloss} to stabilize the training, implemented as L2 regularization in the standard Adam optimizer~\cite{kingma2014adam}.
For comparison, we also simulate numerical solutions without a neural network at the same resolution, denoted as $\Lowres$, which serves as a baseline for comparison against the trained models $\NNeight$ and $\NNsixteen$.

\begin{table}
    \centering
    \begin{tabularx}{\textwidth}{l X X X X X}
    \hline
    Dataset                & No. & $y_s$      & Re  & Sample range  & Unrolled steps \\ \hline
    \multirow{4}{*}{Train} & 1   & 1.0               & 500 & $60 \sim 100$  & 8/16 \\
                           & 2   & 1.5               & 500 & $60 \sim 100$  & 8/16 \\
                           & 3   & 1.0               & 600 & $60 \sim 100$  & 8/16 \\
                           & 4   & 1.5               & 600 & $60 \sim 100$  & 8/16 \\ \hline
    Test                   & 5   & 2.0               & 600 & $70 \sim 100$  & 2000 \\ \hline
    \end{tabularx}
    \caption{
    Training and test setup details for the vortex street corrector.
    Obstacle height $y_{s}$, Reynolds number, sample range for initial state, and unrolled steps.}
    \label{tab:train_test_Karman}
\end{table}

\begin{figure*}[ht]
    \centering
    \includegraphics[height=0.5\textwidth]{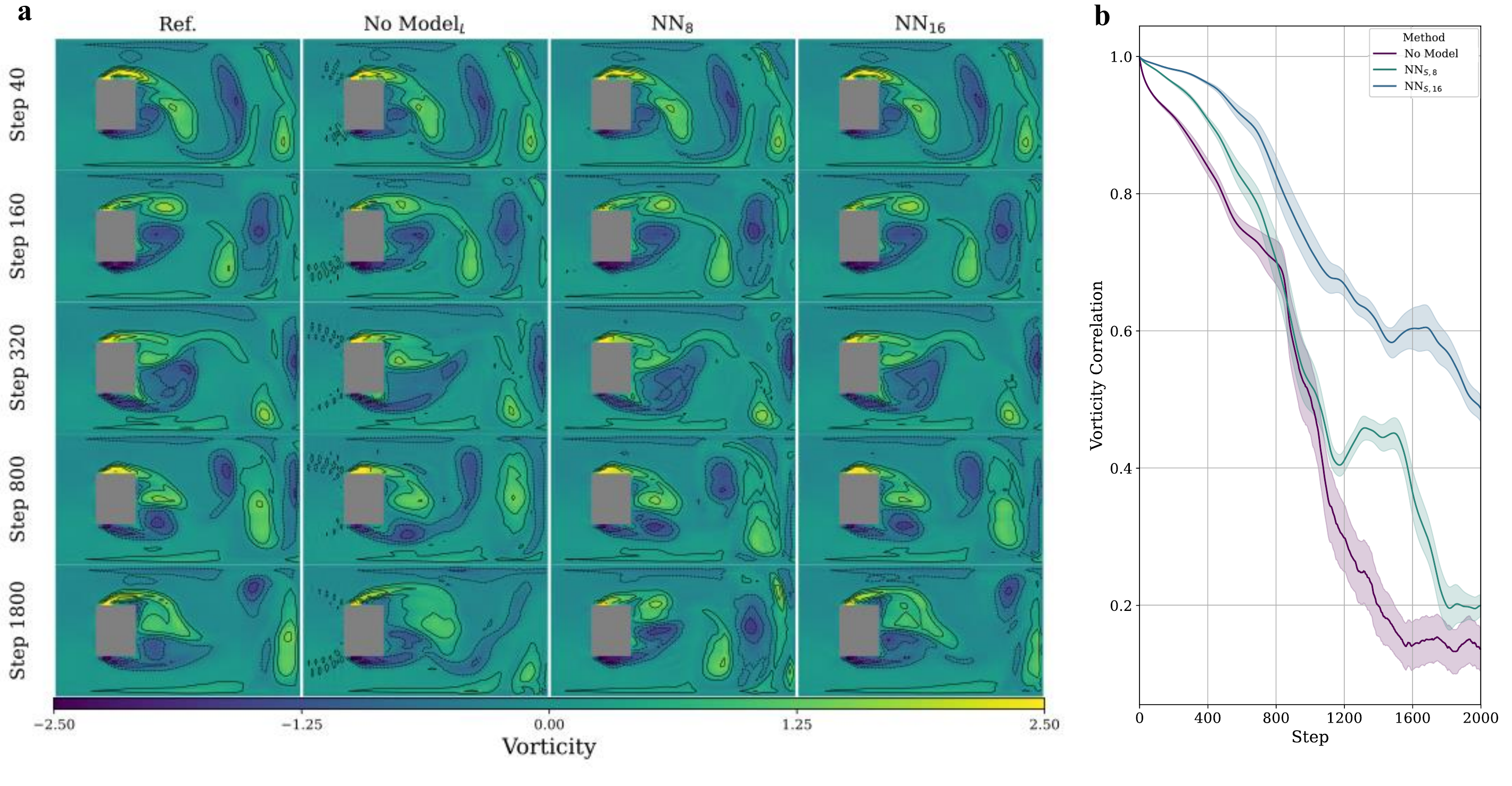}
    \caption{(a) Comparison of vorticity fields at different simulation steps (40, 160, 320, 800, 1800) of the reference solution (Ref.), baseline ($\Lowres$), and our models ($\NNeight$ and $\NNsixteen$). (b) Temporal evolution of vorticity correlation for all test cases, with shaded regions representing scaled standard deviation. Higher vorticity correlation indicates a better alignment with the reference solution.}
    \label{fig:karman_vorticity_compare}
\end{figure*}

\begin{table}[ht]
    \centering
    \renewcommand{\arraystretch}{1.1}
    \setlength{\tabcolsep}{5pt}      
    \small 
    \begin{tabularx}{\textwidth}{l c c c c c c}
    \hline
    \multirow{2}{*}{Method} & \multicolumn{2}{c}{Step = 120} & \multicolumn{2}{c}{Step = 480} & \multicolumn{2}{c}{Step = 2000} \\
                            & $C$ & MSE ($\times 10^{-4}$)                    & $C$ & MSE ($\times 10^{-3}$)                       & $C$ & MSE ($\times 10^{-1}$)                      \\ \hline
    $\Lowres$   & 0.933 $\pm$ 0.010  & $13.674 \pm 3.915$  & 0.805 $\pm$ 0.061  & $17.230 \pm 6.941$  &  0.136 $\pm$ 0.157  & 17.285 $\pm$ 61.45  \\
    $\NNeight$   & 0.976 $\pm$ 0.004  & $7.715 \pm 1.841$  & 0.879 $\pm$ 0.046  & $10.152 \pm 4.531$  & 0.199 $\pm$ 0.080  & $1.424\pm 0.220$  \\
    $\NNsixteen$  & 0.987 $\pm$ 0.004  & $3.801 \pm 1.140$  & 0.947 $\pm$ 0.023  & $3.599 \pm 1.799$  & 0.488 $\pm$ 0.095  & $0.762 \pm 0.216$  \\ \hline
    \end{tabularx}
    \caption{Performance comparison of vorticity correlation and MSE at different forward steps. Higher vorticity correlation and lower MSE indicate better performance. The values represent the mean $\pm$ standard deviation.
    }
    \label{tab:vorticity_mse}
\end{table}

Since the stability of long-term forward simulations is inherently sensitive to initial conditions~\cite{shankar2024differentiable}, we conducted a comprehensive assessment of the stability and accuracy improvements introduced by our methods. Initial states were sampled within a wide time range of $[70, 100]$, with an interval of $0.5$, resulting in a test set of 60 initial conditions. Each of them is advanced by $100$ in time, amounting to 2000 steps of the low-resolution simulator, which is far beyond the number of unrolled steps during training. 
In \myreffig{fig:karman_vorticity_compare}(a) we present a qualitative comparison of the time evolution of the different models.
The baseline, $\Lowres$, exhibits persistent checkerboard oscillations, particularly around the sharp corners of the obstacle. These artifacts remain present throughout the simulation, and the mismatch between the baselines and the reference solution grows over time. $\Lowres$ shows a fast growth of non-physical deterioration, with severe distortions in the vorticity field becoming evident by step 800. 
In contrast, both $\NNeight$ and $\NNsixteen$ suppress non-physical oscillations effectively, with no checkerboard patterns visible in the vorticity fields. The $\NNsixteen$ model, by comparison, achieves superior performance. By step 1800, $\NNsixteen$ produces a vorticity field nearly indistinguishable from the reference, showing its robustness and generalization capabilities for long-term simulations. 

This qualitative comparison provides an initial insight into the models' effectiveness, which is further supported by quantitative evaluations, provided in \myreffig{fig:karman_vorticity_compare}(b) and \myreftab{tab:vorticity_mse}. Herein, the vorticity correlation is calculated as
\begin{equation}
    C(t) = \frac{\sum_{i,j} \omega_{i,j}(t) \cdot \hat{\omega}_{i,j}(t)}{\sqrt{\sum_{i,j} \left(\omega_{i,j}(t)\right)^2} \cdot \sqrt{\sum_{i,j} \left(\hat{\omega}_{i,j}(t)\right)^2}},
\end{equation}
where $\omega_{i,j}(t)$ is the vorticity of the results and $\hat{\omega}_{i,j}(t)$ that of the reference at grid point $(i, j)$ and time $t$.
The baseline shows a rapid decay in vorticity correlation with the reference solution within the first 100 steps. By step 2000, its correlation drops to a mean of 0.136, with significant variability across initial conditions.
In contrast, both $\NNeight$ and $\NNsixteen$ perform significantly better in the initial 300 steps. However, beyond step 400, $\NNeight$ experiences a decline in correlation. These deviations indicate that while $\NNeight$ suppresses the numerical instability, it still suffers from a frequency decay during long-term forward simulations. This is caused by the limited number of unrolled steps in training, which are insufficient to fully capture the non-linear dynamics \cite{list2025differentiability}. $\NNsixteen$ demonstrates the best overall performance. It consistently maintains higher vorticity correlation throughout the simulation, achieving strong alignment with the reference solution, significantly outperforming the baselines and $\NNeight$. This improvement highlights the importance and benefits of training via unrolling, as inherently supported by the PICT solver. It enables the model to better capture and mitigate long-term error accumulation.
To further evaluate the performance of $\NNsixteen$, we analyze the MSE of the aerodynamic coefficients (lift and drag), by comparing time-averaged values from the high-fidelity reference against both the no-model baseline and the $\NNsixteen$ predictions. The baseline is numerically unstable, yielding a drag MSE of $2.14$. In contrast, our model ($\NNsixteen$) successfully stabilizes the simulation, reducing the drag MSE by three orders of magnitude to $4.86 \times 10^{-3}$. A clear improvement is also seen for the lift coefficient, where the MSE is reduced from $0.936$ in the unstable baseline to $1.29 \times 10^{-3}$. This demonstrates the model's success in capturing the fundamental flow dynamics where the baseline solver fails.

\subsection{Backward Facing Step}
Building on the insights gained from the vortex street, we target a backward-facing step (BFS) scenario, a classic separated flow produced by an abrupt change in geometry~\cite{kuehn1980effects, armaly1983experimental, durst1983flows, erturk2008numerical}. 
Unlike the vortex street, where periodic shedding dominates, the BFS introduces new challenges by emphasizing spatial development over long temporal evolutions. The flow evolves spatially along the downstream region, necessitating long-term accuracy and stability to reproduce the statistics of the flow consistently. The focus shifts from periodic shedding to maintaining correct statistical properties over an extended domain. This case tests the solver’s ability to handle flows characterized by separation and reattachment dynamics. 
Notably, while the target is to achieve accurate statistical properties, the training process remains identical to the setup of the vortex street case, which is centered on predicting the instantaneous velocity field, rather than directly targeting statistical metrics.
This approach emphasizes the solver's ability to sustain physical accuracy while inherently preserving statistical consistency.

The geometry of the domain contains a 
gap between the step and the top wall of $h=1$, and a total channel height of $H=5h$, as described in \myrefapp{app:2d_bfs}. In the following,
$U_{\text{b}}$ denotes the bulk velocity, and $\nu$ the kinematic viscosity. The expansion ratio ($ER=H/h$) and the Reynolds number ($Re=2h U_b/\nu$) are considered as two of the most effective factors that influence metrics like reattachment length for BFS~\cite{kuehn1980effects, armaly1983experimental}.
The train and test sets are created by varying the ER and Re as detailed in \myreftab{tab:train_test_BFS}. 
In line with the vortex street case, we target a corrector learning setup, where the model should correct a low resolution simulation
with 4$\times$ spatial downsampling and 4$\times$ temporal downsampling to match the high-resolution reference. We use the same NN architecture as before, and a curriculum-based training approach starting with 10 unrolled steps. Training continues with 30 steps, and concludes with unrolling 40 steps. The resulting models are denoted as $\NNthirty$ and $\NNforty$. 

\begin{table}[!h]
    \centering
    \begin{tabularx}{\textwidth}{l X X X X c c}
    \hline
    \textbf{Dataset}       & \textbf{No.} & \textbf{\boldmath$s/h$} & \textbf{ER} & \textbf{Re} & \textbf{Sample range } & \textbf{Unrolled steps} \\ \hline
    \multirow{3}{*}{\textbf{Train}} & 1            & 0.875 & 1.875    & 1300         & $300 \sim 340$                        & 30/40 \\
                           & 2            & 0.875 & 1.875    & 1350         & $300 \sim 340$                        & 30/40 \\
                           & 3            & 0.85  & 1.85     & 1350         & $300 \sim 340$                        & 30/40 \\ \hline
    \textbf{Test}          & 4            & 1.0   & 2.0      & 1400         & $300 \sim 450$                        & 6000 \\ \hline
    \end{tabularx}
    \caption{Details for train and test sets in the BFS scenario: normalized step height $s/h$, expansion ratio ER, Reynolds number, sample range for initial state, number of unrolled steps.}
    \label{tab:train_test_BFS}
\end{table}

\begin{figure*}[ht]
    \centering
    \includegraphics[height=0.37\textwidth]{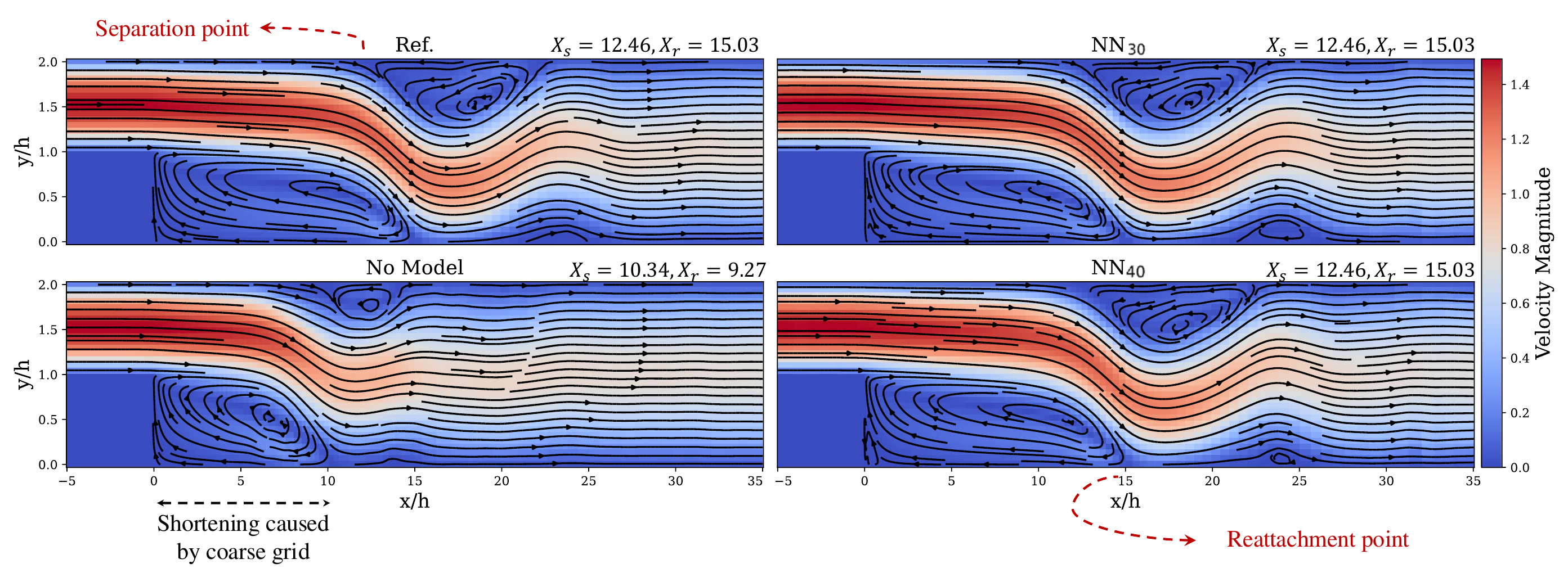}
    \caption{Stream plot, averaged over 6000 $\dt$ across reference, baseline(No model), $\NNthirty$ and $\NNforty$, the contour map visualizes velocity magnitudes $\|\vel\|$.}
    \label{fig:BFS_stream}
\end{figure*}

Figure \ref{fig:BFS_stream} presents the streamlines illustrating the flow patterns, averaged over $ t =6000 \dt$ (equivalent to $t U_b/h=120$) for all cases, including the reference solution, the $\Lowres$ baseline, $\NNthirty$ and $\NNforty$. The contour maps show velocity magnitudes $\|\vel\|$.
From the reference we can obtain the location of the separation point ($X_s=12.46$) and reattachment point ($X_r=15.03$). Compared to the reference, the baseline shows significant deviations, with the separation point shifted upstream to $X_s=10.34$ and the reattachment length of the bottom recirculation bubble drastically shortened to $X_r=9.27$.
Similar discrepancies have been reported by Schafer~\cite{schafer2009dynamics} in 3D simulations. They attributed such errors to grid-induced oscillations, where the coarse grid amplifies unstable shear layers, accelerating the transition from laminar to chaotic flow. The amplified motions increase vertical transport, thereby reducing the primary reattachment length. Although the current case is in 2D and thus lacks the spanwise dynamics inherent to 3D flows, a comparable effect is observed: insufficient resolution leads to a significantly shortened reattachment length. This systematic deviation highlights the limitations of the baseline and the opportunity for learned models to recover more physically accurate behavior. 
Our learned models, $\NNthirty$ and $\NNforty$, significantly improve flow predictions, closely aligning with the reference solution by accurately capturing both the separation point (at $X_S=12.46$) and the reattachment point ($X_R=15.03$) 
with an error of less than $6.3\times 10^{-2}$.

\begin{figure*}
    \centering
    \includegraphics[height=0.32\textwidth]{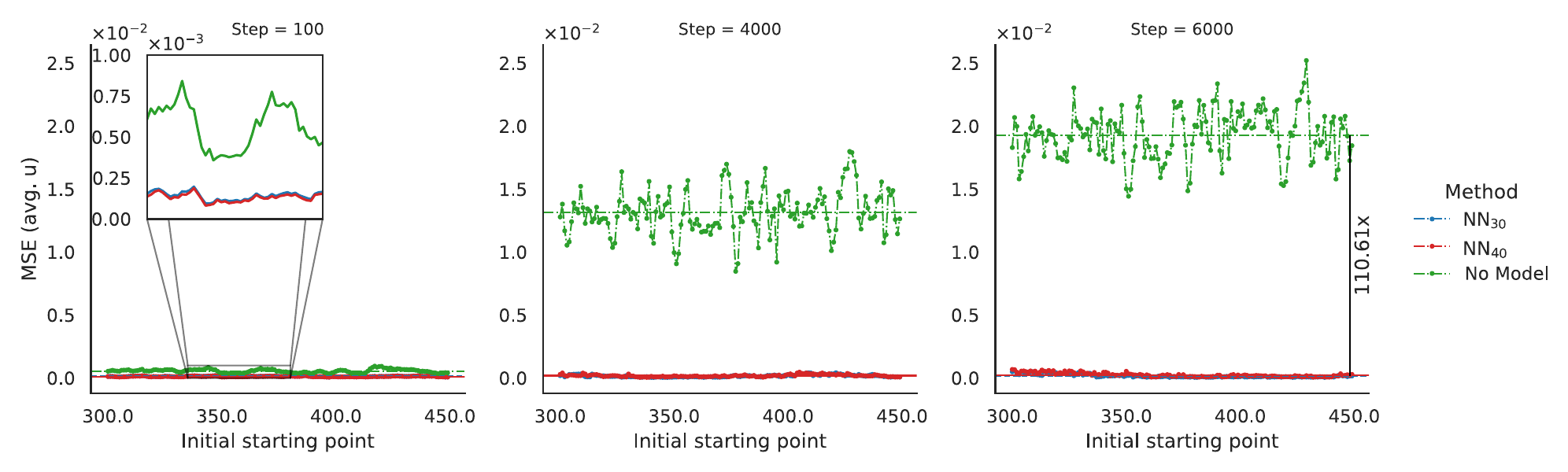} \\
    \caption{MSE (avg. $\vel$) of all methods across a wide sample range of initial conditions after simulating 100, 4000, and 6000 timesteps, respectively.}
    \label{fig:BFS_MSE}
\end{figure*}

Since the statistical result is the primary objective for current task, \myreffig{fig:BFS_MSE} shows the MSE of the temporally averaged
velocity compared to the reference across a sample range of initial conditions at three simulation lengths: 100, 4000, and 6000 $\dt$. Clearly, our methods consistently outperform the baseline approach, with the baseline model exhibiting significantly higher MSE values. At step $=6000$, the MSE of our methods ($1.744 \times 10^{-4}$) is about 110 times lower that than of the baseline ($1.928 \times 10^{-2}$). Despite the elevated MSE values, the baseline shows considerable fluctuations, attributed to varying levels of difficulty for different initial states. In contrast, our models demonstrate consistently stable performance, underscoring their robustness and reliability. Building on previous 
studies~\cite{le1997direct, durst1983flows},  
we also include the comparison of the wall skin-friction coefficient on the top and bottom wall, as well as the streamwise velocity profiles at different locations downstream the change of geometry.

\begin{figure*}[ht]
    \centering
    \includegraphics[width=0.9\textwidth]{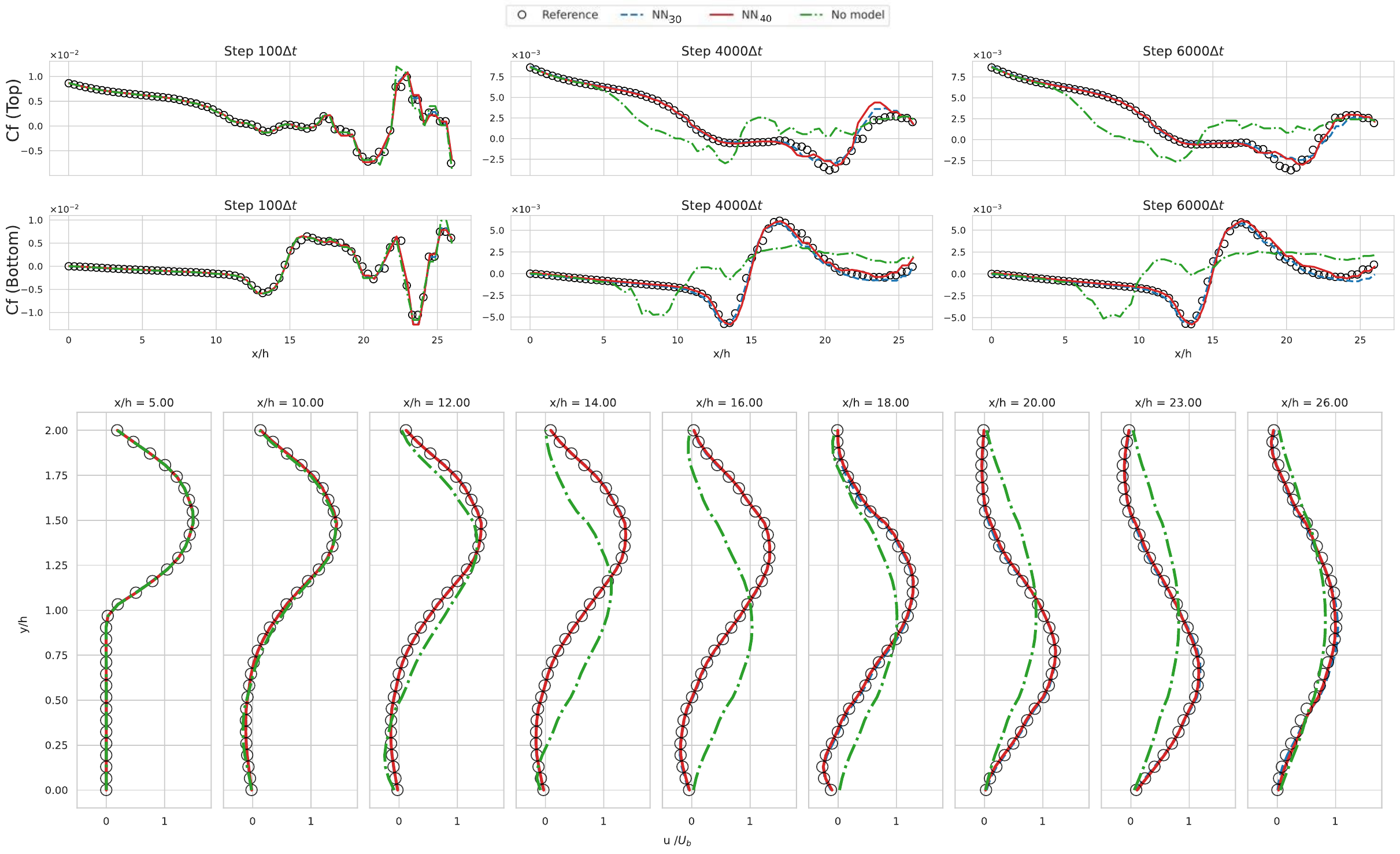}
    \caption{
    Top: Bottom and top wall skin-friction coefficients after three different simulation lengths (100$\Delta t$, 4000$\Delta t$, 6000$\Delta t$),
    Bottom: Velocity profiles at selected streamwise locations after 6000$\Delta t$. In both cases, the black dots show the velocity profiles of the high resolution reference.
    }
    \label{fig:BFS_skin_friction}
\end{figure*}

The wall skin-friction coefficient ($C_f$) is a key metric for assessing wall shear dynamics, providing insights into flow separation, recirculation, and reattachment~\cite{jovic1995reynolds}. It is calculated as 
\begin{equation}
    \label{eq:skin_friction}
    C_f = \frac{\tau_w}{\frac{1}{2} \rho U_{b}^2} ,
\end{equation}
where $\tau_w$ is the wall shear stress, expressed as $ \tau_w = \mu \left. \pderInl{u}{y} \right|_\text{wall} $. Figure~\ref{fig:BFS_skin_friction} (top) illustrates the $C_f$ for different lengths of simulation.
At the beginning of the simulation at 100$\Delta t$, there is barely any difference between the three methods. As the flow evolves over time, the $\Lowres$ baseline exhibits significant deviations from the reference starting from $x/h>5$ for $C_f$ of both top and bottom walls.
Conversely, our methods show a very good agreement with the reference, demonstrating an improved stability and accuracy over extended time steps. The reattachment length, which can be determined as the location where the sign of $C_f$ changes, is accurately captured by both learned models. 
Minor deviations only occur near the outlet, where the grid resolution is about 20 times coarser than near the step.
We also consider the velocity profiles at various streamwise locations ($x/h$) after 6000 $\Delta t$ in \myreffig{fig:BFS_skin_friction} (bottom). Here, the baseline model fails to capture the velocity accurately, particularly in regions of separation and reattachment, leading to discrepancies in both magnitude and shape.
Our methods significantly improve the physical accuracy at lower resolution, although the additional unrolling steps of $\NNforty$ do not further improve the accuracy of the learned correction. The methods are evaluated for more than one characteristic time length of the vortex dynamics~\cite{list2025differentiability}, and they both closely align with the reference solution at all locations.

Overall, the MSE analysis highlights the significant error reduction achieved by the learned methods, demonstrating their effectiveness in correcting coarse grid inaccuracies. This is further substantiated by the wall skin-friction coefficient and velocity profiles, which demonstrate accurate reconstruction of wall-bounded flow dynamics, as well as precise capturing of separation and reattachment points. 
Collectively, these results show the ability of the PICT solver with a learned corrector model to maintain long-term stability and accuracy in coarsely resolved simulations.

\subsection{Turbulent Channel Flow} \label{sec:TCFlearnedSGS}

\begin{figure}
    \centering
    \includegraphics[width=\textwidth]{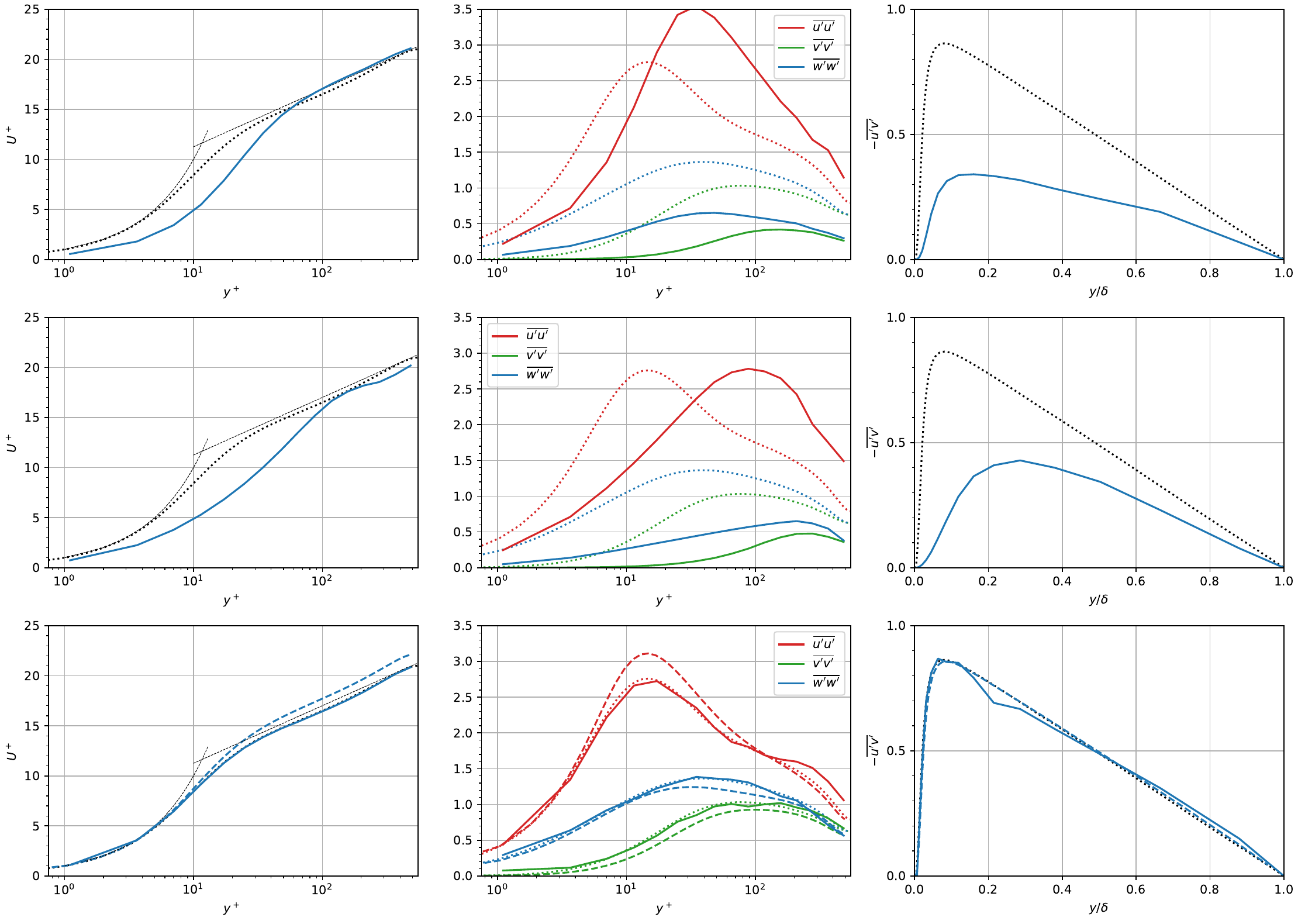}
    \caption{
    Turbulence statistics for 3D TCF at Re 550 with different SGS models.
    Top to bottom: no SGS, SMAG, our learned CNN SGS.
    The average $\Rewall$ resulting from these simulations are 390, 452, and 548, respectively.
    The statistics have been non-dimensionalised with the same expected $\velwall = 0.03658$ for comparability.
    The dotted line is the reference from Hoyas and Jimenez \cite{TCF_2008_10},
    the dashed line represents the statistics of a high-resolution simulation from OpenFOAM.
    } \label{fig:TCF550learnedSGS}
\end{figure}
\begin{figure}
    \centering
    \includegraphics[width=\textwidth]{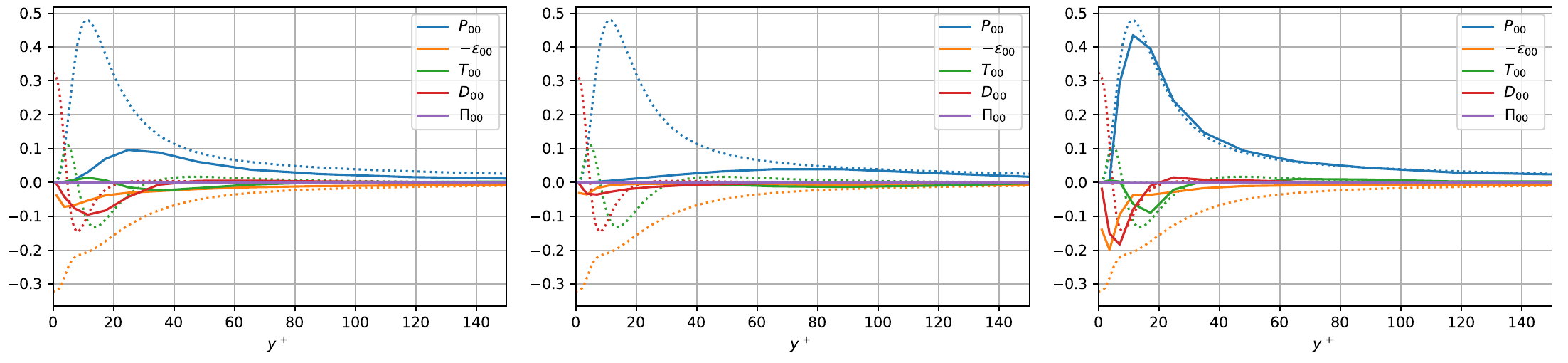}
    \caption{
    Comparison of turbulent energy budgets of different SGS models.
    Left to right: no SGS, SMAG, our learned  CNN SGS.
    The statistics have been non-dimensionalised with the same expected $\velwall = 0.03658$ for comparability.
    The dotted line is the reference from Hoyas and Jimenez \cite{TCF_2008_10},
    the terms are explained in \myrefsec{sec:OnlineMoments}.}
    \label{fig:TCF550learnedSGS_tke}
\end{figure}
\begin{figure}
    \centering
    \includegraphics[width=\textwidth,trim={0.15in 0.15in 0.15in 0.15in},clip]{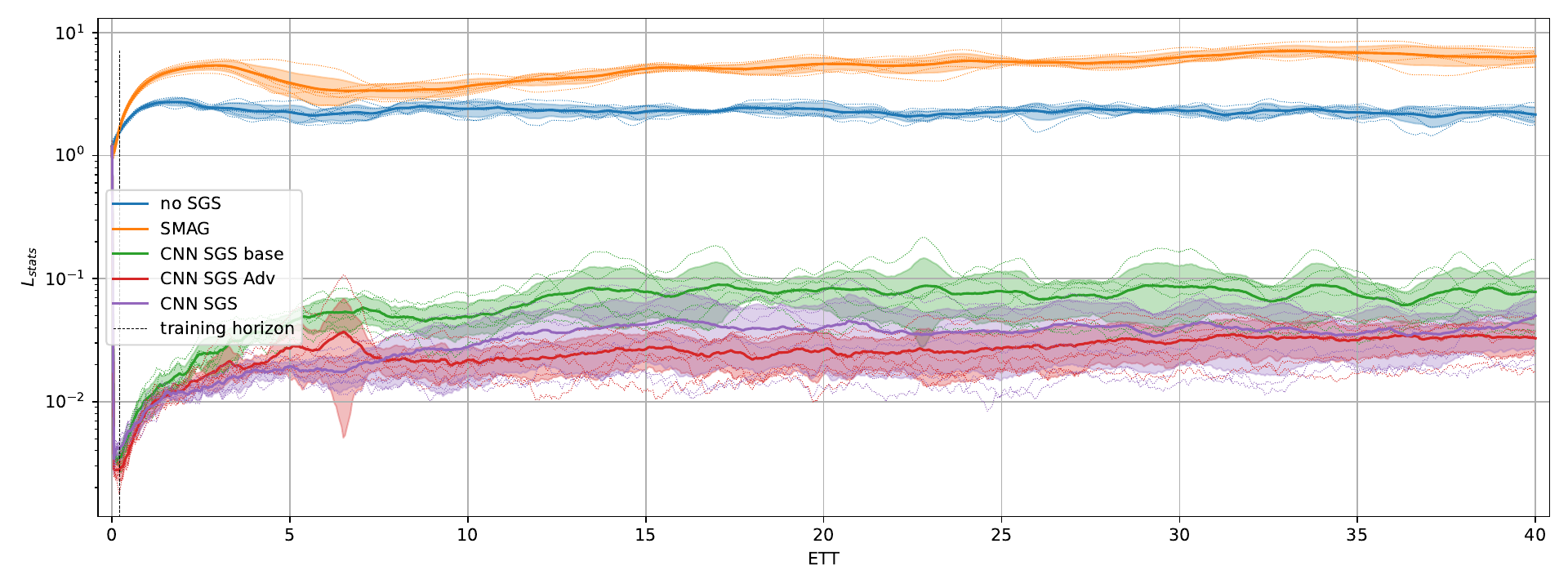}
    \caption{
    A comparison of the per-frame statistics losses from \myrefeq{eq:TCFloss} over a long-term rollout consisting of 20000 simulation steps. In training, the CNN models sees at most 108 steps, indicated by the dashed vertical line.
    All cases are evaluated for 6 different initial conditions that are sampled from an uncorrelated simulation with a turbulence statistics loss close to 1.
    The solid line shows the mean loss over all initial conditions, the fine dotted lines are the individual initial conditions; shaded areas show the standard deviation.
    } \label{fig:TCF550learnedSGS_cmp}
\end{figure}
\begin{figure}
    \centering
    \includegraphics[width=\textwidth]{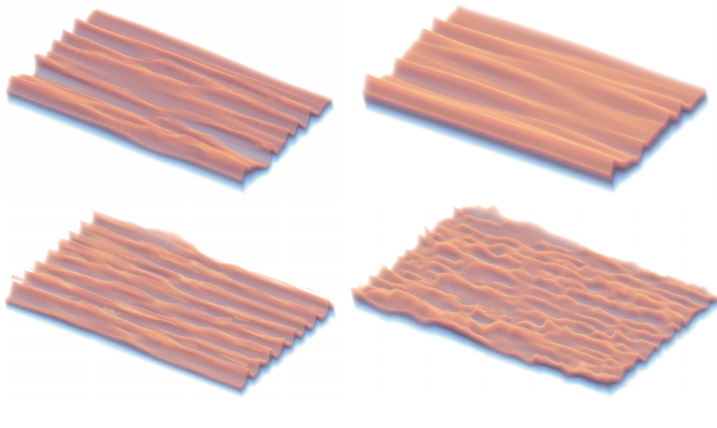}
    \caption{A qualitative visualization of the $\velu$-velocity at the lower boundary of the TCF, comparing the different SGS models to our high-res reference. Top: no SGS and SMAG,
    bottom: our learned SGS and down-sampled reference.}
    \label{fig:TCF550learnedSGS_vis}
\end{figure}
For our final learning setup we target a three dimensional scenario with a turbulent channel flow (TCF) setup.
This setup contains a periodic channel with no-slip boundaries at $\pm y$ which is driven by a dynamic forcing $\viscosity \left. \pderInl{\overline{\velcomp}}{y} \right|_\text{wall}$ to prevent a loss of energy.
The TCF is a well studied scenario that requires high spatial and temporal resolution as well as long simulation times for the turbulence statistics to converge~\cite{TCF_2008_10}. 
In this scenario, coarse spatio-temporal resolutions quickly yield incorrect statistics, and hence it is crucial to introduce a form of numerical modeling of the unresolved scales to obtain an accurate solution without excessive runtimes.
In line with previous experiments, we train an SGS model in form of a corrector $G_\NNw$ that is tasked to estimate a correcting force $\source_\NNw$ at a low spatial resolution of $64 \times 32 \times 32$. For the TCF, $G_\NNw$ receives  the instantaneous velocity and the normalized wall distance $1-|y/\delta|$ as inputs, i.e.,  $\source^n_\NNw = G_\NNw (\vel^n, 1-|y/\delta|)$. 
The term $1-|y/\delta|$ is added to inform the network of the grid refinement in regions near the wall.
As neural network architecture for $G_\NNw$ we employ a simple CNN with layers using 8, 64, 64, 32, 16, 8, 4, and 3 filters, each having a kernel size of $3^3$. Only the last layer uses a kernel size of $1^3$. This gives 198931 trainable parameters in total.
ReLU activations are used for all but the last layer. 
Different from the previous scenarios, the dynamics in the 3D TCF are highly turbulent and matching individual realizations of the flow from simulations at different resolutions no longer provides a physically meaningful learning target.
Hence, we instead aim for matching the turbulence statistics via the statistic loss from~\myrefeq{eq:TCFstatsLossGeneric}, complemented with a regularization term on the generated forcing.

The complete loss we use for training our learned SGS model is given by 
\begin{equation} \label{eq:TCFloss}
    L = L_{stats} + \lambda_\source \frac{1}{N} \sum_{t=0}^{N-1}||\source_\theta^n||_2^2.
\end{equation}
We additionally constrain the forcing to the $[-2,2]$ range to stabilize early training.
Aside from the loss terms guiding the model towards producing the desired turbulence statistics, it proved essential to prevent un-physical network outputs, which in our case means violating the incompressibility assumption. 
To ensure divergence free flow motions, we include the gradient modification from~\myrefeq{eq:SgradPsgrad} for $\source_\NNw$. 

As starting point for the training rollouts we simulate 36\ETT~without any SGS model, and store 160 equally spaced frames from the last 16\ETT. We simulate further 22\ETT~to obtain a starting point for evaluation.
During training we apply warm-up steps where the simulation with the corrector is rolled out for a number of non-differentiable steps and backpropagation is activated afterwards \cite{prantl2022guaranteed}. This mitigates distribution shift, and allows us to train with longer time horizons with a stabilizing effect on the learning process, without requiring more memory for backpropagation operations.

Based on our results from \myrefsec{sec:gradPathAblation}, we consider two different approaches for the training of the CNN model: in the first one, we exclude the gradients of the linear solves from the optimization (i.e. only using $J^{\text{none}}$ from \myrefeq{eq:jnone}). In the second variant, we start in the same way, but include the terms at a later stage of training for fine tuning.
The network is trained in four different phases 
where the warm-up steps are sampled uniformly random from $[0,0],~[0,12],~[0,24],$ and $[0,96]$, respectively. The first phase consists of 6k optimization steps, while the other three include 20k steps. The warm-up is always followed by 12 steps of unrolling, for which we backpropagate through the PICT solver.
We denote the result of this initial training as \textit{CNN SGS base}.
We then fine-tune the network with gradients from the advection linear solve, called \textit{CNN SGS Adv}, for another 20k optimization steps, again with $[0,96]$ steps of warm-up.
To ensure a fair comparison, we also continue training with the initial model as \textit{CNN SGS} for another 20k optimization steps, but without the additional gradient terms.
As can be seen from \myreffig{fig:TCF550learnedSGS_cmp}, the additional gradients from the advection solve only marginally improve the results while roughly doubling the training time.
This is consistent with our results from \myrefsec{sec:gradPathAblation} where a rollout of 10 steps did not benefit from the additional gradient terms.
The learning task for the TCF models potentially also benefits from the fact that the network still receives feedback for every simulation step via the turbulence statistics loss, making a backpropagated transport of the gradients through the advection unnecessary.

As baselines for comparison we use a low resolution simulation without modeling the sub-grid scales (no SGS) as well as a LES version with the Smagorinsky model (SMAG)~\cite{smagorinsky} with $C_s=0.1$. 
Since this model is not correct near solid surfaces, we also apply van-Driest scaling towards the walls to avoid excessive wall friction as a result of the added viscosity.
An evaluation of the learned SGS model is shown in \myreffig{fig:TCF550learnedSGS}, where the statistics where accumulated over the course of 20\ETT, equivalent to 10000 steps of simulation.

For the no SGS and SMAG variants, the mean streamwise velocity $U^+$ deviates noticeably from the reference, although SMAG is closer to the target $\Rewall$ despite the model having almost no influence in the near wall region.
When comparing second order statistics, we can observe how the non-modelled simulation and the simple turbulence model are insufficient to correctly simulate the channel flow at the chosen resolution.
Our learned corrector, on the other hand, matches the target statistics very well, and the resulting $\Rewall= 548$ is very close to the target value of 550.
Since matching the statistics used in a training loss is expected for a well trained network, we also compare the resulting turbulent energy budgets to corresponding reference values in \myreffig{fig:TCF550learnedSGS_tke}.
Despite not being trained on these quantities, our model accurately matches the reference budget terms, 
although slight deviations are noticeable in this evaluation, in particular close to the wall and in the dissipation term.
However, it is worth noting that both baselines fail to match any of the budget terms.
When investigating the higher order moments, skewness and flatness, the results start to deviate from the reference, with both the CNN SGS and no SGS matching the reference only qualitatively. The CNN SGS also shows deviations near the wall, as shown in \myrefapp{app:3DTCF:moments}.

As is evident from \myreffig{fig:TCF550learnedSGS_cmp}, where the per-frame errors in terms of statistics over a long rollout are shown, our model is not only about two orders of magnitude more accurate than no model and SMAG, but also stays stable for a rollout more than 180$\times$ longer than the training horizon. All versions quickly correct the statistics from the initial state, and, after a slight deterioration beyond the training horizon, successfully keep the error at a low level. In comparison, the Smagorinsky model maintains a level of accuracy that is comparable to the simulation without turbulence modeling.
The learned SGS model does not show artifacts in the temporal correlations, in line with the reference, while no SGS exhibits some spurious periodicity, see \ref{app:3DTCF:convergence}.
When observing the instantaneous velocity fields, pictured in \myreffig{fig:TCF550learnedSGS_vis}, we can observe how the flow in the simulation with the learned corrector maintains the topological features of flows at this Reynolds number, in particular the streamwise streaks spacing is closer to the reference than the no SGS and SMAG simulations. The latter shows a overly-smooth flow close to the wall when compared to the reference. 

Overall, our results on the 3D TCF show a highly stable corrector network that matches the spectral reference even better than simulations at much higher resolution, as is evident from the comparison to a high resolution simulation with  OpenFOAM in \myreffig{fig:TCF550learnedSGS}.
Aggregating the normalized errors in the statistics towards the statistics of the spectral reference, our learned SGS model has a MSE of $8.78\times10^{-3}$ while the MSE of OpenFOAM is $36\%$ higher with $1.19\times10^{-2}$.
Details of the aggregated error calculations are provided in \myrefapp{app:errorTcf}.

\subsection{Runtime Performance}

As computational resources are a crucial aspect of CFD simulations, we report the runtime performance of our solver-network simulators in the various scenarios. However, due to the inherent difficulty of comparing different hardware and different software implementations, these performance numbers only serve to provide a rough estimate.
Our comparison is performed considering simulations with similar accuracy, independently from the number of points used in each simulation.
The reference simulations for each task, detailed in the Appendix, are compared against the learned hybrid solver employing PICT together with a neural network component.

For the two dimensional vortex street case, training the $\NNeight$ and $\NNsixteen$ requires 15h and 20h, respectively. During inference, the learned solver requires 181.9s to simulate a sequence of 100s. The trained CNN in this case uses 21\% of the runtime, with 79\% being accounted for by the PICT solver. For this scenario we compare to a solution obtained by PICT alone with a higher spatial resolution of $100 \times 54$. This medium resolution and our PICT+NN solver both run on a single Nvidia RTX~2080~Ti GPU.
With this hardware, only running PICT requires 240s, while yielding a lower accuracy (with MSE = 0.104) than the hybrid simulator (MSE = 0.076). Due to the relatively small spatial resolution, the runtime improvement of 32\% for PICT+NN is modest, but it is nonetheless interesting to note that the corrections inferred by the NN improve temporal stability: the oscillations caused by the coarse grid lead to small time steps for the No Model simulation, and certain simulations led to runtimes of up to 305.3s when no NN was used.
The BFS case, executed on the same hardware,
requires a runtime of 986.4s for a typical evaluation run simulating 120s (with MSE = $1.744 \times 10^{-4}$). In contrast, a medium resolution simulation at $256 \times 64$  with a similar level of accuracy (MSE = $2.578 \times 10^{-4}$) requires 1729.5s. Thus, the full simulation is 75\% slower than the version with the neural network.
As scaling effects become particularly important for larger systems with more degrees of freedom,
the three-dimensional TCF is the most challenging and interesting case.

For the TCF cases, 
training a corrector NN takes about 42 hours for the initial four phases combined, using a single Nvidia GTX~1080~Ti GPU. The fine-tuning with gradients from the advection linear solve takes another 26 hours, 46\% of which are needed for the backwards advection solve, while fine-tuning without them takes 13 hours.
For inference, we compare the runtime performance of our GPU-based solver to OpenFOAM's PISO implementation, running on 32 cores of a Xeon Gold 5220R. These CPUs provide 34.4 TFLOPS of compute resources, in comparison to 11.34 TFLOPS available to the PICT solve. 
In each case, we measure the wall clock time it takes to simulate 20ETT.
OpenFOAM needs 2.5h for the finely resolved simulation at $192 \times 96 \times 96$. 
Despite the higher resolution, it yields a 36\% higher aggregated error in comparison to PICT with the learned SGS model, where the error is computed across all turbulence statistics.
Our solver with learned SGS model takes 223s, 38.7\% of which are required for the neural network.
This means that the corrected simulation with PICT is ca.~40$\times$ faster than OpenFOAM, 
while still matching the reference statistics with a substantially higher accuracy, and running on a single consumer-grade GPU. Naturally, the learned solver is less general than OpenFOAM, but the substantial speedup and its accuracy nonetheless point to the very significant potential of solver-NN hybrids for turbulent, three-dimensional flows.

\section{Conclusion} \label{sec:conclusion}

In this paper we presented the differentiable fluid simulator PICT.
We validated the solver's accuracy 
and analysed the correctness of the gradients provided by our simulator both with numerical methods and in simple optimization tasks.
For optimization tasks with shorter rollout lengths, the modularity of our solver also provides the option to exclude the most expensive parts of the backpropagation, namely the backwards linear solves, to gain runtime improvements without adverse effects on the optimization.

Having established the simulator's applicability to learning tasks, we showed its efficacy in a number of learning applications. In two challenging 2D settings, we trained stable corrector networks to yield accurate solutions with low-resolution simulations over long rollouts.
The learning setup was adapted to three-dimensional turbulent channel flows, enabling the corrector to recover the statistical properties of turbulence over rollout horizons exceeding the training window by orders of magnitude, without requiring direct supervision.
Despite the uncertainties inherent to deep learning, such as generalization and the lack of guarantees for unconstrained aspects of the solution, our learned corrector remains remarkably stable, even for different initial conditions. And while it is limited to the geometry seen during training and not necessarily reproduces the details of higher order moments, it still matches the expected energy budgets.
It additionally outperforms traditional solvers like OpenFOAM both in terms of accuracy and computational performance.
Together, these tests show the efficacy of using learned corrector networks in conjunction with coarse simulations to retain high fidelity behavior in the flows. 
The availability of the PICT solver as open source\footnote{\url{https://github.com/tum-pbs/PICT}} provides a powerful foundation for advancing learned turbulence modeling for the research community.

All trained models are CNNs, which limits them to regular grid inputs and requires custom padding for multi-block setups. An extension to unstructured meshes is possible via GNNs (graph neural networks), which are typically more expensive. Investing into a differentiable re-gridding to transfer from unstructured to regular meshes is another possibility to run CNNs on unstructured simulations.
Our models use the complete velocity fields as input, making the availability of the dense fields a necessity, but allow the model to use more than local information, e.g., the model employed as TCF SGS has a receptive field of $15^3$, running on a grid of size $64 \times 32 \times 32$. If sparse sensors or actors are targeted, local models like those employed in SciMARL~\cite{SciMARL2022} could also be trained with our solver.
While our solver is implemented with GPU support for increased performance, it is currently limited to a single GPU. 
As future work, our simulator could be extended to work in multi-GPU setups, and to include multi-grid solvers for the advection and pressure systems, from which we expect further performance benefits. 
Overall, the presented results are indicative of our solver's ability to address optimization tasks like initial and boundary value reconstruction, learning for control problems, and, with an extension for differentiable transformations, potentially also for tasks like shape optimization~\cite{chen2021numerical}.
Moreover, the solver’s full differentiability enables promising opportunities in uncertainty quantification for fluid dynamics through probabilistic and diffusion-based learning frameworks~\cite{liu2024airfoils,zhuang2025spatially}.

\newpage

\appendix

\section{Method Details} \label{app:method}

\subsection{Definitions}
{
\centering
\begin{tabular}{rl}
    Velocity & $\vel$ \\
    Pressure & $\pressure$ \\
    Viscosity & $\viscosity$ \\
    External Sources & $\source$ \\
    Time & $t$ \\
    Timestep (superscript) & $n$ \\
    Correction, Update & $^*, ^{**}, ^{***}$ \\
    Spatial Location, Cell (subscript) & $i$ \\
    Spatial Gradient & $\nabla$ \\
    Divergence & $\nabla \cdot$ \\
\end{tabular}

}

\subsection{PISO Algorithm} \label{app:PISO}
Here we reproduce the formulation of the original PISO algorithm we implemented in our solver as a  notationally consistent baseline for further derivations.
The PISO algorithm comprises a predictor step to solve the momentum equation (\ref{eq:momentum}) and advance the simulation in time, followed by typically 2 corrector steps to enforce continuity (\ref{eq:continuity}) on the result.
For the predictor step
\begin{equation} \label{eq:advection}
    \frac{1}{\Delta t} \vel^{\guess}
    + \nabla \cdot (\vel^{n} \vel^{\guess})
    - \viscosity \nabla^2 \vel^{\guess}
    = \frac{1}{\Delta t} \vel^{n} - \nabla \pressure + S^{n}
\end{equation}
the velocity is split into velocity from previous step $\vel^{n}$ (advecting) and the velocity guess $\vel^{\guess}$ (advected) to linearize the equation.
In matrix form this becomes the linear system
\begin{equation} \label{eq:advectionMat}
    \AdvMat \vel^\guess = \frac{1}{\Delta t} \vel^{n} - \nabla \pressure + S,
\end{equation}
which is solved for $\vel^\guess$.

For the corrector step the matrix $\AdvMat$ is split into its diagonal $\AdvDiag$ and off-diagonal entries $\AdvNDiag$.
With
\begin{equation} \label{eq:pressureRHS}
    \pressureRHSvec = \AdvDiagInv \left( - \AdvNDiag \vel^{\guess} + \frac{\vel^n}{\Delta t} + \source^n \right),
\end{equation}
the pressure correction comes from the linear system
\begin{equation} \label{eq:pressure}
    \nabla^2 (\AdvDiagInv \pressure^{\guess}) = \nabla \cdot \pressureRHSvec,
\end{equation}
which is solved for the pressure $\pressure^{\guess}$.
This pressure is then used to compute the corrected, divergence-free velocity $\vel^{\guess \guess}$ with
\begin{equation} \label{eq:correction}
    \vel^{\guess \guess} = \pressureRHSvec - \AdvDiagInv \nabla \pressure^{\guess}.
\end{equation}

The pressure correction, equations \ref{eq:pressureRHS}, \ref{eq:pressure} and \ref{eq:correction}, are repeated twice \cite{issa1986solution}, with an additional $^*$ indicating the second round of updates. The velocity of the next time-step is then $\vel^{n+1} := \vel^{\guess\guess\guess}$.

\subsection{Discretization}  \label{app:discretization}

We discretize the PISO algorithm following Maliska \cite{maliska2023fundamentals} and Kajishima and Taira \cite{kajishima2016computational}. 
Since we focus on the finite volume method, we adopt a set-based notation to indicate discrete cells, their direct neighbors, and the connecting faces.
This allows us to write the equations in a general, dimension independent formulation while avoiding an index-based notation for referencing cells, which would conflict with indexing components of vector quantities like $\velcomp_i$.
We further define:

{
\centering
\begin{tabular}{rl}
    Current cell & $\cell$ \\
    Set of valid neighbour cells of $\cell$ & $\nb \in \Snb$ \\
    Set of faces between $\cell$ and $\Snb$ & $\nbf \in \Snbf$ \\
    Set of (virtual) boundary neighbour cells of $\cell$ & $\nbb \in \Snbb$ \\
    Set of boundary-faces between $\cell$ and $\Snbb$ & $\nbfb \in \Snbfb$ \\
    Set of diagonal neighbors 'tangential' to $\nb$ & $\nbd \in \Snbd$\\
    Set of diagonal boundary faces 'tangential' to $\nbfb$ & $\nbdfb \in \Snbdfb$\\
    Sign of logical face direction on computational grid & $\facesign_\nbf$ \\
    Evaluated/value at current cell & $\evalcell{\Box}$ or $\Box_\cell$ \\
    Evaluated/value at neighbour cell & $\evalnb{\Box}$ or $\Box_\nb$ \\
    (Linear) Interpoaltion & $\lerp{\Box}$
\end{tabular}
}

\vspace{10pt} Using this notation to reference values in one of the matrices is less straight-forward, since the matrix entries relate to two cells. Thus, with current cell $\cell$ with index $i$ and neighbor $\nb \in \Snb$ with index $j$, the entries of a matrix $\AdvMat$ are referenced as:

{\centering
\begin{tabular}{rl}
    Current entry of current cell & $\evalcell{\AdvMat_\cell} = \AdvMat_{ii}$ \\
    Neighbour entry of current cell & $\evalcell{\AdvMat_\nb} = \AdvMat_{ij}$ \\
    Current entry of neighbour cell & $\evalnb{\AdvMat_\cell} = \AdvMat_{ji}$ \\
    Neighbour entry of neighbour cell (its own center) & $\evalnb{\AdvMat_\nb} = \AdvMat_{jj}$
\end{tabular}
}

\vspace{10pt} We write $\nb \rightarrow j$ to indicate that $j$ is the vector component that corresponds to the computational axis of the direction from $\cell$ to $\nb$.
Cell references may be omitted when an equation is purely per-cell.

\subsubsection{Finite Volume Method}
At its core, the finite volume method uses the divergence theorem
\begin{equation}
    \int_V (\nabla \cdot \vel) dV = \oint_S (\vel \cdot \normal) dS,
\end{equation}
which relates the divergence of a vector quantity $\vel$ in a finite volume $V$ to the flux $\vel \cdot \normal$ over its surface $S$.
In its discrete form this becomes a sum over the faces $\nbf \in \Snbf$ of a discrete cell $\cell$
\begin{equation} \label{eq:divtheorem}
    \evalcell{ \nabla \cdot \vel } \approx \sum_{\Snbf} \vel_{\nbf} \cdot \normal_{\nbf} \area_{\nbf},
\end{equation}
where $\normal_{\nbf}$ is the face normal and $\area_{\nbf}$ its area.

\subsubsection{Grid Transformations}

The vertices of the regular grids that make up the blocks need to be transformed to align grid axes to physical boundaries and support refinement in areas of interest.
Since we use a FVM-based formulation, the face fluxes created from eq. (\ref{eq:divtheorem}) need to take the new physical size and orientation of the now-transformed cells and faces into account.
To handle these mesh transformations we use the generalized coordinate system as described
by Kajishima and Taira \cite{kajishima2016computational} and Maliska \cite{maliska2023fundamentals}, which effectively scales $\area_{\nbf}$ and rotates $\normal_{\nbf}$, but allows to precompute the required factors from the mesh coordinates.

Given a physical space with coordinates $x_i$
and a computational space with coordinates $\laxis^j$
the transformation metrics can be computed from the mesh coordinates as matrices
$\transforminv_{ji} := \laxis^j_{x_i} = \pderInl{\laxis^j}{x_i}$.
Following previous work, we use a superscript for $\laxis^j$ in the following.
Together with the determinant $\transformJ = \text{det}(\transform)$ these metrics are used to compute, e.g., fluxes over the face with normal $\laxis^j$ from the physical velocity $\vel$ as
$\cvelcomp^j = \transformJ \laxis^j_{x_i} \velcomp_i$,
with implied summation over $i$.
The transformation metrics $\transforminv$ and $\transformJ$ are computed for the cell centers. The face-fluxes needed for advection are computed with the velocity and transformations from the cell center and then interpolated to the faces.

\subsubsection{Predictor Step} 
To solve the predictor step, \myrefeq{eq:advectionMat}, the matrix $\AdvMat$ and the right-hand-side (RHS) need to be calculated.
$\AdvMat$ is a sparse square matrix where every row and column corresponds to one cell. For every row, the diagonal entry is the cell $\cell$ itself, while off-diagonal entries are its neighbors $\Snb$.
The matrix entries contain the temporal, advective, and diffusive terms as $\AdvMat = \AdvMat^{t} + \AdvMat^\text{adv} + \AdvMat^\viscosity$, as can be seen from \myrefeq{eq:advection}. The temporal term is simply $\AdvMat^{t} = \transformJ/\Delta t$.

\paragraph{Advection Term}
For the finite volume advection, we consider the fluxes $\cvelcomp^j_\nbf$ over the faces $\Snbf$ of each cell. Since we use a collocated grid, we interpolate the fluxes from neighboring cells for each face $\nbf$ between a cell $\cell$ and its neighbor $\nb$ as
\begin{equation}
    \cvelcomp^j_\nbf =
    \evalnbf{\lerp{\cvelcomp^j}} =
    \frac{\cvelcomp^j_\cell + \cvelcomp^j_\nb}{2} =
    \frac{ \evalcell{ \transformJ \left( \transforminv_j \cdot \vel \right) } +
    \evalnb{ \transformJ \left( \transforminv_j \cdot \vel \right) } }{2},
\end{equation}
where $j$ is the component that correspods to the logical/computational direction between $\cell$ and $\nb$.
With this we can compute the central and neighbor entries for each row $i$ on $\AdvMat^\text{adv}$ as
\begin{equation} \label{eq:advMatDiscrete}
    \begin{split}
        \evalcell{\AdvMat^\text{adv}_\nb} &
        := 0.5 \facesign_\nbf \lerp{\cvelcomp^j_\nbf}, \\
        \evalcell{\AdvMat^\text{adv}_\cell} &
        := \sum_{\nb \in \Snb} \evalcell{\AdvMat^\text{adv}_\nb}.
    \end{split}
\end{equation}
Boundary neighbors are included on the RHS and do not appear in the advection term on the matrix.

\paragraph{Diffusion Term}
The second order diffusion term includes squared transformation metrics, which are called $\alpha$ for clarity
\begin{equation}
    \begin{split}
        \alpha_{jk} = \alpha_{kj} &= \transformJ \sum_i \pder{\laxis^j}{x_i} \pder{\laxis^k}{x_i} = \transformJ \transforminv_{j} \cdot \transforminv_{k}
    \end{split}
\end{equation}
For a purely orthogonal transformation, the diffusive matrix components are, with $j$ being the computational axis of $\cell \rightarrow \nb$,
\begin{equation} \label{eq:difMatDiscrete}
    \begin{split}
        \evalcell{ \AdvMat^\viscosity_\nb } &= - \evalnbf{ \lerp{ \alpha_{jj} \viscosity } },\\
        \evalcell{ \AdvMat^\viscosity_\cell } &= \sum_{\Snbf} \evalnbf{ \lerp{ \alpha_{jj} \viscosity } } + \sum_\Snbfb 2 \evalcell{ \alpha_{jj} \viscosity }.
    \end{split}
\end{equation}
The boundary term in the second equation appears only for Dirichlet boundaries, not for Neumann.
For non-orthogonal transformations the tangential gradients at a logical face also influence the diffusive flux, leading to additional terms for direct neighbors and extending the stencil to include also diagonal neighbors (for a total stencil of 9 in 2D and 19 in 3D).
The non-orthogonal additions to the direct neighbor components on matrix are (c.f. Maliska \cite{maliska2023fundamentals} eq. (12.184)):
\begin{equation} \label{eq:difNonOrthoMatDiscrete}
    \begin{split}
        \evalcell{ \AdvMat^\viscosity_\nb } &= -
        \sum_{ \Snbd } \left( \evalloc{ \lerp{ \alpha_{ij} \viscosity } }{\nbdf} + \evalnbf{\lerp{ \alpha_{ij} \viscosity}} \right) \facesign_\nb \facesign_\nbd / 4 \\
        \evalcell{ \AdvMat^\viscosity_\cell } &= - \sum_\Snb \sum_{ \Snbd } \evalnbf{ \lerp{ \alpha_{ij} \viscosity } } \facesign_\nb \facesign_\nbd / 4 
    \end{split}
\end{equation}
where $i$ is the axis of $\nb$ and $j$ that of $\nbd$. The $\evalnbf{\lerp{ \alpha_{ij} \viscosity}}$ terms all cancel out unless the cell is at a boundary.
Handling of the diagonal neighbors is described in \ref{sec:nonOrthoDiscrete}.

\paragraph{Boundaries and Sources}
Boundary values appear on the RHS akin to source terms,
where any external sources $\source$ are also added. Thus, the RHS for the velocity contains a part of the discrete temporal derivative and the advective and diffusive boundary fluxes as
\begin{equation} \label{eq:velRHSdiscrete}
    \velRHS = \frac{\vel^n}{\Delta t} + \frac{1}{\transformJ_\cell} \sum_\Snbfb \evalnbfb{ \vel^n \left( 2\alpha_{jj}\viscosity - \cvelcomp^j \facesign \right) } + \source.
\end{equation}
The viscosity term again only appears for Dirichlet boundaries. Since $\vel$ and $\transforminv$ are defined directly at the boundary face $\nbfb$, $\cvel$ and $\alpha$ are not interpolated.
To handle non-orthogonal grids with diagonal neighbors on the RHS, \myrefeq{eq:velRHSnonOrtho} is added.

\subsubsection{Pressure Correction}
Using the divergence theorem \myrefeq{eq:divtheorem},
the term $\nabla^2 \pressure = \nabla \cdot \left( \nabla \pressure \right)$ takes the discrete form
\begin{equation}
    \evalcell{\nabla^2 \pressure} = \sum_\Snbf \evalnbf{\transformJ \transforminv_j \cdot \left( \transforminv^T \nabla_\laxis \pressure \right)},
\end{equation}
where the resulting matrix entries appear very similar to the diffusion terms, only using $\AdvDiagInv$ instead of $\viscosity$ and a different sign:
\begin{equation} \label{eq:pMatDiscrete}
    \begin{split}
        \evalcell{ \PMat_\nb } &= \evalnbf{ \lerp{ \alpha_{jj} \AdvDiagInv }}\\
        \evalcell{ \PMat_\cell } &= \sum_{\Snbf} - \evalnbf{ \lerp{ \alpha_{jj} \AdvDiagInv }}
    \end{split}
\end{equation}
The non-orthogonal treatment is also equivalent,
\begin{equation}
\begin{split}
        \evalcell{ \PMat_\nb } &= \sum_{ \Snbd } \left( \evalloc{ \lerp{ \alpha_{ij} \AdvDiagInv } }{\nbdf}
        \right) \facesign_\nb \facesign_\nbd /4,\\
    \end{split}
\end{equation}
where we omit any $\evalnbf{\lerp{ \alpha_{ij} \AdvDiagInv}}$ terms since they will always cancel out.
The (implicit) Neumann boundaries need to be handled differently, however.
For simplicity we ignore the prescribed boundary gradient and instead use one sided differences for face-tangential gradients on boundary-orthogonal faces.

The pressure RHS, $\nabla \cdot \pressureRHSvec$, includes $\pressureRHSvec$ from \myrefeq{eq:pressureRHS} and the velocity boundary terms from \myrefeq{eq:velRHSdiscrete}. In full:
\begin{equation} \label{eq:pressureRHSdiscrete}
    \pressureRHSvec = \AdvDiagInv \left( \frac{\vel^n}{\Delta t} + \frac{1}{\transformJ_\cell} \sum_\Snbfb \evalnbfb{ \vel^n \left( 2\alpha_{jj}\viscosity - \cvelcomp^j \facesign \right) } + \source -  \AdvNDiag \vel^\guess \right) 
\end{equation}
The divergence is computed with the divergence theorem and $\pressureRHSvec$ as vector field as
\begin{equation} \label{eq:pressureRHSdivDiscrete}
    \pressureRHS_\cell :=
    \evalcell{\nabla \cdot \pressureRHSvec} = \sum_{\nb \rightarrow j} \evalnbf{ \lerp{ \transformJ \transforminv_j \cdot \pressureRHSvec } } \facesign_\nb
\end{equation}
The non-orthogonal pressure components from diagonal neighbors must be added after the divergence, see \myrefeq{eq:pressureRHSnonOrtho}.

\paragraph{Velocity Correction}
Finally, to make the velocity divergence free, the gradient of the computed pressure is applied to $\pressureRHSvec$:
\begin{equation} \label{eq:velCorrDiscrete}
    \vel^{\guess \guess} = \pressureRHSvec - \AdvDiagInv \nabla \pressure
\end{equation}
The spatial pressure gradient $\nabla \pressure$ needed here is computed after Maliska \cite{maliska2023fundamentals}, eq. (12.193) - (12.195), as 
$\nabla \pressure = \transforminv^T \nabla_\laxis \pressure$,
where $\nabla_\laxis$ is the spatial gradient on the untransformed (computational) grid, computed with finite differences over the cell
\begin{equation}
    \left( \nabla \pressure \right)_i = \sum_j \transforminv_{ji} \frac{\pressure_{j+1} - \pressure_{j-1}}{2},
\end{equation}
where we use $j \pm 1$ to indicate the neighbor cells in the respective logical direction.

\subsubsection{Non-Orthogonal Grids} \label{sec:nonOrthoDiscrete}
For non-orthogonal grids, the second order derivatives (diffusion and pressure) extend the stencil to also include diagonal neighbors, leading to additional entries in the matrix.
As an alternative, the entries from the diagonal neighbors can be moved to the RHS \cite{maliska2023fundamentals},
which keeps the matrix stencil small, but can slow down convergence on highly non-orthogonal grids.
In this case multiple linear solves, sometimes called non-orthogonal corrector steps, with intermediate updates to the diagonal neighbor entries on the RHS to include the updated $\vel^\guess$ or $\pressure^\guess$ may be necessary, depending on the mesh.
If this approach is chosen, as we have in our simulator, the RHS of the predictor step $\velRHS$ is 
extended by
\begin{equation} \label{eq:velRHSnonOrtho}
    \sum_{\Snb} \sum_\Snbd - \facesign_\nb \evalnbf{\lerp{\alpha_{jk} \viscosity}} \facesign_\nbd \vel^{\guess-1}_\nbd + 
    \sum_\Snbfb \sum_\Snbdfb - \facesign_\nbfb \alpha_{jk\nbfb} \facesign_\nbdfb \viscosity_\nbdfb \vel^n_\nbdfb /2,
\end{equation}
where the sum over $\Snbfb$ only includes Dirichlet boundaries. For the pressure, $\nabla \cdot \pressureRHSvec$ is extended by
\begin{equation} \label{eq:pressureRHSnonOrtho}
    \sum_{\Snb} \sum_\Snbd - \facesign_\nb \evalnbf{\lerp{\alpha_{jk} \AdvDiagInv}} \facesign_\nbd \pressure^{\guess-1}_\nbd,
\end{equation}
which is applied after the divergence computation.

\subsection{Boundary Conditions}
For connections between blocks, the boundary specification merely complicates the neighbor cell lookup, but does not otherwise influence the algorithm.
For prescribed boundaries like Dirichlet and Neumann conditions, any required values, e.g., velocities and transformation metrics, are defined directly on the cell boundary face, as opposed to on a virtual boundary cell outside the domain.
The pressure conditions for Dirichlet velocity boundaries are implicitly 0-Neumann. For the implementation these prescribed boundaries can be largely ignored as the pressure correction should not change the boundary velocity.

In addition to Dirichlet boundary conditions, we implement a non-reflecting advective outflow boundary \cite{advout1992}.
The advective outflow updates a Dirichlet boundary between each PISO step by advecting the block's boundary cell layer into the boundary with a predetermined characteristic velocity $\vel_m$ to satisfy
\begin{equation} \label{eq:advOut}
    \frac{\partial \vel}{\partial t} + \vel_m \frac{\partial \vel}{\partial x_i} = 0,
\end{equation}
which prevents the boundary from reflecting flow structures back into the domain.
The discrete update before every PISO step is
\begin{equation}
    \vel_\nbfb^{n+1} = \vel_\nbfb^{n} -
    \left( 1- \frac{1}{1 - 2 \Delta t \transforminv_j \cdot \vel_m} \right)
    \left( \vel_\nbfb^{n} - \vel_\cell^n \right),
\end{equation}
where $\vel_\cell^n$ is the velocity in the cell at the boundary.
During the PISO step the boundary is then treated as a fixed Dirichlet boundary.
Since our solver is incompressibe, the updated boundary velocities $\vel_\nbfb^{n+1}$ are also scaled such that the in and out fluxes of the domain are balanced.

\subsection{Gradients} \label{app:gradients}

Backpropagation is generally based on the chain rule, where the partial derivative of some composite $g(f(x))$ can be expressed as $\pder{g(f(x))}{x} = \pder{f(x)}{x} \pder{g(f(x))}{f(x)}$.
Since in AD the derivatives are computed by chaining gradient functions of the output gradient,
we write the gradient function for a function $y = f(x)$ as $\gradfnN{x}{y}{f} = \pder{f}{x} \partial y$,
which backpropagates some given gradient $\partial y$ to the inputs of $f$ to obtain $\partial x$. This also allows us to shorten some expressions.
We further exclude the function subscript if it is clear from the context.
Since the derivation of the gradient functions is based on the discrete implementation, the equations are per-cell. Overlapping output gradients, e.g., when an operation provides gradients for neighbor cells, are implicitly accumulated (additive) on the respective cells.

Note that we do not compute derivatives with respect to the transformations. Thus, the transformation metrics stay scalars in the formulation that modify the gradient with respect to other quantities. The differentiable quantities are $\vel, \viscosity, \dens$, and $\source$, where boundary values for $\vel$ and $\dens$ are also differentiable, as well as any derived intermediate quantities like matrices and RHS of the linear systems.

\subsubsection{Pressure Correction}
Since backpropagation goes backwards through the algorithm we start our derivation of the gradient equations at the end.
For the gradients of the velocity correction, \myrefeq{eq:velCorrDiscrete}, we compute
\begin{equation}\label{eq:diff:velcorr_grad_advdiag}
    \begin{split}
        \gradfn{\AdvDiag}{\velcomp^{\guess\guess}} &= - \sum_i \left( \nabla \pressure \right)_i (-1)\AdvDiag^{-2} \partial \velcomp^{\guess\guess}_i,\\
    \end{split}
\end{equation}
\begin{equation}\label{eq:diff:velcorr_grad_pressure}
    \begin{split}
    \gradfn{\pressure}{\velcomp^{\guess\guess}} &= \sum_{\Snb \rightarrow j} -0.5 \facesign_\nbf \evalnb{\left( \transforminv \AdvDiagInv \partial \velcomp^{\guess\guess} \right)_j}
    \end{split}
\end{equation}
via
\begin{equation}\label{eq:diff:velcorr_grad_pressure_via}
    \begin{split}
        \partial \pressure \left( \partial \nabla_\laxis \pressure \right) &= \sum_{\Snb \rightarrow j} -0.5 \facesign_\nbf \evalnb{\left( \partial \nabla_\laxis \pressure \right)_j}\\
        \partial \nabla_\laxis \pressure \left( \partial \nabla \pressure \right) &= \transforminv \partial \nabla \pressure\\
        \partial \nabla \pressure \left( \partial \velcomp^{\guess\guess} \right) &= \AdvDiagInv \partial \velcomp^{\guess\guess},
    \end{split}
\end{equation}
and finally for the re-used RHS of the pressure system simply
\begin{equation}\label{eq:diff:velcorr_grad_pressurerhs}
    \gradfn{{\pressureRHSvec}}{\vel^{\guess\guess}} = \partial \vel^{\guess\guess}.
\end{equation}
After computing
$\gradfn{\PMat}{\pressure}$ and $\gradfn{\pressureRHS}{\pressure}$
through the backwards linear solve,
the gradients of the pressure matrix, \myrefeq{eq:pMatDiscrete}, are
\begin{equation}
    \begin{split}
        \gradfn{\AdvDiag}{\PMat} &=
        \sum_{\Snb \rightarrow j} 0.5 \left( \evalcell{ \partial \PMat_\nb - \partial \PMat_\cell } + \evalnb{ \partial \PMat_\cell - \partial \PMat_\nb } \right)
        \evalcell{- \AdvDiag^{-2} \alpha_{jj}},
    \end{split}
\end{equation}
excluding gradients from the non-orthogonal treatment for simplicity.

For the divergence in pressure system's RHS, \myrefeq{eq:pressureRHSdivDiscrete} ($\pressureRHS = \nabla \cdot \pressureRHSvec$), we have
\begin{equation}\label{eq:diff:divpressurerhs_grad_pressurerhs}
    \begin{split}
        \gradfn{\pressureRHSvec_\cell}{\pressureRHS} &=
        \sum_{\Snb \rightarrow j} 0.5 \left( \evalcell{\transformJ \transforminv_j \partial \pressureRHS } + \evalnb{\transformJ \transforminv_j \partial \pressureRHS} \right) \facesign_\nb\\
        &= \sum_{\Snb \rightarrow j} \evalnbf{ \lerp{ \transformJ \transforminv_j \partial \left( \nabla \cdot \pressureRHSvec \right) } } \facesign_\nb
    \end{split}
\end{equation}
and, if non-orthogonal corrector steps are used,
\begin{equation}\label{eq:diff:divpressurerhs_grad_advdiag}
    \begin{split}
        \gradfn{\AdvDiag_\cell}{\pressureRHS}
        &= \sum_{\Snb \rightarrow j} \sum_{\Snbd \rightarrow k}  0.5 \facesign_\nb \evalloc{0.25 \pressure^{\guess-1} \facesign}{\nbd} \evalcell{\alpha_{jk} \AdvDiag^{-2} \partial \pressureRHS} \\
        &+\sum_{\Snb \rightarrow j} \sum_{\Snb_{\perp} \rightarrow k} 0.5 \facesign_\nb \evalloc{0.25 \pressure^{\guess-1} \facesign}{\nb_{\perp}} \evalcell{\alpha_{jk} \AdvDiag^{-2}} \evalnb{\partial \pressureRHS}
    \end{split}
\end{equation}
\begin{equation}\label{eq:diff:divpressurerhs_grad_pressure}
    \begin{split}
        \gradfn{\pressure^{\guess-1}_\cell}{\pressureRHS} &= \sum_{\Snb \rightarrow j} \sum_{\Snbd \rightarrow k} - 0.125 \facesign_\nbf \facesign_\nbd \left( \evalnb{\alpha_{jk} \AdvDiagInv} + \evalloc{\alpha_{jk} \AdvDiagInv}{\nbd} \right) \partial \pressureRHS_\nbd, \\
    \end{split}
\end{equation}
where $\pressure^{*-1}$ is the pressure result of the previous non-orthogonal step, $\Snb_\perp$ is the set of neighbors of $\cell$ in the directions orthogonal to $\nb$, and the influence of tangential boundaries has been omitted for clarity.
Finally, we compute the gradients of the pressure RHS, \myrefeq{eq:pressureRHSdiscrete},
\begin{equation}\label{eq:diff:pressurerhs_grad_vel}
    \gradfn{\vel^n}{\pressureRHSvec} = \frac{1}{\AdvDiag \Delta t} \partial \pressureRHSvec
\end{equation}
\begin{equation}
    \begin{split}
        \gradfn{\velcomp^{n}_{i \nbfb}}{\pressureRHSvec} &= \evalcell{\AdvDiagInv \frac{1}{\transformJ}} \left( \evalnbfb{ 2 \alpha_{ii} \viscosity - \cvelcomp^i \facesign } \partial \pressureRHSveccomp_i  
        - \sum_{j} \evalnbfb{ \velcomp^n_{i} \facesign \transforminv_{ij} } \partial \pressureRHSveccomp_j \right)
    \end{split}
\end{equation}
\begin{equation}\label{eq:diff:pressurerhs_grad_velsource}
    \gradfn{\source}{\pressureRHSvec} = \AdvDiagInv \partial \pressureRHSvec
\end{equation}
\begin{equation}\label{eq:diff:pressurerhs_grad_velguess}
    \begin{split}
    \gradfn{\vel_\cell^{\guess}}{\pressureRHSvec} &= - \sum_{\Snb} \evalnb{\AdvNDiag_\cell \AdvDiagInv \partial \pressureRHSvec},
    \end{split}
\end{equation}
\begin{equation}\label{eq:diff:pressurerhs_grad_visc}
    \begin{split}
        \gradfn{\viscosity}{\pressureRHSvec} &= \AdvDiagInv \frac{1}{\transformJ} \sum_i \sum_\Snbfb \evalnbfb{ 2 \velcomp_{i} \alpha_{ii} } \evalcell{\partial \pressureRHSveccomp_i}
    \end{split}
\end{equation}
\begin{equation}\label{eq:diff:pressurerhs_grad_advdiag}
        \gradfn{\AdvDiag}{\pressureRHSvec} = -\AdvDiag^{-2} \sum_i ({\partial \pressureRHSveccomp}_i \AdvDiag)
\end{equation}
\begin{equation}\label{eq:diff:pressurerhs_grad_advndiag}
    \gradfn{\evalcell{\AdvNDiag_\nb}}{\pressureRHSvec_\cell} = - \sum_i \AdvDiagInv \evalnb{\velcomp_i^*} \evalcell{\partial \pressureRHSveccomp_i}
\end{equation}

\subsubsection{Predictor Step}
For differentiation of the prediction step, the gradients $\pder{}{\vel^\guess}$ are first passed through the linear solve
to obtain $\pder{\vel^{\guess}}{\AdvMat}$ and $\pder{\vel^{\guess}}{\velRHS}$, in addition to any direct gradients $\pder{}{\AdvMat} = \pder{}{\AdvDiag} + \pder{}{\AdvNDiag}$ from the pressure backwards step.
Then gradients of the advection matrix, \myrefeq{eq:advMatDiscrete} and (\ref{eq:difMatDiscrete}), can be computed as
\begin{equation}\label{eq:diff:advmat_grad_vel}
    \begin{split}
        \gradfn{\vel^{n}}{\AdvMat}
        &= \sum_{\Snb \rightarrow j} \facesign_\nbf 0.25 \evalcell{ \transforminv_j \transformJ }
        \left( \frac{1}{\transformJ_\cell} \evalcell{\AdvMat_\cell + \AdvMat_\nb}
        + \frac{1}{\transformJ_\nb} \evalnb{\AdvMat_\cell + \AdvMat_\nb} \right)
    \end{split}
\end{equation}
\begin{equation}\label{eq:diff:advmat_grad_visc}
    \begin{split}
        \partial \viscosity \left( \partial\AdvMat \right)
        &= \sum_{\Snb \rightarrow j} 0.5 \left( \evalcell{ \partial \AdvMat_\nb - \partial \AdvMat_\cell } + \evalnb{ \partial \AdvMat_\cell - \partial \AdvMat_\nb } \right)
        \evalcell{\alpha_{jj}} \\
        &+ \sum_{\Snbfb \rightarrow j} 2 \evalcell{\alpha_{jj} \partial \AdvMat },
    \end{split}
\end{equation}
again excluding gradients from the non-orthogonal treatment.

For the advection RHS, \myrefeq{eq:velRHSdiscrete}, the gradients are
\begin{equation}\label{eq:diff:advrhs_grad_vel}
    \gradfn{\vel^n}{\velRHS} = \frac{1}{\Delta t} \partial \velRHS
\end{equation}
\begin{equation}\label{eq:diff:advrhs_grad_velbound}
    \begin{split}
        \partial \velcomp^{n}_{i \nbfb} \left( \partial \velRHS \right) &=
        \frac{1}{\transformJ_\cell} \left( \evalnbfb{2 \alpha_{ii} \viscosity - \cvelcomp^i \facesign } \partial \velcompRHS_i
        - \sum_{j} \evalnbfb{ \velcomp^n_{i} \facesign \transforminv_{ij} } \partial \velcompRHS_j \right)
    \end{split}
\end{equation}
\begin{equation}\label{eq:diff:advrhs_grad_velsource}
    \gradfn{\source}{\velRHS} = \partial \velRHS
\end{equation}
\begin{equation}\label{eq:diff:advrhs_grad_visc}
    \begin{split}
        \partial \viscosity \left( \partial \velRHS \right) &= \frac{1}{\transformJ} \sum_\Snbfb \sum_i  2 \evalnbfb{ \velcomp_{i} \alpha_{ii} } \partial \velcompRHS_i
    \end{split}
\end{equation}
and for the corresponding non-orthogonal correction, \myrefeq{eq:difNonOrthoMatDiscrete},
\begin{equation}
    \gradfn{\velcomp^{\guess-1}_{i}}{\velRHS} = \sum_{\nb \rightarrow j} \sum_{\Snbd \rightarrow k} - 0.125 \facesign_\nbf \facesign_\nbd \left( \evalnb{\alpha_{jk} \viscosity} + \evalloc{\alpha_{jk} \viscosity}{\nbd} \right) \partial \velRHS_\nbd
\end{equation}
\begin{equation}
    \gradfn{\velcomp_{i\nbd}^{\guess-1}}{\velRHS} = - \sum_{\nb_\perp,\Snbdfb \rightarrow k} \facesign_\nbfb \evalloc{\alpha_{jk} \viscosity \facesign}{\nbdfb} \evalloc{\velcompRHS_i}{\nb_\perp}
\end{equation}
\begin{equation} \label{eq:NO_velRHS_grad_visc}
    \begin{split}
        \gradfn{\viscosity_\cell}{\velRHS} &= \sum_i \sum_{\nb \rightarrow j} \left( \sum_{\Snbd \rightarrow k}  -0.5 \facesign_\nb \evalloc{0.25 \pressure^{\guess-1} \facesign}{\nbd} \evalcell{\alpha_{jk} \partial \velcompRHS_i} \right.\\
        &+ \left. \sum_{\nb_{\perp} \rightarrow k} -0.5 \facesign_\nb \evalloc{0.25 \pressure^{\guess-1} \facesign}{\nb_{\perp}} \evalcell{\alpha_{jk}} \evalnb{\partial \velcompRHS_i} \right) \\
        &+ \sum_i \sum_{\Snbfb \rightarrow j} \sum_{\Snbdfb \rightarrow k} - 0.5 \evalnbfb{\facesign \alpha_{jk}} \evalloc{\facesign \velcomp_i^n}{\nbdfb} \evalcell{\partial \velcompRHS_i}
    \end{split}
\end{equation}
assuming $\viscosity$ is global in \myrefeq{eq:NO_velRHS_grad_visc}.

\subsection{Implementation}
As we target a tight integration with machine learning, we implement our differentiable simulator as Python module containing custom GPU operations written in C++/CUDA alongside the necessary data structures.
The multi-block structure is realized as Domain, Block and Boundary classes that each contain their children (Domain has Blocks, Blocks have Boundaries) and the data fields relevant for their level (Domain contains global Matrices and RHS, Blocks the velocity and pressure fields, and Boundaries the boundary values).
It also includes the data structure necessary to store the multi-block structure with its tensors and connections and make them accessible on the GPU.

The underlying fields, e.g., velocity and pressure, are represented as PyTorch tensors, which allows connecting those directly to optimization and machine learning tasks.
From these components, we build the final, combined PISO algorithm in Python to allow for easier customization, integration of learned components, or replacement of individual operations.
The linear systems for prediction and correction are solved with suitable linear solvers implemented with the cuBLAS and cuSparse libraries.
Due to the modularity of the implementation, these solvers could be switched relatively easily, e.g., we plan support for faster multi-grid solvers in future versions of the solver.

The gradient operations for the backwards pass are also implemented as custom CUDA operations, where we implement a single backwards kernel for each forward function (see \myreffig{fig:PISOgraph}) that provides all necessary gradients.
To make this work with PyTorch's native autodiff, each backwards kernel further needs to be wrapped in Python to make it compatible with PyTorch's tensor tracking and compute graph building.
For the adjoint backwards linear solves we can re-use the forward solver code with an option to transpose the matrix.

As linear system solver we use conjugate gradient (CG) for the pressure and bi-conjugate gradient stabilized (BiCGStab) for advection-diffusion. 
These are standard algorithms implemented via cuBLAS and cuSparse library functions as standalone linear solvers, oblivious of the multi-block structure.
For the BiCGStab we also support preconditioning based on an incomplete LU-decomposition. 
This preconditioning is necessary for meshes with significantly varying cell sizes, and hence is enabled on a case-by-case basis.
We also support an option to only use the preconditioner when the un-preconditioned solve has failed.

To facilitate training neural networks on multi-block domains, we provide a custom Multi-Block convolution to seamlessly run the convolution over every block's tensor. This is done by resolving the block connection to perform the typical padding of the convolutional layers with values or features of connected blocks. Otherwise, when using standard zero padding, convolution are prone to  cause artifacts along the block boundaries. With our padding the convolution is essentially oblivious of the block structure.

\section{Additional Validation Cases} \label{app:validation}

Here, we show additional validations for the forward simulation capabilities of our solver in increasingly complex standard benchmark scenarios, showing its accuracy and long-term stability.

\subsection{Plane Poiseuille Flow}
\begin{figure}
    \centering
    \includegraphics[width=\textwidth]{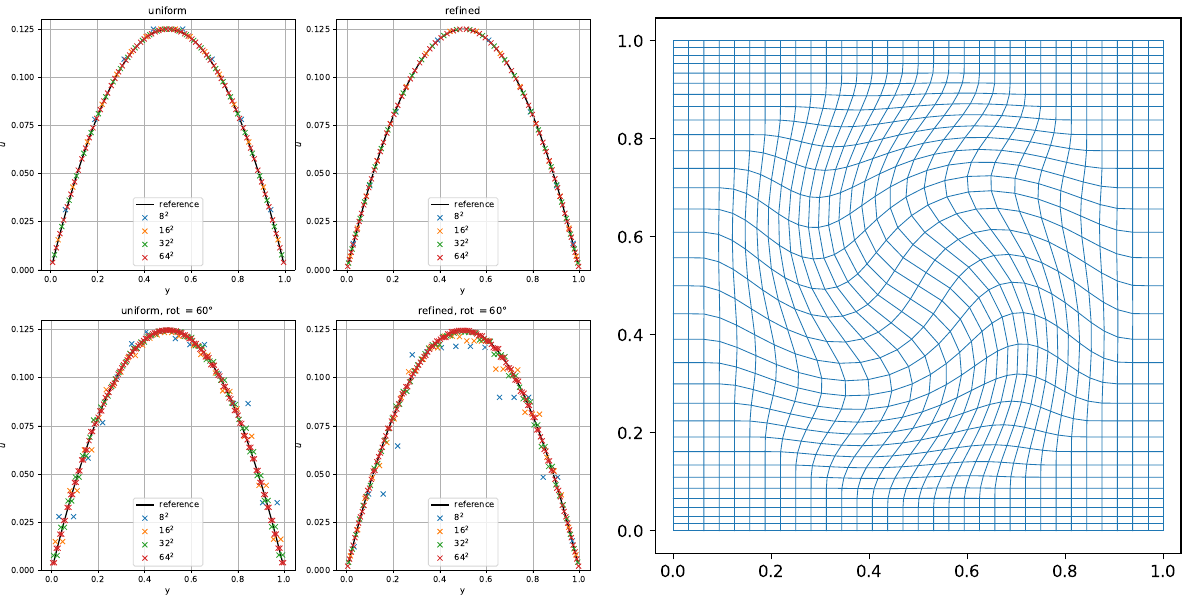}
    \caption{The graphs on the left show vertical u-velocity profiles for the plane Poiseuille flow for increasing resolution from $8^2$ to $64^2$. 'refined' uses a grid refined towards the closed boundaries.
    The lower plots show results from a rotationally distorted grid, for which the refined version is shown on the right.
    The vertically aligned samples come from using nearest-neighbor interpolation to obtain a center line profile.
    } \label{fig:PHPF}
\end{figure}
As the laminar version of the TCF, the plane Poiseuille flow is a simple 2D test case in which the NS equations simplify to have the analytic solution 
$\velu = \frac{G}{2\viscosity} y (1-y)$.
It is a flow through a periodic channel with closed no-slip boundaries and a constant forcing $G$.
In our test we use $y=1$, $\viscosity=1$ and $G=1$, for which the maximum velocity is $\velu_{\text{max}} = 0.125$,
and tested growing resolutions and refinement towards the closed boundaries. All resolutions agree well with the analytic solution, as can be seen in fig. \ref{fig:PHPF}.
For non-orthogonal grid transformations we also tested a grid with rotational distortion around the center of the grid.

\subsection{Lid-Driven Cavity} \label{app:lidcavityFWD}

\begin{figure}
    \centering
    \begin{subfigure}{0.49\textwidth}
        \includegraphics[width=\textwidth]{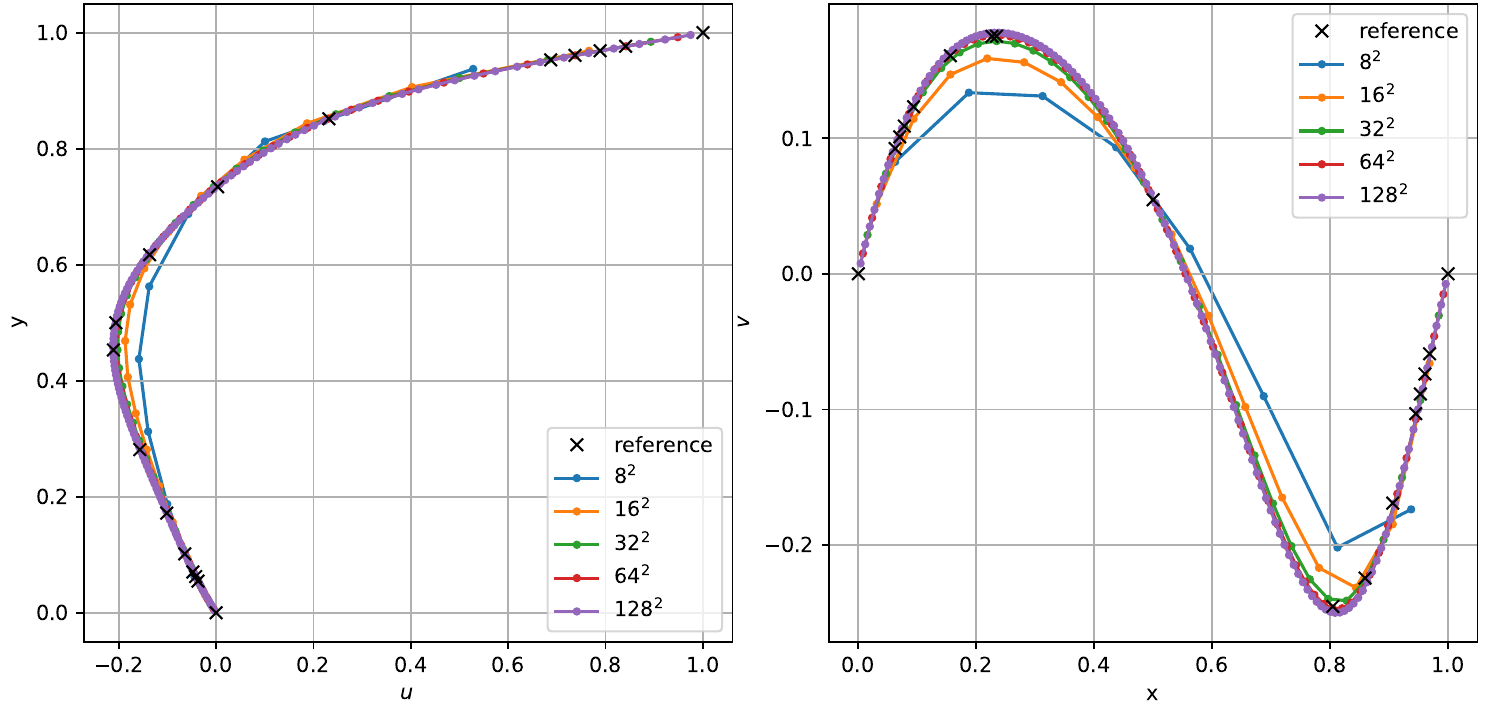}
        \caption{Uniform grid, Re $=100$} \label{fig:app:Lid2d:uniform100}
    \end{subfigure}
    \hfill
    \begin{subfigure}{0.49\textwidth}
        \includegraphics[width=\textwidth]{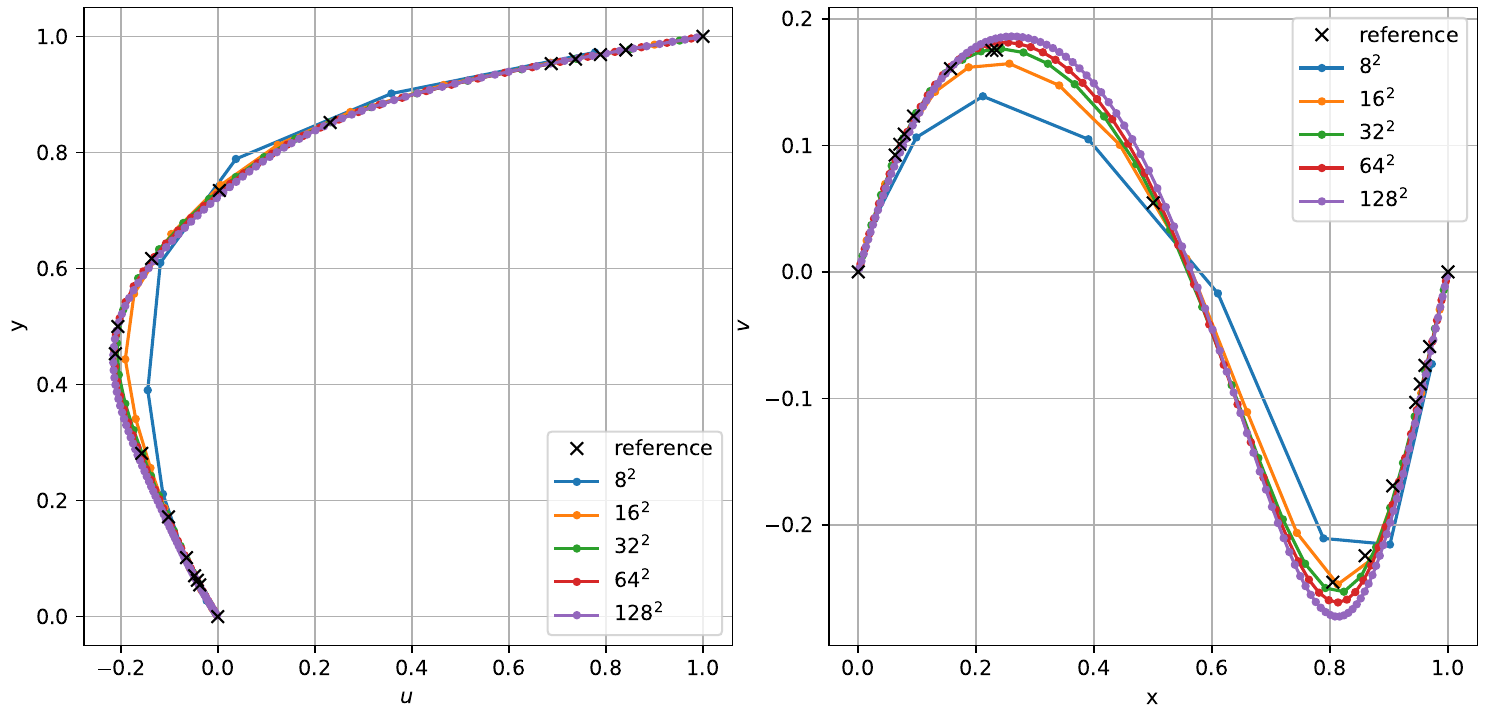}
        \caption{Refined grid, Re $=100$} \label{fig:app:Lid2d:refined100}
    \end{subfigure}\\
    \begin{subfigure}{0.49\textwidth}
        \includegraphics[width=\textwidth]{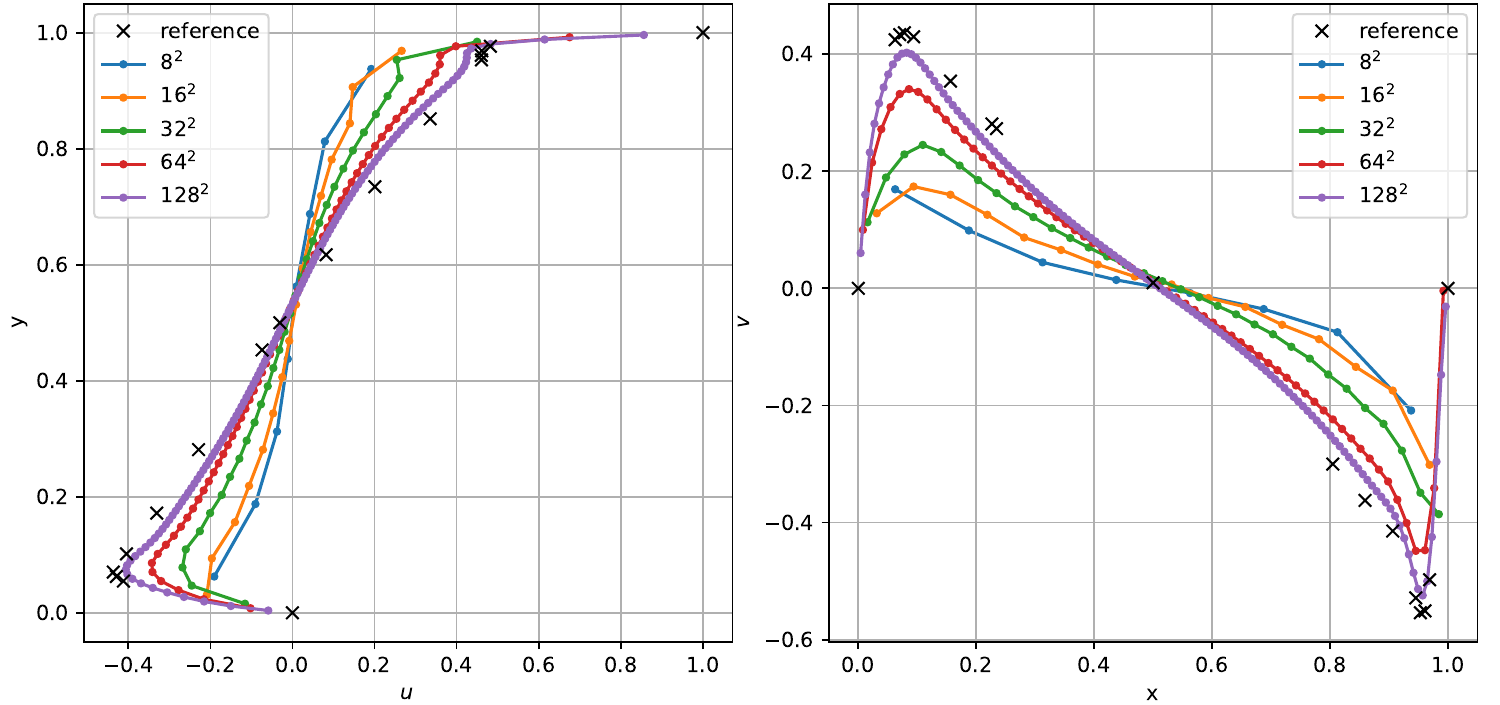}
        \caption{Uniform grid, Re $=5000$} \label{fig:app:Lid2d:uniform5000}
    \end{subfigure}
    \hfill
    \begin{subfigure}{0.49\textwidth}
        \includegraphics[width=\textwidth]{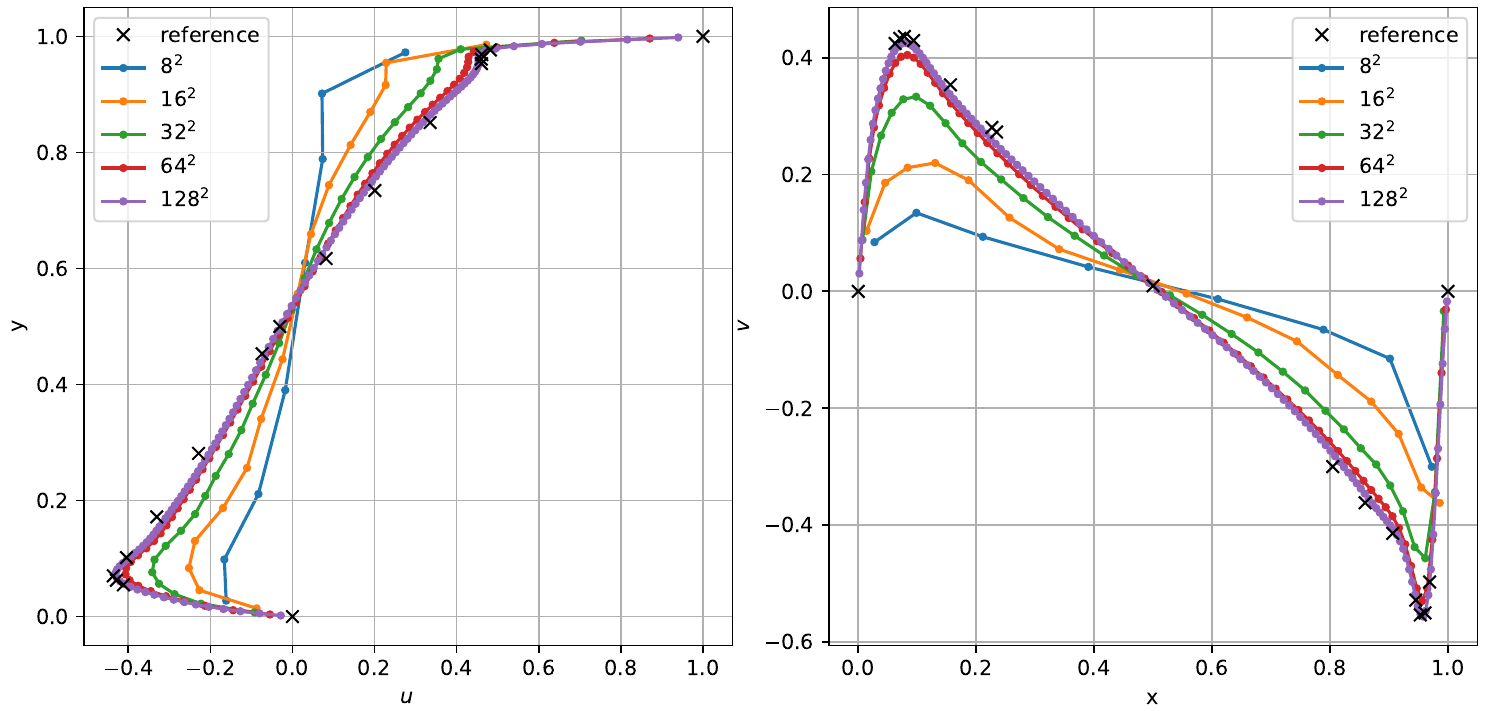}
        \caption{Refined grid, Re $=5000$} \label{fig:app:Lid2d:refined5000}
    \end{subfigure}
    \caption{Velocity profiles for the 2D lid-driven cavity with Re $=100$ and Re $=5000$ for increasing resolutions. The left image of a subfigure is the u-velocity on the vertical center line, and the right is the v-velocity on the horizontal center line. (b) and (d) use a grid that was refined towards all boundaries. The reference is a high-resolution DNS~\cite{Lid2D}.
    } \label{fig:app:Lid2d}
\end{figure}

\begin{figure}
    \centering
    \begin{subfigure}{0.49\textwidth}
        \includegraphics[width=\textwidth]{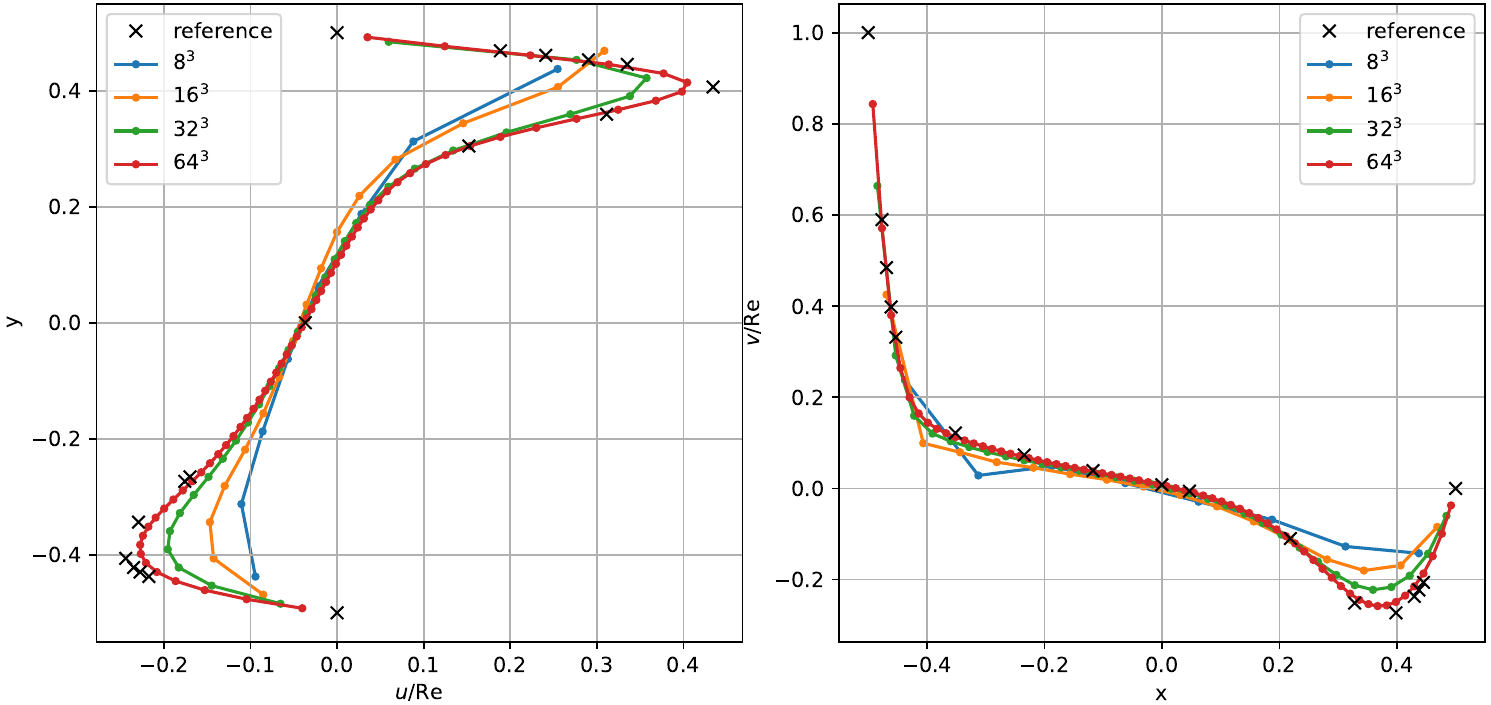}
        \caption{Uniform grid} \label{fig:app:Lid3d1000:uniform}
    \end{subfigure}
    \hfill
    \begin{subfigure}{0.49\textwidth}
        \includegraphics[width=\textwidth]{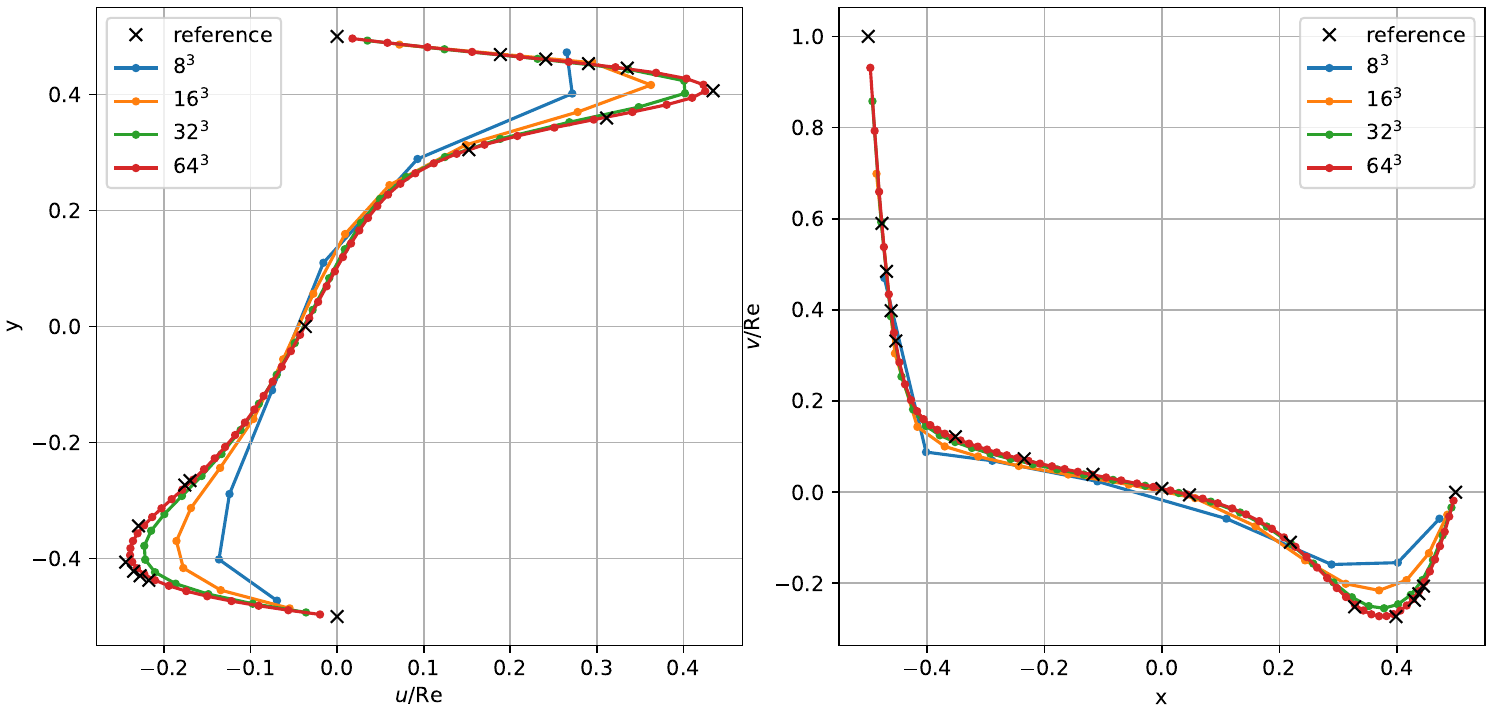}
        \caption{Refined grid} \label{fig:app:Lid3d1000:refined}
    \end{subfigure}
    \caption{Velocity profiles for the 3D lid-driven cavity with $=1000$ for increasing resolutions.
    The plots show the same quantities as in the 2D case, but the velocities are additionally normalized with the Reynolds number.
    The reference is a high-resolution DNS~\cite{Lid3D}.
    } \label{fig:app:Lid3d1000}
\end{figure}

We compare a converged lid-driven cavity simulation to high-res DNS references for 2D \cite{Lid2D}, \myreffig{fig:app:Lid2d}, and 3D \cite{Lid3D}, \myreffig{fig:app:Lid3d1000}, for different Reynolds numbers and with and without grid refinement towards the boundaries. As is evident from the plots, the solution converges to the reference with increasing resolution. For higher Reynolds numbers the refined grid, shown in the right pair, further improves the results, while at lower Reynolds numbers the uniform grid (left pair) performs better. 
Additionally, we tested permutations of the lid and its velocity direction, as well as rotational distortions of the grid similar to those of the plane Poiseulle flow (not shown). The results of the permutations are all identical, while those on a distorted grid are impacted by the worse mesh quality but are still stable and close to the reference.
In the 3D setting we also tested different aspect ratios, meaning the scaling of the x and z size of the cavity, and periodic and closed z-boundaries, where reference values were available.

\subsection{Obstacle Flows}

\begin{figure*}
    \centering
    \includegraphics[width=\textwidth]{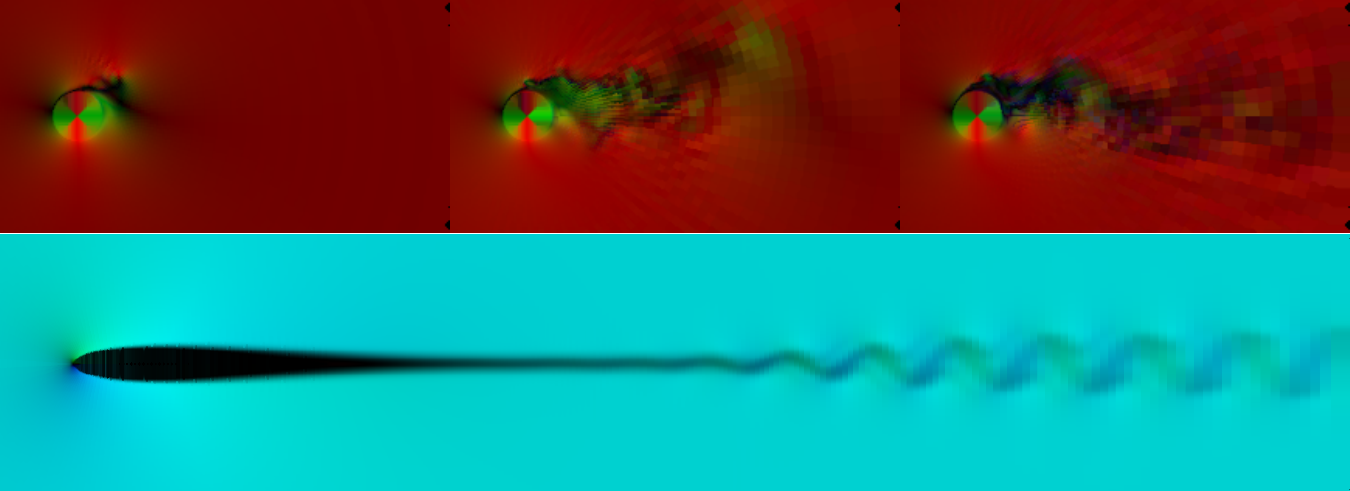}
    \caption{Visualization of the velocity field of two forward simulations. 
    Cells are visualized with a constant value per cell to indicate changing cell sizes of the computational mesh.
    These showcase cropped results from the non-orthogonal transformed grids shown in \myreffig{fig:meshes}.
    For visualization, the velocity fields have been resampled to a regular grid using nearest neighbor interpolation.
    The upper row one shows the center slice of the velocity field of frames 8, 30, and 200 from a 3D flow around a rotating cylinder, where the absolute velocity is directly mapped to RGB.
    The second shows the evolved 2D flow around and behind a NACA 0012 airfoil profile where the 2D velocity is mapped to a color circle.
    } \label{fig:results1}
\end{figure*}
In \myreffig{fig:results1} we show visualizations of two additional flow scenarios around obstacles that make use of non-orthogonal meshes. These results show qualitatively that our solver supports stable simulations on non-orthogonal meshes like O- and C-grids. The corresponding meshes are visualized in \myreffig{fig:meshes}.

\subsection{2D Vortex Street} \label{app:2d_vs}

\begin{figure*}
    \centering
    \includegraphics[height=0.25\textwidth]{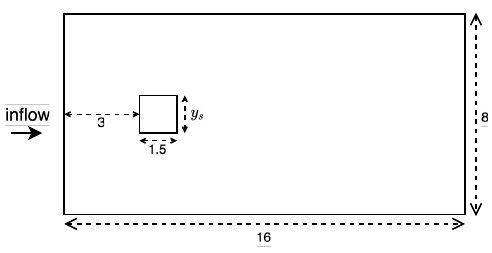}
    \caption{Schematic of the 2D vortex street geometry. The domain extends \(16\) in the streamwise direction and \(8\) in the transverse direction. The square obstacle, with a width of \(1.5\) along the x-axis, is positioned \(3\) downstream from the inlet. The obstacle's vertical position (\(y_{in}\)) and its height (\(y_s\)) vary across cases, as detailed in \myreftab{tab:train_test_Karman}. }
\label{fig:VortexStreetGeometry}
\end{figure*}

Here, we provide additional details about the 2D vortex street setup used in our evaluations. All the geometry parameters have been non-dimensionalized with the characteristic length $y_s$.
The computational domain extends $16$ in the streamwise direction and $8$ in the transverse direction. The obstacle is positioned $3$ downstream from the inlet, with a width of $1.5$ along the x-axis.
This geometry is visualized in \myreffig{fig:VortexStreetGeometry}.
The obstacle height, denoted as $y_s$, varies along the y-axis across different cases, as detailed in \myreftab{tab:train_test_Karman}. The inlet velocity follows a Gaussian profile given by $u_{in} = u \frac{1}{{\sqrt{2\pi \sigma ^2}}} e^{-\frac{y^2}{2\sigma^2}}$, where $u = 1$ and $\sigma = 0.4$. The viscosity is then set as $\nu = \frac{u y_s}{Re}$. An advective boundary condition is applied at the outlet, while the top and bottom walls are no-slip. To enhance accuracy, the grid is refined by a factor of 3 near the top, bottom, and around the obstacle.
Additionally, the resolution is further increased in the blocks surrounding the bluff body with a ratio of 1.5. The final grid resolution varies with the obstacle height $y_s$, which takes values in $\{0.5, 1.0, 1.5, 2.0\}$ . Specifically, the corresponding grid resolutions are $268 \times 132$, $268\times 136$, $268\times 140$, and $268\times 144$, with the increasing vertical extent of the computational domain with larger obstacle heights. 

\subsection{2D Backwards Facing Step} \label{app:2d_bfs}

\begin{figure*}
    \centering
    \includegraphics[height=0.125\textwidth]{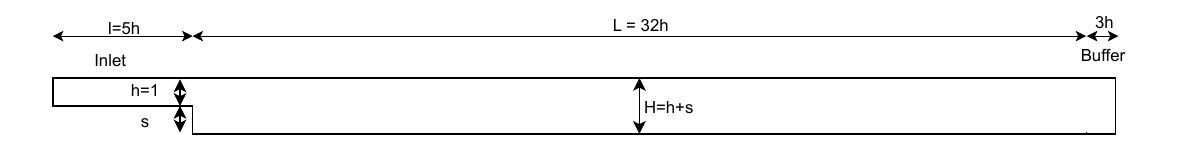}
    \caption{Schematic of the 2D BFS geometry. The gap between step and top wall is $h=1$, and the inlet length is $l=5h$. The main channel has a length of $L=32h$, with a buffer region of $3h$ at the outlet. The total height of the domain is $H=h+s$, where $s$ is the step height offset below the inlet. }
\label{fig:BFSgeometry}
\end{figure*}

\begin{figure*}
    \centering
    \includegraphics[height=0.30\textwidth]{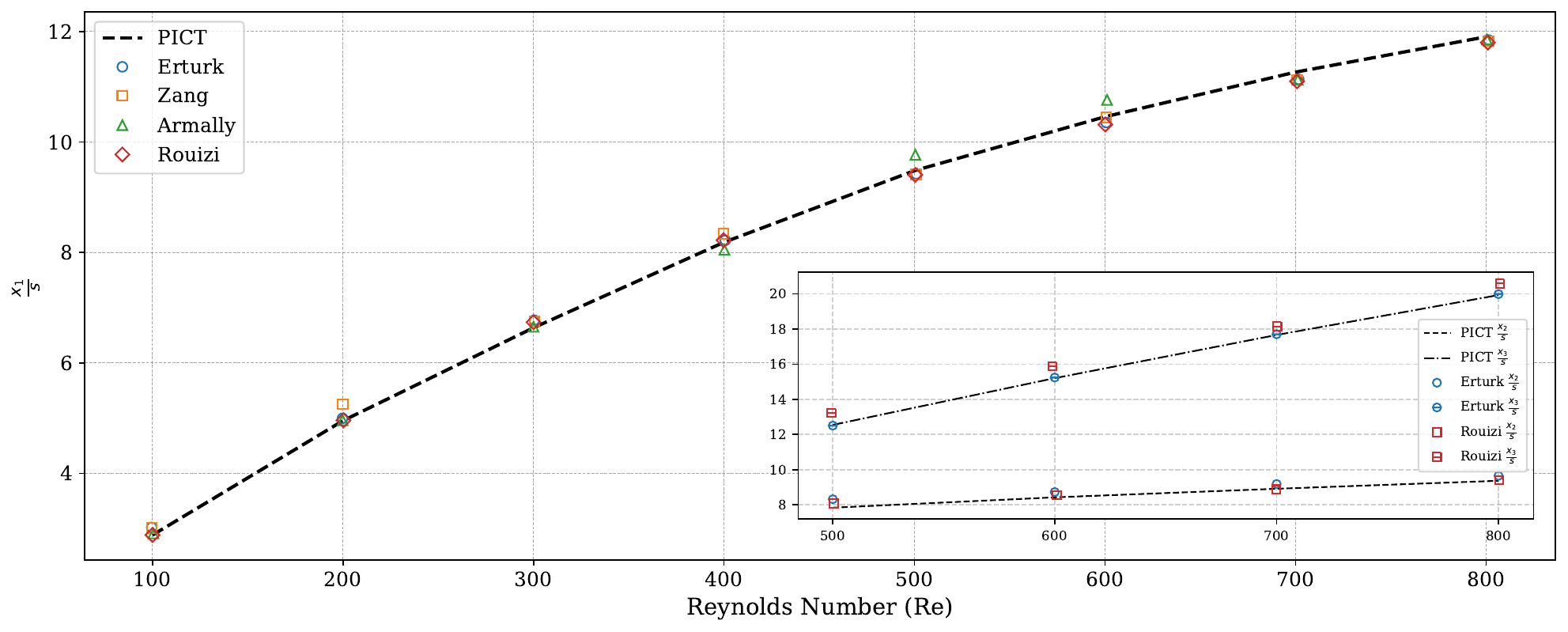} \\
    \includegraphics[height=0.325\textwidth]{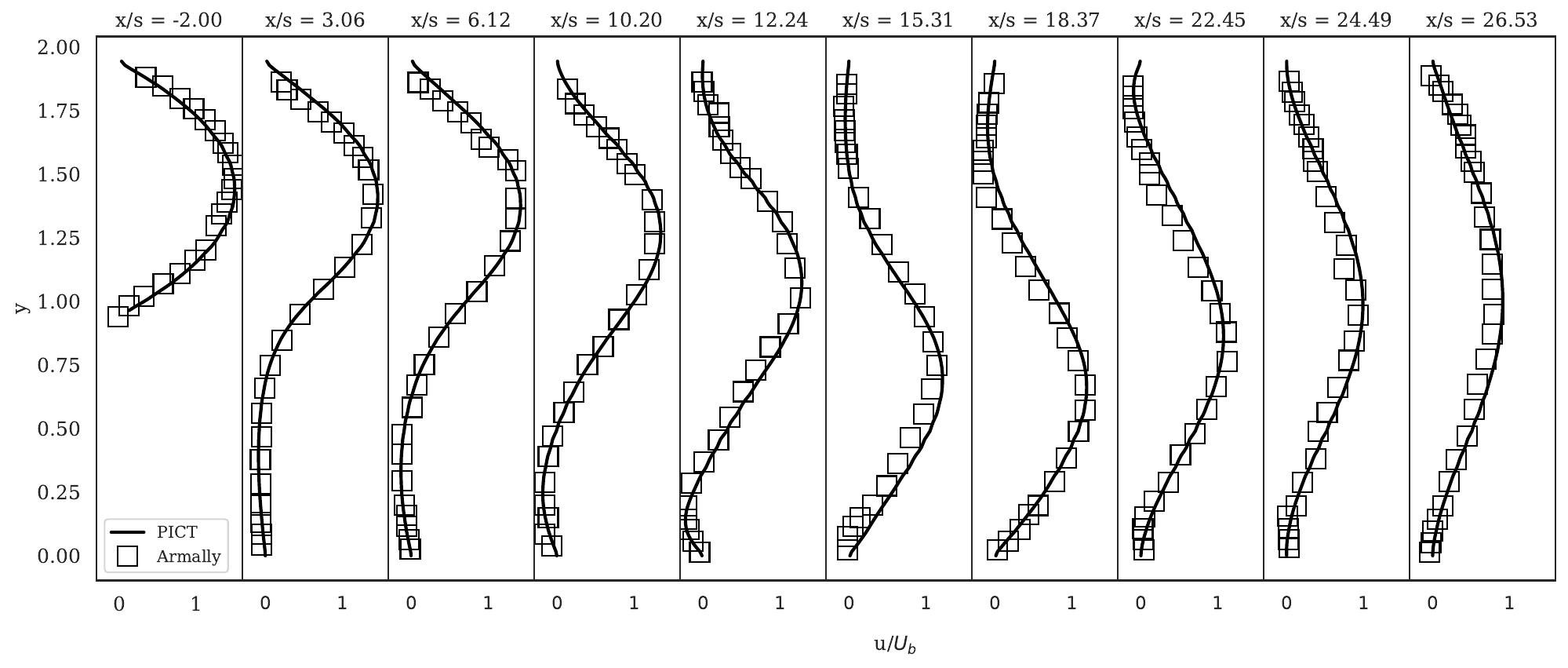}
    \caption{Accuracy validataion of the BFS case. Top: Size of the reattachment length $\frac{x_1}{s}$ respective to the Reynolds number (Re). The locations of the detachment point $\frac{x_2}{s}$ and the reattachment point $\frac{x_3}{s}$ of top wall respective to Re are shown in the inset. Bottom: Velocity profiles comparison for Re = 1290.
    } \label{fig:BFS_accuracy}
\end{figure*}

Above, we tested our solver on the backward-facing step (BFS) flow with different Reynolds numbers. The BFS flow generally includes an inflow with parabolic velocity distribution and a sudden step on the lower side,
as shown in \myreffig{fig:BFSgeometry}.
The geometry setup follows the work by Rouizi et al.~\cite{rouizi2009numerical} with a fixed $h=1$ for the height of the channel before the step. The length before the step is set as $5h$. The length after the step is set as $35h$.
A suitable computational grid with refinement to the step, top and bottom wall has been chosen to ensure the critical regions have been sufficiently resolved. The refinement factor for the vertical direction is 2, resulting in the smallest grid y-size ($\Delta y = 0.01h$) at the top and bottom wall and the horizontal line behind the step, while the maximum grid y-size is $\Delta y = 0.02h$.
For the horizontal direction, the refinement factor of the middle area is 20, resulting in a minimal grid x-size behind the step ($\Delta x = 0.01h$), with the maximum ($\Delta x = 0.2h$) close to the outlet. Refinement factors for the inlet and outlet blocks were adjusted to ensure smooth transitions between adjacent grid sizes, thereby avoiding abrupt changes in cell sizes. 
For the boundary conditions, the inlet parabolic velocity is defined by $U = 6 U_{\text{b}} \frac{y}{h} \left(1 - \frac{y}{h}\right)$, where $U_{\text{b}}=1$.
For the outlet,  advective boundary conditions are used, and in order to avoid the outlet influencing the upstream area, a stabilizing buffer layer of $3h$ with slightly increased viscosity has been applied \cite{list2022_Learned}. For the top, bottom, and step, the no-slip wall boundary condition has been applied. The simulation is run for a period of $t U_b / h = 600$. Generally, the domain in front of the step is discretized using $64 \times 32$ grids. Behind the step, a resolution of $544 \times 128$ is employed, including a block with a resolution of $32 \times 128$ for the buffer layers.
As can be seem from \myreffig{fig:BFS_accuracy}, the flow reattachment lengths and velocity profiles resulting from our solver closely match the reference across all investigated Reynolds numbers.
Training the  models for the BFS scenario took 25h and 32h for $\NNthirty$ and $\NNforty$, respectively.


\subsection{3D Turbulent Channel Flow} \label{app:3DTCF}

For the turbulent channel flow (TCF) scenario, we use a coordinate system with streamwise and spanwise directions along $x$ and $z$, while the wall normal direction is $y$.
A dynamic forcing $\viscosity \left. \pderInl{\overline{\velcomp}}{y} \right|_\text{wall}$ is applied in streamwise direction to drive the flow. The mesh is a regular grid which is refined against the walls as described in the reference.
We consider the case of $\Rewall = 550$ \cite{TCF_2008_10} with a spatial resolution of $192 \times 96 \times 96$ and exponential refinement with a base of $1.095$.
The domain has a physical size of $2\pi\delta \times 2\delta \times \pi\delta$ with $\delta = 1$ being the channel half-width.
For time stepping we used an adaptive time stepper that satisfies CFL$<$0.1, which was necessary to reduce numerical diffusion introduced by larger time steps.
The velocity is initialized using a Reichardt profile~\cite{reichardt1951} with small, divergence free perturbations and we set $\viscosity = \delta / \Recenter$ where $\Recenter$ is the expected centerline Reynolds number calculated via $\Recenter := (\Rewall / 0.116)^{1/0.88} \approx 15037$.

The simulations were run for 20 \ETT~to ensure convergence before the statistics were accumulated over another 20 \ETT, which results in ca. $11000t^+$.
The resulting turbulence statistics can be found in \myreffig{fig:TCF180-TCF550}.
Using $\velwall = \sqrt{\viscosity \left.  \pderInl{\overline{\velcomp}}{y} \right|_\text{wall} } $,
the non-dimensionalization is performed via $\vel^+ = \vel / \velwall$, $y^+ = y \velwall / \delta$,  and $t^+ = t \velwall^2 / \viscosity$. 
The eddy turnover time is \ETTtext $= \ETTeq$.
The flow of the validation simulations is statistically stable and the averaged statistics are close to the spectral reference given sufficient resolution.
The resulting averaged  $\Rewall = \velwall / \viscosity$
is likewise close to the target with a value of 541. 

When training our learned SGS model in this scenario,
we use $\lambda^n_{stats}=0.5$, $\lambda_{U_0}=1$, $\lambda_{U_1}=\lambda_{U_2}=0.5$, $
\lambda_{u'_{ii}}=\lambda_{u'_{01}}=1$, $\lambda_{\nabla\cdot\source} = 10^{-4}$, and $\lambda_\source = 1$
for balancing the loss terms.

\subsubsection{Convergence} \label{app:3DTCF:convergence}

\begin{figure}
    \centering
    \includegraphics[width=\linewidth]{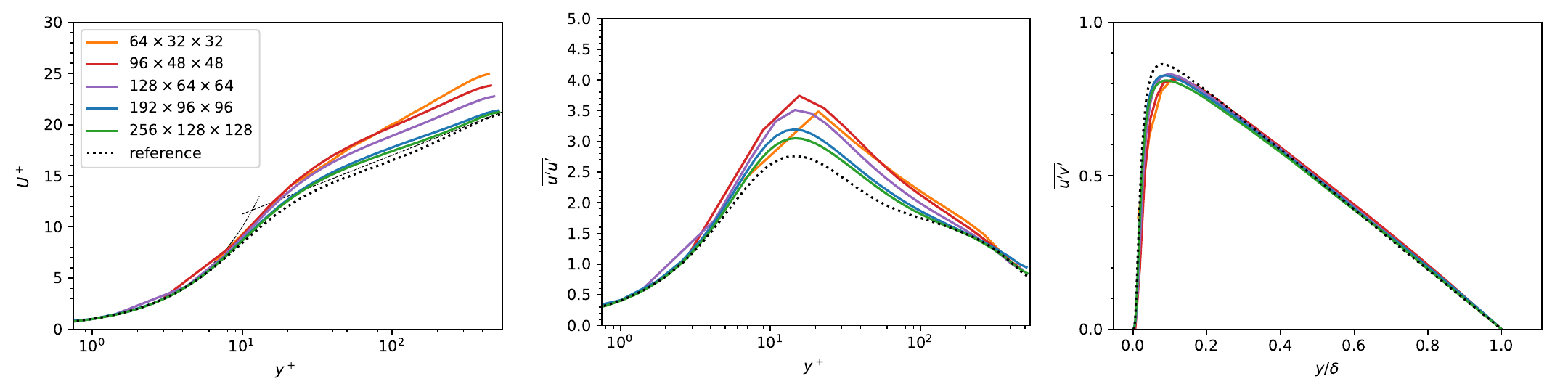}
    \caption{
    A convergence study simulating the TCF at increasing resolutions. The statistics have been non-dimentionalized with their respective average $\velwall$, which approaches the expected with increasing resolution. The reference is from Hoyas and Jim\'enez \cite{TCF_2008_10}.}
    \label{fig:app:TCFconvergence}
\end{figure}
We run a convergence study simulating the TCF at increasing resolutions, key statistics of which are shown in \myreffig{fig:app:TCFconvergence}. Both the mean flow $U^+$ and the turbulent fluctuations $\overline{\velcomp_i'^+ \velcomp_i'^+}$ converge towards the high-fidelity reference~\cite{TCF_2008_10}, while $\overline{\velu'^+ \velv'^+}$ stays close in any case.
The turbulent fluctuations do not seem to fully converge on the reference, which might be due to dissipation caused by the low order discretization scheme of PICT.
While this might make the present implementation not suited for high-fidelity turbulence simulation, it is nevertheless a close approximation suitable for training hybrid solvers.
The respective average $\Rewall$ also converge towards the target of $\Rewall = 550$:
$\text{Re}_{\tau, 64 \times 32 \times 32} = 467$,
$\text{Re}_{\tau, 96 \times 48 \times 48} = 479$,
$\text{Re}_{\tau, 128 \times 64 \times 64} = 502$,
$\text{Re}_{\tau, 192 \times 96 \times 96} = 534$,
$\text{Re}_{\tau, 256 \times 128 \times 128} = 541$.

\subsubsection{Higher Order Moments} \label{app:3DTCF:moments}

\begin{figure}
    \centering
    \begin{subfigure}{0.49\textwidth}
        \includegraphics[width=\linewidth]{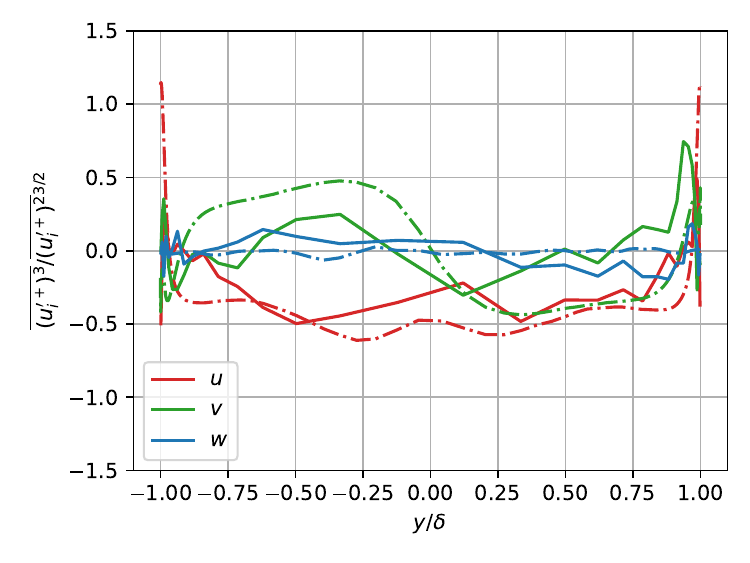}
        \caption{Skewness CNN SGS}
    \end{subfigure}
    \hfill
    \begin{subfigure}{0.49\textwidth}
        \includegraphics[width=\linewidth]{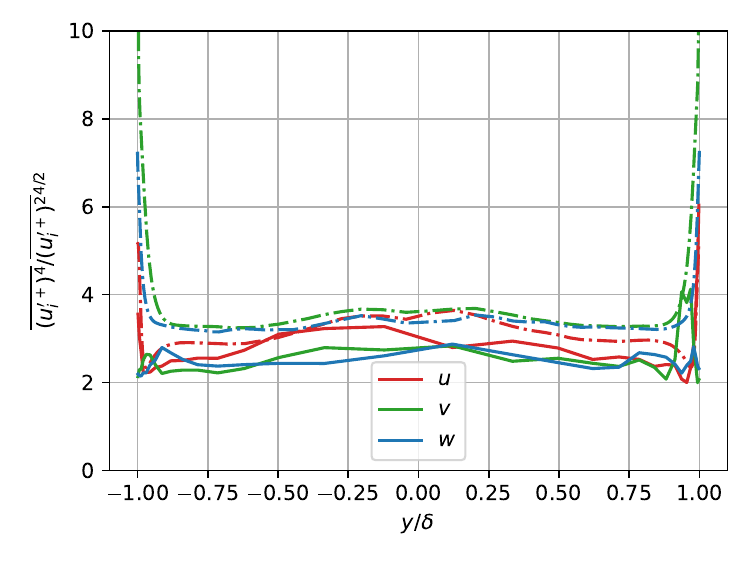}
        \caption{Flatness CNN SGS}
    \end{subfigure}
    \\
    \begin{subfigure}{0.49\textwidth}
        \includegraphics[width=\linewidth]{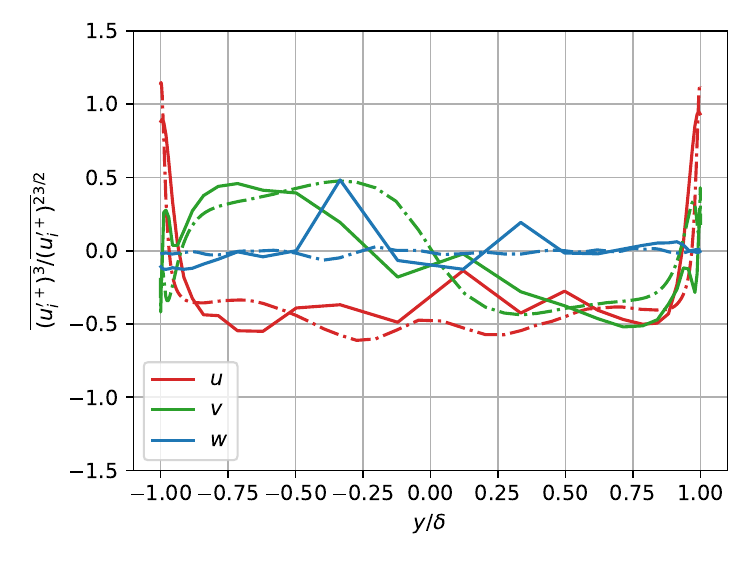}
        \caption{Skewness no SGS}
    \end{subfigure}
    \hfill
    \begin{subfigure}{0.49\textwidth}
        \includegraphics[width=\linewidth]{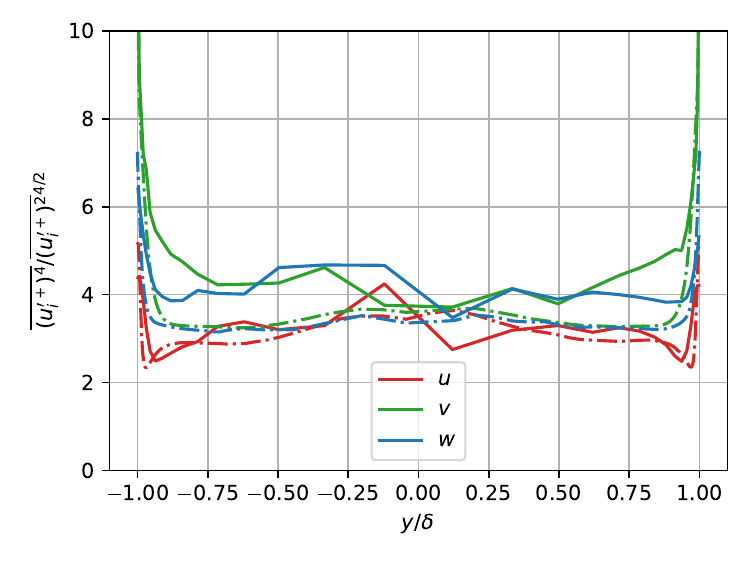}
        \caption{Flatness no SGS}
    \end{subfigure}
    \caption{
    Skewness (3rd standardized moment) and flatness (4th standardized moment) of our learned CNN SGS and no SGS, accumulated over 20 ETT. The dash-dotted lines are a high-resolution $(256 \times 128 \times 128)$ reference simulated with PICT, which is close to the values reported by Kim et al.~\cite{Kim_Moin_Moser_1987}.}
    \label{fig:app:highMoments}
\end{figure}
In addition to the mean and (co)variance, we show the higher order moments skewness and flatness of our learned CNN SGS in \myreffig{fig:app:highMoments} and compare to a high resolution simulation run with PICT.
The moments do not match the reference perfectly, but they nevertheless follow the same behavior qualitatively in the same value range.
Further, it is visible that the learned CNN SGS, which matches the mean and (co)variance it was trained on almost perfectly, does not show the correct behavior near the wall for both skewness and flatness. Here it is outperformed by the no SGS variant, which does not match the lower order statistics as well.
The discrepancy in the near-wall behavior is likely a consequence of the training setup. In fact, the corrector is optimized to minimize the error in the lower order statistics, without imposing any constraint on other physical quantities.

\subsubsection{Temporal Two-Point Correlation} \label{app:3DTCF:tempCorr}

\begin{figure}
    \centering
    \begin{subfigure}{0.32\textwidth}
        \includegraphics[width=\linewidth]{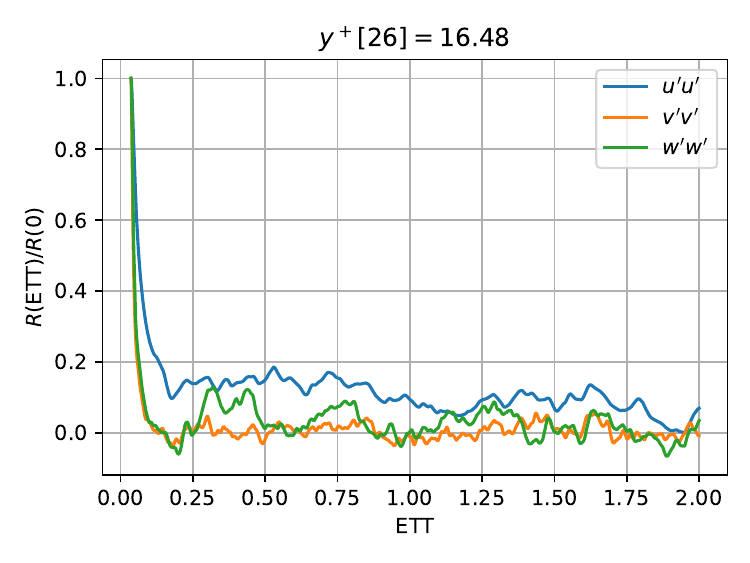}
        \caption{Temporal correlation HR} \label{fig:app:tempCorr:HR}
    \end{subfigure}
    \begin{subfigure}{0.32\textwidth}
        \includegraphics[width=\linewidth]{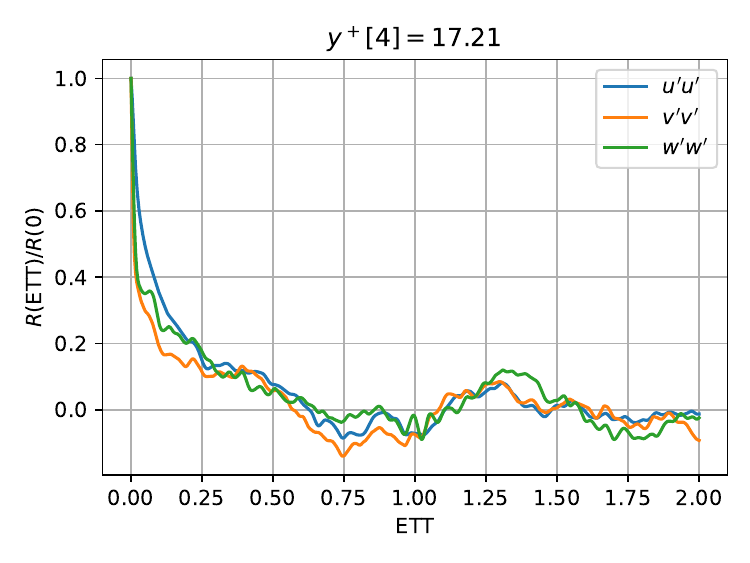}
        \caption{Temporal correlation CNN SGS} \label{fig:app:tempCorr:CNN}
    \end{subfigure}
    \begin{subfigure}{0.32\textwidth}
        \includegraphics[width=\linewidth]{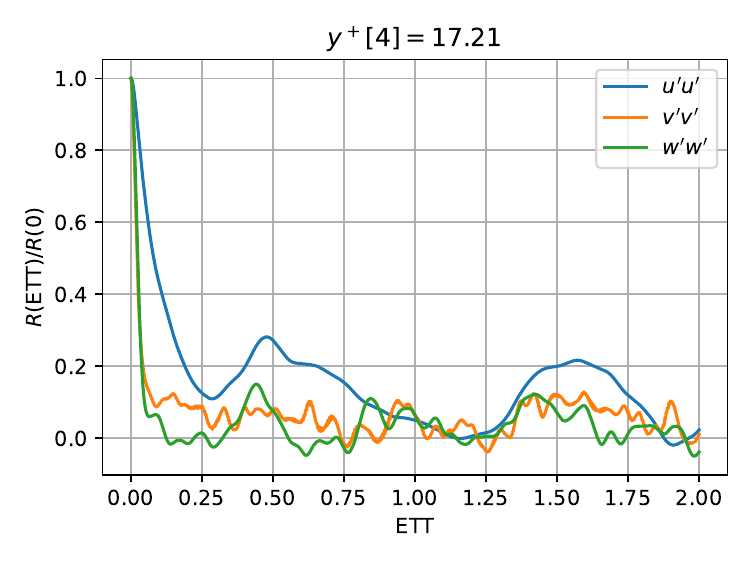}
        \caption{Temporal correlation no SGS} \label{fig:app:tempCorr:no}
    \end{subfigure}
    \caption{
    Temporal two-point correlation over 2 ETT, HR has a resolution of $(256 \times 128 \times 128)$ while CNN SGS and no SGS use $(64 \times 32 \times 32)$. The plots are per-model as the available wall-normal slices differ due to the different discretization of high and low-resolution. The shown wall-normal positions have been chosen to be close to $y^+ = 17$ for all cases.
    }
    \label{fig:app:tempCorr}
\end{figure}
In \myreffig{fig:app:tempCorr} we plot the Temporal two-point correlation coefficient~\cite{Chen2019twopointturbulent}
\begin{equation}
    R_{ij}(x, \Delta x, \tau) = \frac{\overline{\velcomp_i'(x, t_0) \velcomp_j'(x + \Delta x, t_0 + \tau)}}{\sqrt{\overline{\velcomp_i'^2(x, t_0)}} \sqrt{\overline{\velcomp_j'^2(x + \Delta x, t_0 + \tau)}}},
\end{equation}
with $\Delta x = 0$.
While all plots show a certain degree of noise, the PICT HR reference and the CNN SGS show significantly reduced spurious periodicity in the temporal correlation, which is present in the streamwise fluctuations of the no SGS case.
The CNN SGS also needs longer to de-correlate (until ca. 0.5 ETT) than the no SGS version.

\subsection{Aggregated Errors for TCF}\label{app:errorTcf}

\begin{table}
    \centering
    \begin{tabularx}{\textwidth}{X X X}\hline
        $\stats$ & PICT + CNN SGS & OpenFOAM \\\hline
        $U^+$ & $\mathbf{}4.91\times10^{-6}$ & $2.44\times10^{-3}$ \\
        $\overline{u'u'}$ & $3.09\times10^{-3}$ & $\mathbf{}1.25\times10^{-3}$ \\
        $\overline{v'v'}$ & $\mathbf{}2.18\times10^{-3}$ & $5.64\times10^{-3}$ \\
        $\overline{w'w'}$ & $2.42\times10^{-3}$ & $\mathbf{}2.30\times10^{-3}$ \\
        $\overline{u'v'}$ & $1.10\times10^{-3}$ & $\mathbf{}2.76\times10^{-4}$ \\\hline
        $\statsSet_\text{MSE}$ & $\mathbf{}8.78\times10^{-3}$ & $1.19\times10^{-2}$ \\\hline
    \end{tabularx}
    \caption{Individual and aggregate statistics errors for a TCF simulated with our learned SGS model and with OpenFOAM.
    The statistics used are plotted in the bottom row of \myreffig{fig:TCF550learnedSGS}.
    }
    \label{tab:TCFerrorCMP}
\end{table}

To quantify the accuracy of the TCF simulations, we employ an error calculation that aggregates different normalized error quantities as
\begin{equation}
    \begin{split}
        \statsSet_\text{MSE} &= \sum_{\stats \in \statsSet} \frac{1}{\text{max}( | \hat{\stats} | )} \frac{1}{Y} \sum_y \left| [ \stats ]_y - [ \overline{\hat{\stats}} ]_y \right|^2 \Delta y , \\
    \end{split}
\end{equation}
considering all statistics $\statsSet = \{U^+, \overline{u'u'}, \overline{v'v'}, \overline{w'w'}, \overline{u'v'}\}$ with equal weight.
$\hat{\stats}$ are the corresponding reference statistics~\cite{TCF_2008_10}, re-sampled
at the y-locations ($[\overline{\hat{\stats}}]_y$) of the computational mesh.
The individual statistics are normalized with the maximum of the reference $1/\text{max}(|\hat{\stats}|)$ to account for different magnitudes.
Due to the refinement of the mesh towards the walls, the discrete integral over all sample points $y$ is a mean weighted with the cells' size $\Delta y$. $Y$ is the total size of all cells under consideration. The resulting errors are presented in \myreftab{tab:TCFerrorCMP}.

\section{Additional Validation of PICT Gradients}

The gradients of the PICT solver are validated with a set of optimization problems below.

\subsection{Simple Optimization Problems} \label{app:diropt}
\begin{figure}
    \centering
    \includegraphics[width=\textwidth]{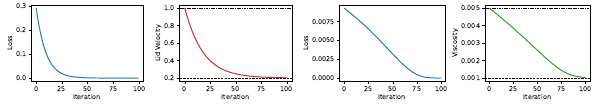}
    \caption{Convergence plots for a simple optimization on a lid-driven cavity setup. Left pair: optimization of the lid velocity. Right pair: optimization of the viscosity.
    } \label{fig:diropt}
\end{figure}
For an initial, simple optimization test, we run direct optimizations on two different low-dimensional flow quantities. These optimizations do not involve neural networks. The optimized quantities are viscosity and lid velocity, in the same lid-driven cavity setup that we also used for the validation of the forward simulation in \myrefapp{app:lidcavityFWD}.
Here, we use the 2D setup with a resolution of $32 \times 32$ and closed no-slip boundaries. The boundary at the lower y-border moves in x-direction as the driving lid.
As objective for the optimization we use a L2 loss to the velocity of a reference simulation,
which is backpropagated through the complete simulation rollout.
Hence, the resulting gradient contains terms from all operations in the simulator, among others the pressure solver.
For the optimization we use simple gradient descent without momentum, with a learning rate of $6\times10^{-2}$ for the lid velocity and $2\times10^{-5}$ for the viscosity.
When optimizing a quantity, it is initialized as $\velu_{init} = 1$ for lid velocity or $\viscosity_{init} = 0.005$ for viscosity. The target values are $\velu_{tar} = 0.2$ and $\viscosity_{tar} = 0.001$ respectively. This results in both the initial and target state having $\text{Re} = 200$.

We run the optimization for 100 iterations. Each iteration runs a simulation for 10 time units with adaptive time-step sizes based on the current lid velocity, which results in up to 40 simulation steps, and yields one update of the quantity to be optimized.
Both quantities converge against their respective value used in the reference simulation with a residual of
$6.44\times10^{-6}$ for lid velocity optimization and 
$5.46\times10^{-6}$ for viscosity.
The convergence is shown in \myreffig{fig:diropt}.

\begin{figure}
    \centering
    \includegraphics[width=0.72\textwidth,trim={0.0in 2.4in 0.0in 0.0in},clip]{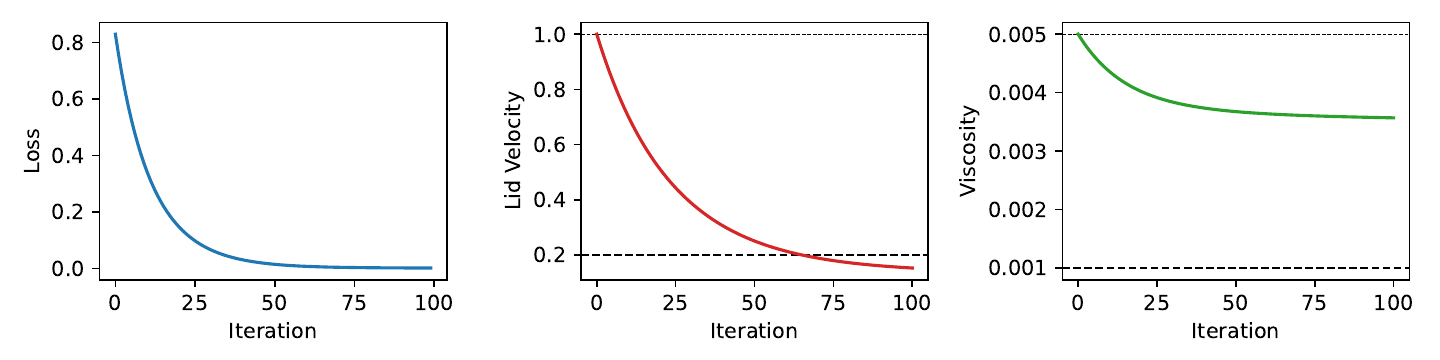}
    \caption{Convergence plots for a joint optimization of lid velocity and viscosity in a lid-driven cavity setup.
    } \label{fig:diroptJoint}
\end{figure}

For increased difficulty, we also target the task to jointly optimize viscosity and lid velocity.
In this setup, there is no unique solution when jointly optimizing viscosity and lid velocity, given the simple objective and fixed time horizon.
The combination of lid velocity and viscosity defines the magnitude of the velocity in the field at the final step, which is the objective,
meaning a higher velocity can compensate for a lower viscosity and vice versa.
While this results in flows of different Reynolds number that are visually distinct, it still causes the optimization to converge to a solution with low loss.
The exact solution found depends on the relative learning rates used.
A representative optimization run is shown in \myreffig{fig:diroptJoint}.

\section{Acknowledgements}
Funding: This work was supported by the European Research Council (ERC-2019-COG \#863850 SpaTe), and by the DFG Research Unit FOR 2987/1.

\newpage

\bibliography{main,felix-neurips,benjamin-icml,nils-merged-full}

\end{document}